\documentclass[iicol,pdflatex,sn-mathphys-ay]{sn-jnl}

\usepackage{graphicx}
\usepackage{multirow}
\usepackage{array}
\usepackage{amsmath,amssymb,amsfonts}
\usepackage{bm}
\usepackage{textcomp}
\usepackage{url}
\usepackage{xcolor}
\usepackage{booktabs}
\usepackage{enumitem}
\usepackage{algorithm}
\usepackage{algorithmicx}
\usepackage{algpseudocode}
\usepackage[caption=false]{subfig}
\usepackage{microtype}

\newenvironment{scaledtable}[1][t]
  {\begin{tableorg}[#1]\centering}
  {\end{tableorg}}
\IfFileExists{preamble.tex}{\input{preamble}}{}

\providecommand{\eg}{e.g.}

\begin{document}

\title[HETA++]{HETA++: Global Structure-from-Motion with Hybrid Explicit Translation Averaging}

\author[1,2]{\fnm{Peilin} \sur{Tao}}
\author*[1,2]{\fnm{Hainan} \sur{Cui}}\email{hncui@nlpr.ia.ac.cn}
\author[1,2]{\fnm{Mengqi} \sur{Rong}}
\author*[1,2]{\fnm{Shuhan} \sur{Shen}}\email{shshen@nlpr.ia.ac.cn}

\affil[1]{\orgname{Institute of Automation, Chinese Academy of Sciences},
  \orgaddress{\city{Beijing}, \postcode{100190}, \country{China}}}
\affil[2]{\orgname{School of Artificial Intelligence, University of Chinese Academy of Sciences},
  \orgaddress{\city{Beijing}, \postcode{100049}, \country{China}}}

\abstract{
Global Structure-from-Motion (SfM) offers advantages over incremental methods in terms of efficiency and error distribution. However, the task of translation averaging remains challenging. Many existing methods rely solely on relative translations or feature tracks, which either degrade under collinear camera motion or are susceptible to outliers.
In this paper, we propose a novel hybrid explicit translation averaging framework that incorporates both relative translations and feature tracks. Specifically, we first refine the relative translations using global camera rotations and remove globally inconsistent relative translations. Next, we employ convex distance-based objective functions to estimate the initial camera positions and 3D points, followed by refinement using a non-bilinear angle-based objective function.
Furthermore, since camera rotations are fixed during translation averaging, inaccurate camera rotations can severely limit the accuracy of camera positions.
To address this issue, we then robustly refine both camera rotations and camera positions with selected feature tracks through bounded angle-based refinement and subsequent reprojection-based bundle adjustment.
In this step, feature tracks are selected to maintain a balanced spatial distribution and improve optimization efficiency.
Finally, we perform a complete bundle adjustment using all reliable feature tracks to refine the camera parameters and 3D points.
Extensive experiments on various sequential and unordered real-world datasets demonstrate the superior accuracy, robustness, and scalability of our approach, outperforming state-of-the-art methods in both accuracy and computational efficiency.
}


\maketitle

\section{Introduction}
Recovering camera motion and 3D structure from image collections is a fundamental problem in 3D computer vision, with broad applications in areas such as navigation~\citep{ORBSLAM3_TRO, murAcceptedTRO2015}, augmented reality~\citep{Liu2019AR,peng2022epar, quesada2023design}, multi-view stereo (MVS)~\citep{unimvsnet,schoenberger2016mvs}, and novel view synthesis~\citep{Mipnerf,kerbl3Dgaussians}. Structure-from-Motion (SfM) is a common and effective approach to solving this problem. 
It begins with feature extraction, image retrieval, and feature matching to establish correspondences between image pairs and construct a view graph. Based on the resulting multi-view feature correspondences, the camera parameters and 3D scene structure are then jointly refined by minimizing the reprojection error through bundle adjustment (BA)~\citep{triggs2000bundle,ren2022megba}.
However, 2D image feature points are often incorrectly matched under conditions such as significant illumination changes, occlusions, and duplicate structures. The highly non-convex nature of BA makes it sensitive to outlier feature tracks. 
Hence, enhancing robustness and accuracy while maintaining efficiency remains a longstanding challenge in SfM techniques.

The primary paradigm of Structure-from-Motion is incremental, as exemplified by COLMAP~\citep{schoenberger2016sfm}. This method begins by carefully selecting an image pair to create an initial model. 
Next, images with sufficient 2D--3D correspondences are registered using a Perspective-n-Point (PnP) solver~\citep{gao2003complete,ding2023revisiting}, together with RANSAC to reject outlier correspondences. The scene structure and camera poses are then progressively refined through repeated camera registration, triangulation, and BA. Although incremental methods~\citep{snavely2006photo,theia-manual,schoenberger2016sfm} achieve high reconstruction accuracy and strong robustness to outliers, their results can depend on the image registration order, leading to error accumulation and drift~\citep{holynski2020reducing}. Additionally, the repetitive, non-convex bundle adjustments significantly impede efficiency, rendering them unsuitable for large-scale scenes.

As an alternative paradigm, global approaches first estimate all global camera rotations simultaneously through rotation averaging~\citep{chatterjee2017robust, zhang2023revisiting}, followed by the estimation of global camera positions and 3D points through translation averaging~\citep{zhu2018very,liu2019robust,cai2021pose,manam2022correspondence} and triangulation~\citep{schoenberger2016sfm}, leading to substantially improved efficiency and uniform error distribution. Subsequently, all camera parameters and structures are refined through BA. Therefore, the key problem of global methods is to robustly and accurately estimate the initial camera poses and 3D points from relative camera poses and feature correspondences in the view graph.
For global rotation estimation, given relative rotations in the view graph, existing methods \citep{govindu2004lie,chatterjee2013efficient} based on Lie algebra structures have been extensively studied. By contrast, relative translations estimated from two-view geometry suffer from scale uncertainty and are more sensitive to low-parallax conditions \citep{liu2019robust, concha2021instant} and outlier feature matches \citep{olsson2010outlier,sim2006removing}, thereby complicating the global translation estimation problem. 
Classical global translation estimation methods \citep{ozyesil2015robust,zhuang2018baseline,manam2022correspondence} rely exclusively on relative translations. Although these methods are highly efficient, they encounter degeneracy issues when the camera motion trajectory is collinear \citep{cui2015global}. In the more common case where the camera motion trajectory is nearly collinear and camera triplets typically form skew triangles \citep{manam2023sensitivity}, slight perturbations in relative translations can lead to substantial changes in estimated camera positions, leading to instability in the estimation process.
Moreover, methods relying solely on relative translations require a sufficiently constrained camera-direction graph and may become degenerate when the graph lacks parallel rigidity~\citep{arrigoni2021viewing}.

To alleviate these limitations, some methods incorporate additional camera-to-point constraints from feature tracks into their objective functions.
Each feature track consists of cameras and corresponding feature rays that observe the same 3D point. Depending on whether the corresponding 3D points of feature tracks are estimated during optimization, these methods can be categorized as either implicit or explicit. In most implicit methods, 3D points are represented by feature rays and relative translations between cameras in their corresponding feature tracks. Using implicit 3D points, camera baseline scales are estimated based on the depth consistency of feature rays~\citep{jiang2013global,cui2015global,cui2016robust}, or additional camera-to-camera constraints are incorporated into the objective function~\citep{cui2015linear,liu2019robust,cai2021pose}. Although implicit methods avoid explicit 3D point optimization, they yield less robust and accurate representations.

For explicit methods, the 3D points are represented as three-dimensional vectors and explicitly estimated during global translation estimation.
A typical explicit method, 1DSfM~\citep{wilson2014robust}, incorporates both camera-to-camera and camera-to-point constraints into a non-convex, angle-based objective function to directly optimize camera positions and 3D points with random initialization using the Levenberg–Marquardt (LM) method. However, it often produces noisy solutions when dealing with outlier feature points in feature tracks. 
Similarly, the recent method, GLOMAP~\citep{pan2024glomap}, also leverages the LM method to optimize camera positions and 3D points with random initialization. 
It utilizes a bilinear angle-based objective function~\citep{zhuang2018baseline} by introducing additional variables for each feature ray. This approach achieves comparable accuracy to incremental methods, which demonstrates the robustness and accuracy of explicit methods with angle-based objective functions.
Although the bilinear objective function offers better convergence, the random initialization of this non-convex problem undermines robustness. 
The bilinear objective jointly estimates camera positions, 3D points, and the scales of relative translations or feature rays. These scale variables are redundant once camera positions and 3D points are optimized, increasing both time and memory costs and limiting scalability.
Our previous work, HETA~\citep{tao2024revisiting}, first uses a convex $L_1$ distance-based objective to jointly initialize camera positions and 3D points from refined relative translations and selected feature tracks. It then applies an $L_2$ angle-based objective with hybrid camera-to-camera and camera-to-point constraints. However, its block coordinate descent optimizer limits the effectiveness of this refinement.

\begin{figure}[tp]
    \centering
    \includegraphics[width=\linewidth, trim = 30mm 0mm 30mm 0mm, clip]{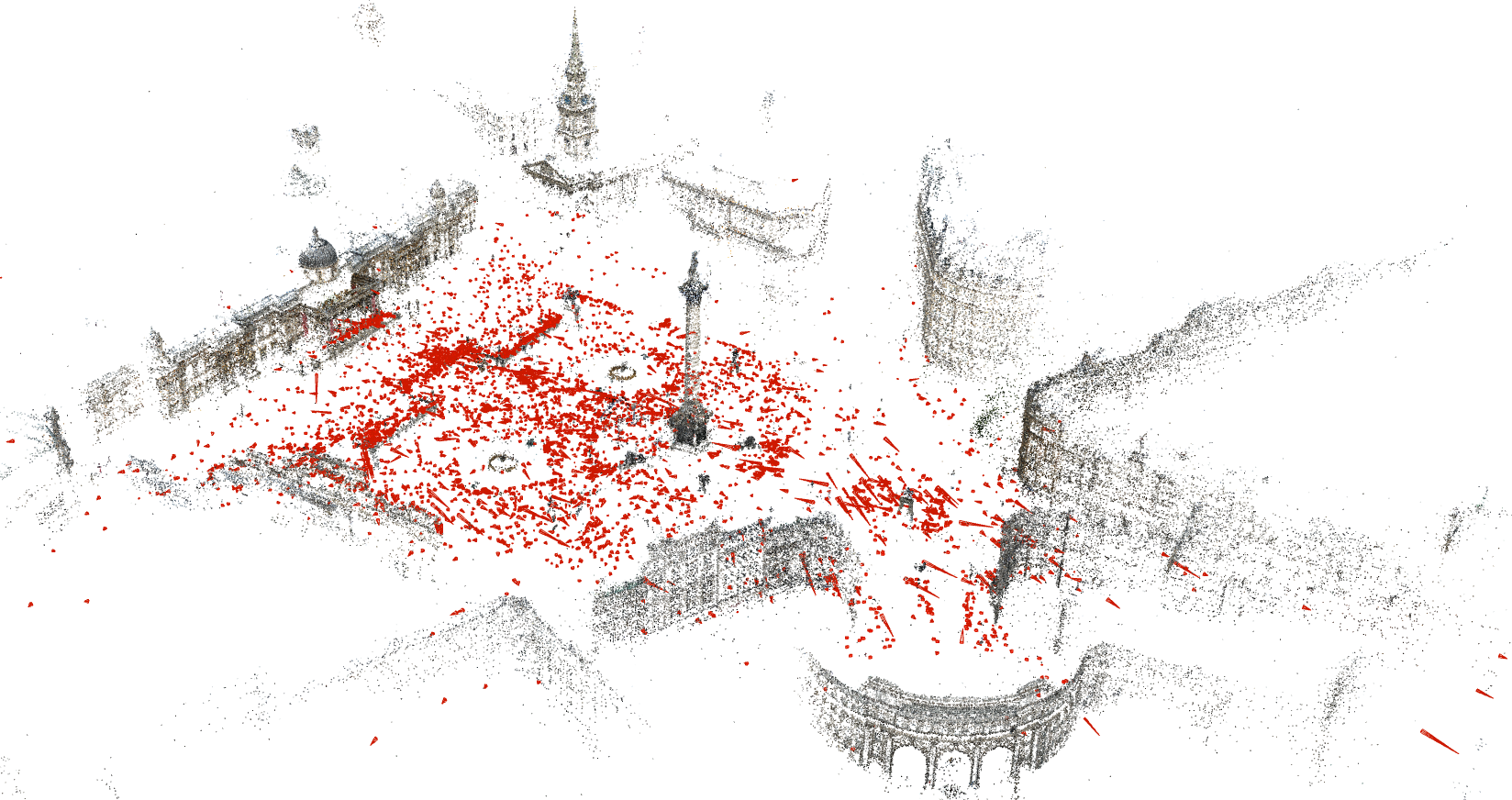}
    \caption{Reconstruction on Trafalgar~\citep{wilson2014robust}, containing over 5,000 images. Our method achieves more accurate camera positions than HETA~\citep{tao2024revisiting} and GLOMAP~\citep{pan2024glomap}, while running twice as fast as HETA and three times as fast as GLOMAP
}
    \label{fig:TFG}
    \vspace{-0.2cm}
\end{figure}

In this paper, we present an upgraded version of HETA, called HETA++. Compared to traditional explicit methods, such as 1DSfM \citep{wilson2014robust} and GLOMAP~\citep{pan2024glomap}, our method offers three key contributions. 
First, we introduce a hybrid explicit translation averaging approach for initialization, where camera positions and 3D points are initially estimated using a convex cross-product form distance-based objective function, and then refined with a compact non-bilinear angle-based objective function. This angle-based refinement minimizes the number of unknown parameters by focusing on camera positions and 3D points, thereby improving efficiency and reducing memory usage.
Second, since translation averaging is sensitive to the quality of relative translations, we re-estimate the relative translations using global camera rotations and remove globally inconsistent relative translations before initialization.
Third, we propose a joint optimization module to refine camera parameters and 3D points after initialization. Since camera rotations are fixed during initialization, insufficient precision in them can severely limit the accuracy of initialization.
To reduce reliance on the accuracy of initial camera rotations, we first improve camera poses with selected feature tracks through angle-based refinement, followed by reprojection-based bundle adjustment. 
In this step, a spatially balanced subset of feature tracks is selected to improve efficiency.
Finally, we perform a complete bundle adjustment to refine camera parameters and 3D points using all reliable feature tracks.

We improve HETA on four main aspects: 1) After re-estimating relative translations, we additionally filter relative translations based on global consistency to reduce the influence of outliers on translation averaging. 
2) In contrast to jointly estimating initial camera positions and 3D points with selected feature tracks in HETA~\citep{tao2024revisiting}, we first estimate camera positions via translation averaging using only relative translations and then apply alternating triangulation to estimate camera positions and 3D points respectively with all feature tracks in parallel, which improves the efficiency of initialization.
3) To avoid estimating redundant scale variables in each iteration of the BCD optimizer, we employ the LM optimizer for the non-bilinear angle-based objective function to improve convergence and efficiency.
4) Before the final bundle adjustment, we additionally refine camera poses using selected feature tracks in the joint optimization module.
Figure~\ref{fig:TFG} shows a large-scale reconstruction on Trafalgar~\citep{wilson2014robust}, demonstrating the robustness, accuracy, and scalability of our method.

\section{Related Work}
\label{sec:related}
\subsection{Global Rotation Averaging}

Global rotation averaging estimates absolute camera rotations from pairwise relative rotations~\citep{Hartley2013RotationA} and has been extensively studied. \citet{chatterjee2017robust} formulate the problem in the Lie algebra, while spectral and semidefinite relaxations under the chordal distance provide strong recovery guarantees~\citep{arie2012global,rosen2019se}. \citet{eriksson2018rotation} analyze its Lagrangian dual using spectral graph theory, and \citet{dellaert2020shonan} seek globally optimal solutions through a sequence of higher-dimensional lifts. Hierarchical initialization~\citep{lee2022hara}, end-to-end estimation~\citep{yang2021end}, and reliable weighting of relative rotations~\citep{zhang2023revisiting,sidhartha2021all} have also been explored. In this work, we adopt the robust method of \citet{chatterjee2017robust}.

\subsection{Global Translation Averaging}
Given global camera rotations, translation averaging methods estimate camera positions either from relative translations alone or by incorporating feature tracks. The latter can be further classified as implicit or explicit, depending on whether the associated 3D points are directly optimized.

\noindent\textbf{Pure Translation Averaging.} 
\citet{govindu2001combining} proposes a least-squares solution of
global translation estimation and refines translation with iterative weights.
Several works \citep{sim2006recovering,ke2007quasiconvex,kahl2008multiple,moulon2013global} use $L_\infty$ norm-based quasi-convex optimization to estimate camera translations. However, these methods necessitate meticulous handling of outliers \citep{ozyesil2015robust,wilson2014robust}.
\citet{ozyesil2015robust} propose a least unsquared deviations (LUD) formulation to enhance the robustness.
\citet{goldstein2016shapefit} propose to minimize the projection of $\bm{t}_i - \bm{t}_j$ on the orthogonal complement of $\bm{v}_{ij}$ under $L_1$ norm by the Alternating Direction Method of Multipliers (ADMM) method.
\citet{zhuang2018baseline} propose an angle-based formulation with an iteratively reweighted least squares (IRLS) scheme to mitigate the impact of different camera baselines.
\citet{zhu2018very} introduce a distributed framework to enhance the system efficiency in large-scale scene reconstructions.
\citet{manam2022correspondence} propose an iterative averaging scheme to filter outliers and refine relative translations with re-weighted feature matches.
While efficient, such methods are sensitive to outliers and produce inaccurate results in low-parallax scenes or collinear camera motions.

\noindent\textbf{With Implicit Feature Tracks.}
For implicit methods, 3D points are not estimated but are represented with feature rays. 
One category of approaches \citep{jiang2013global, cui2015global, cui2016robust} leverages the depth consistency of feature points to compute the camera baseline scales. 
\citet{jiang2013global} use constraints from the co-planarity of camera triplets and estimate baseline ratios to tackle the collinear motion problem. 
\citet{cui2015global} estimate the camera baseline scales based on a satellite graph, and then compute the camera motions by similarity averaging.
Similarly, based on the adjacent triangles in feature tracks, \citet{cui2016robust} utilize the law of sines to estimate the scale of camera baselines.
However, these methods are highly sensitive in low-parallax scenes. In these scenes, relative translation estimation is inaccurate, leading to incorrect angles between relative translations and feature rays. On the other hand, the low parallax angles result in numerical instability when estimating the relative baseline ratios in camera triplets. 

An alternative category of methods \citep{cui2015linear, liu2019robust, cai2021pose} represents 3D points with feature tracks and constrains cameras based on these represented 3D points and their corresponding feature rays. \citet{cui2015linear} propose to represent 3D points with relative translations and feature rays based on the rotation trick \citep{jiang2013global}.
Then a linear constraint is derived for cameras seeing a common 3D point. 
However, in low parallax scenes, the represented 3D points are unstable and the relative translations are error-prone.
To enhance the stability of representation, \citet{liu2019robust} propose to represent each 3D point with two cameras featuring a sufficient parallax angle.
Then, a linear constraint is constructed between the represented 3D point and the remaining cameras in each feature track.
However, the representation of 3D points in \citep{cui2015linear} and \citep{liu2019robust} still relies on relative translations, whose errors are accumulated into the represented 3D points.
To handle these concerns, \citet{cai2021pose} propose a linear global translation (LiGT) constraint, where 3D points are represented solely with feature rays, aiming to avoid the impact of errors in relative translations.

\begin{figure*}	
	\setlength{\abovecaptionskip}{0.0cm}
	\setlength{\belowcaptionskip}{-0.15cm}
	\centering
	\includegraphics[width=\linewidth, trim = 0mm 65mm 0mm 65mm, clip]{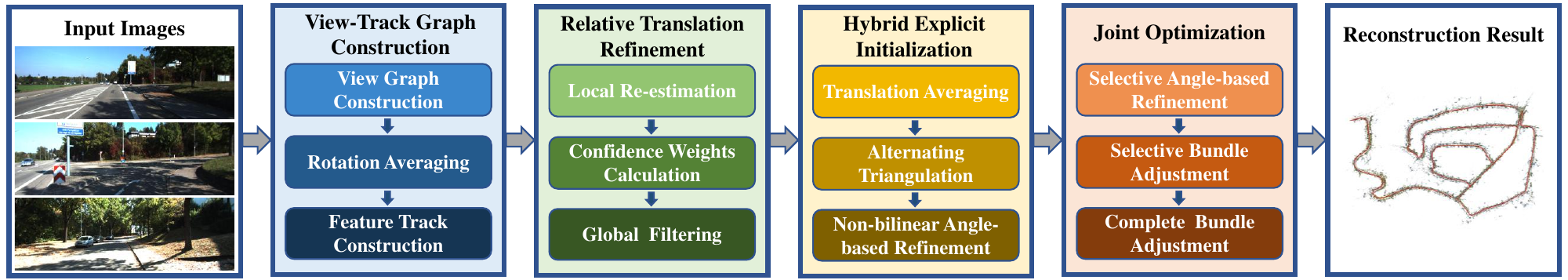}
	\caption{Pipeline of our method: Given a set of images, we first construct a view-track graph, where the nodes represent either images or 3D points, and the edges encode camera-to-camera and camera-to-point constraints. Next, the relative translations are locally refined and filtered based on global consistency. A hybrid explicit initialization module is then introduced to estimate the initial camera positions and 3D points, followed by a joint optimization that progressively refines camera parameters and 3D points. Finally, the reconstruction result is obtained, including the camera intrinsics, camera poses, and the 3D scene}

	\label{fig:pipline}
\end{figure*}

\noindent\textbf{With Explicit Feature Tracks.}
For explicit methods, camera translations and 3D points are estimated simultaneously.
Since the mathematical expressions for the camera-to-camera constraint and camera-to-point constraint are equivalent, the core idea is to estimate 3D points using the same objective function employed for estimating camera translations.
Compared to the error-prone relative translations in low parallax scenes, feature rays, as raw information derived from images, naturally exhibit higher precision.
Hence, using feature rays in explicit methods can theoretically deliver superior performance compared to methods that rely solely on relative translations.
For example, \citet{crandall2012sfm} employ a discrete Markov Random Field formulation to estimate cameras and 3D points. 
\citet{wilson2014robust} propose to first eliminate relative translation outliers through multiple one-dimensional projections and subsequently integrate both relative translations and feature tracks into a non-convex objective function.
\citet{pan2024glomap} utilize camera-to-point constraints from feature tracks to formulate a bilinear angle-based objective function, which can be optimized with random initialization and yields a good scene structure and camera poses for bundle adjustment.

In terms of robustness and accuracy, explicit methods typically perform better than implicit methods, since the 3D point for each feature track is estimated and refined with all feature rays by minimizing angle-based geometric errors in explicit methods, while the 3D point is usually linearly encoded only with two feature rays from base cameras in implicit methods. 

\subsection{Feed-Forward 3D Reconstruction}
Recent feed-forward methods directly predict camera poses and scene geometry from images, enabling efficient and robust reconstruction without a complete geometric SfM pipeline. DUSt3R~\citep{wang2024dust3r} predicts pairwise point maps in a common coordinate frame, while MASt3R~\citep{leroy2024grounding} further introduces dense descriptors for accurate correspondence estimation. VGGT~\citep{wang2025vggt} jointly predicts camera parameters, depth, point maps, and tracks from multiple views. $\pi^3$~\citep{wang2025pi} adopts a permutation-equivariant formulation to reduce dependence on image ordering and reference-view selection. 
MapAnything~\citep{keetha2026mapanything} supports flexible combinations of images and geometric priors, whereas 
Depth Anything 3~\citep{lin2025depth} provides a unified framework for depth and multi-view reconstruction. These methods are efficient, generalizable, and robust, but their lack of explicit multi-view geometric optimization limits reconstruction accuracy and global consistency compared with geometry-based SfM.

\section{Hybrid Explicit Translation Averaging} \label{sec:method}

Given a set of images, we first construct the view-track graph, which involves view graph construction, global rotation averaging, and feature track generation. To better interpret the objective function used in our global translation estimation, we compare several convex and non-convex objective functions related to camera-to-camera and camera-to-point constraints. Next, we introduce a local-to-global strategy for refining the relative translations. With hybrid inputs, including refined relative translations and feature tracks, we propose a two-step scheme for initializing camera positions and 3D points. The first step involves robust convex initialization using translation averaging and alternating triangulation under the $L_1$ norm, followed by non-bilinear angle-based refinement using both camera-to-camera and camera-to-point constraints under the $L_2$ norm. 
Finally, to provide a good starting point for the complete BA, we select a spatially balanced subset of reliable feature tracks for each image to improve camera poses with angle-based refinement and bundle adjustment, where the selected feature tracks ensure a balanced error distribution and improve optimization efficiency. 
An overview of our framework is shown in Fig.~\ref{fig:pipline}.
 

\subsection{View-Track Graph Construction}
\label{sec:preprocessing}
Given an image collection $\mathcal{I}=\{I_1,\cdots,I_N\}$, we first extract 
feature points (\eg, SIFT \citep{lowe2004distinctive}) and use the NetVLAD ~\citep{arandjelovic2016netvlad} method to retrieve the top-K most similar images for each image.
Next, we evaluate each potential image pair by performing brute-force feature matching and then using the RANSAC algorithm to identify inlier feature matches through two-view geometry verification.
When the camera intrinsics are approximately known, the relative rotations are estimated from the fundamental, essential, or homography matrices through two-view geometry. Valid relative translation directions are retained only for image pairs with sufficient translation and parallax, while unreliable translation constraints are discarded before translation averaging.

Given matched image pairs, a view graph $\mathcal{G}_c = \{\mathcal{V}_c, \mathcal{E}_c\}$ is constructed, where each node in $\mathcal{V}_c$ represents an image or its corresponding camera, and each edge $ij \in \mathcal{E}_c$ represents the relative pose  $(\bm{R}_{ij},\bm{t}_{ij})$ and corresponding feature matches between two matched images $i,j \in \mathcal{V}_c$.
The global camera poses $(\bm{R}_{i},\bm{t}_{i})$ and the relative camera poses $(\bm{R}_{ij},\bm{t}_{ij})$ satisfy the following equations:
\begin{equation}
\label{equ:camera pose}
\bm{R}_j \bm{R}_i^T = \bm{R}_{ij}, \quad \frac{\bm{t}_i-\bm{t}_j}{||\bm{t}_i-\bm{t}_j||_2} = \bm{R}_j^T\bm{t}_{ij} = \bm{v}_{ij}.
\end{equation}
Here $\bm{v}_{ij}$ denotes the relative translation in the global coordinate system.

Given the relative rotations in $\mathcal{E}_c$, the global camera rotations are estimated through rotation averaging methods.
The objective function is generally formulated as follows:
\begin{equation}
\label{equ:rotation averaging}
\{\tilde{\bm{R}}_i\}_{i\in\mathcal{V}_c} = \arg\min_{\bm{R}_i,i\in \mathcal{V}_c}\sum_{ij\in \mathcal{E}_c}\rho(d(\bm{R}_{ij},\bm{R}_j\bm{R}_i^T)),
\end{equation}
where $\rho(\cdot)$ represents the robust estimator function, and $d(\cdot)$ represents the distance metric like geodesic distance, chordal distance, or quaternion distance.
In this paper, we use the method proposed by \citet{chatterjee2017robust} to estimate global camera rotations. This method leverages geodesic distance by first initializing the rotations under the $L_1$ norm for a few iterations, followed by refinement using the iteratively reweighted least squares (IRLS) method.

Given the estimated global camera rotations, we compute the globally consistent relative rotations as $\tilde{\bm{R}}_{ij}=\tilde{\bm{R}}_j\tilde{\bm{R}}_i^T$. An image pair is discarded if the angular difference between its original two-view relative rotation $\bm{R}_{ij}$ and the globally consistent relative rotation $\tilde{\bm{R}}_{ij}$ exceeds $10^\circ$.
Next, based on the feature correspondences in the filtered view graph, we utilize the union-find algorithm~\citep{moulon2012unordered} to construct the view track graph $\mathcal{G} = \{\mathcal{V}_c \bigcup \mathcal{V}_p, \mathcal{E}_c \bigcup \mathcal{E}_p\}$, where each node in $\mathcal{V}_p$ represents a 3D point, and each edge $ik \in \mathcal{E}_p$ indicates that the 3D point $k \in \mathcal{V}_p$ is observed by the camera $i \in \mathcal{V}_c$.
We denote the global position coordinates of the 3D point $k$ and camera $i$ as $\bm{P}_k$ and $\bm{t}_i$, respectively, and the pixel coordinates of the feature point in image $i$ observing 3D point $k$ as $\bm{x}_{ki}=(u_{ki},v_{ki})^T$.
The feature point in the normalized camera coordinate system is given by $\bm{X}{ki}=\bm{\pi}i^{-1}(u{ki},v{ki},1)^T$, where $\bm{\pi}_i$ denotes the camera intrinsics.
The relationship between the 3D point $k$ and camera $i$ is given by:
\begin{equation}
\label{equ:camera to point}
\frac{\bm{P}_k-\bm{t}_i}{||\bm{P}_k-\bm{t}_i||_2} =\bm{R}_i^T \frac{\bm{X}_{ki}}{||\bm{X}_{ki}||_2}= \bm{R}_i^T \bm{\hat{X}}_{ki} = \bm{f}_{ki}.
\end{equation}
Here $\bm{f}_{ki}$ denotes the normalized feature ray from camera $i$ to 3D point $k$.
Both the camera-to-camera and camera-to-point relations are directional bearing constraints between two unknown 3D positions. Although they differ in their geometric roles and observation properties, they share the same normalized displacement form. Therefore, the same class of direction-based objective functions can be applied to both types of constraints.

\subsection{Definition of Objective Function}
\label{sec:objective}
\begin{figure}[tp]
    \setlength{\abovecaptionskip}{0.0cm}
	\setlength{\belowcaptionskip}{-0.1cm}
    \centering
    \includegraphics[width=\linewidth, trim = 11mm 45mm 6mm 52mm, clip]{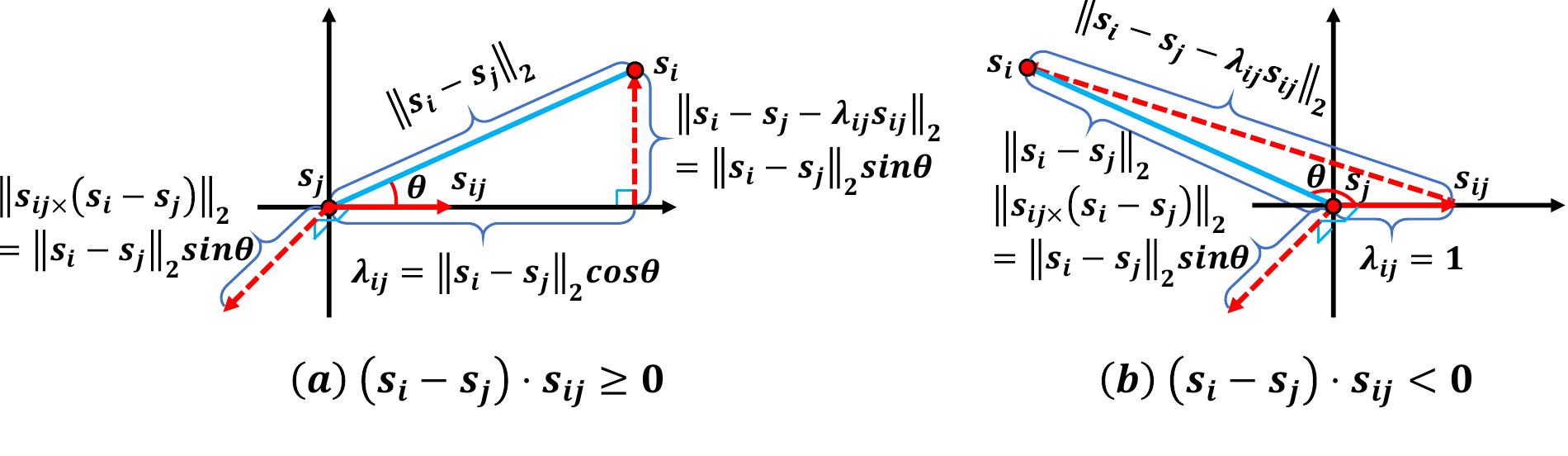}
    \caption{
    Residuals of two linear objective functions: $||\bm{s}_{ij\times}(\bm{s}_{i}-\bm{s}_{j})||_2$ and $||\bm{s}_{i}-\bm{s}_{j}-\lambda_{ij}\bm{s}_{ij}||_2$ under different circumstances}
    \label{fig:objective}
\end{figure}
For Eq.~\eqref{equ:camera pose} and Eq.~\eqref{equ:camera to point}, with known global camera rotations, both camera-to-camera and camera-to-point constraints can be represented by the following formulation:
\begin{equation}
\label{equ:model}
\bm{s}_i - \bm{s}_j = ||\bm{s}_i - \bm{s}_j||_2 \cdot \bm{s}_{ij},
\end{equation}
where $\bm{s}_i,\bm{s}_j$ represent camera positions or 3D points and $\bm{s}_{ij}$ is a known normalized observation vector from $\bm{s}_j$ to $\bm{s}_i$, such as a feature ray or a relative translation.
Convex distance-based objective functions, such as LUD~\citep{ozyesil2015robust}, exhibit good convergence properties but place greater emphasis on observations with large scales, such as long baselines for relative translations or large depths for feature rays. In contrast, non-convex angle-based objective functions, such as 1DSfM~\citep{wilson2014robust}, are sensitive to outlier feature matches and often suffer from poor convergence; however, they treat all observations with varying scales equally. Therefore, we use convex objective functions to initialize the solution and then employ angle-based objective functions for refinement. Below, we compare several convex distance-based objective functions in Section \ref{sec:convex} and several non-convex angle-based objective functions in Section \ref{sec:non-convex}.

\subsubsection{Convex distance-based Objective Function}
\label{sec:convex}
We compare two types of convex distance-based objective functions: the cross-product form, $\left\|\bm{s}_{ij}\times(\bm{s}_i-\bm{s}_j)\right\|_2$, and the scale form, $\left\|\bm{s}_i-\bm{s}_j-d_{ij}\bm{s}_{ij}\right\|_2$, where $\bm{s}_{ij}$ is the known unit observation direction, while $\bm{s}_i$, $\bm{s}_j$, and the auxiliary scale $d_{ij}$ are jointly estimated. The auxiliary scale represents the unknown displacement magnitude associated with the directional observation. To eliminate the trivial zero-scale solution and preserve the orientation of each directional observation, we impose $\bm{s}_{ij}^{T}(\bm{s}_i-\bm{s}_j)\geq 1$ for the cross-product form and $d_{ij}\geq 1$ for the scale form. Although $d_{ij}$ is jointly optimized with the positions, its role can be understood by considering its conditional optimum for fixed $\bm{s}_i$ and $\bm{s}_j$. Specifically, the corresponding one-dimensional subproblem is $\min_{d_{ij}\geq 1}\left\|\bm{s}_i-\bm{s}_j-d_{ij}\bm{s}_{ij}\right\|_2$. Since $\bm{s}_{ij}$ is normalized, its solution is $d_{ij}^{*}=\max\left\{1,\bm{s}_{ij}^{T}(\bm{s}_i-\bm{s}_j)\right\}$. Therefore, when $\bm{s}_{ij}^{T}(\bm{s}_i-\bm{s}_j)\geq 1$, the optimal scale equals the projection length of the estimated displacement onto the observed direction. This conditional solution shows that $d_{ij}$ does not encode an additional independent geometric quantity, although it remains an explicit variable in the joint optimization.

When the observed direction contains a severe error and points to the opposite hemisphere of the ground-truth displacement, the unconstrained projection becomes negative. In the scale formulation, the positivity constraint prevents the auxiliary scale from reversing the observation direction, and its conditional optimum reaches the lower bound $d_{ij}^{*}=1$. The incorrect observation is therefore represented by a finite residual rather than imposing a hard directional constraint on the estimated positions. In contrast, the inequality constraint of the cross-product formulation directly restricts the displacement to the half-space defined by the erroneous observation direction, which may bias the feasible solution.

Cross-product direction constraints have been widely used in global motion and translation estimation since early work such as~\citep{govindu2001combining}.
Compared with the scale form, the cross-product form avoids explicitly optimizing one auxiliary scale variable for each directional observation and provides a more direct inequality constraint. When the observation scales vary substantially, as in cases involving disparate camera baselines or feature-ray depths, the scale form is more difficult to optimize efficiently, whereas the cross-product form generally exhibits better convergence and lower computational cost, as demonstrated by \citet{tao2024revisiting}. However, its inequality constraint may introduce bias when an observation contains a severe directional error. Since such erroneous relative translations are substantially reduced through local re-estimation and global consistency filtering, we employ the cross-product form with the retained relative translations to initialize the camera positions.

\subsubsection{Non-convex Angle-based Objective Function}
\label{sec:non-convex}
Compared to the convex distance-based objective function, the non-convex angle-based objective function is unbiased \citep{zhuang2018baseline} and bounded \citep{pan2024glomap}, making it suitable for filtering outliers and robustly providing a good starting point for unbounded reprojection-based bundle adjustment.

We denote $\bm{\hat{s}}_{ij}=\frac{\bm{s}_i-\bm{s}_j}{||\bm{s}_i-\bm{s}_j||_2}$ and $\theta=\arccos{(\bm{s}_{ij}\cdot\bm{\hat{s}}_{ij})}$.
Pioneering method 1DSfM \citep{wilson2014robust} formulates the angle-based objective function by directly calculating the chordal distance between $\bm{s}_{ij}$ and $\bm{\hat{s}}_{ij}$ by:
\begin{equation}
\label{eq:chordal}
    \mathbb{C}(\bm{s}_{ij},\bm{\hat{s}}_{ij})=||\bm{s}_{ij} - \bm{\hat{s}}_{ij}||_2 = 2\sin{(\theta/2)}.
\end{equation}
However, this chordal distance objective function, when initialized randomly, is prone to converging to local optima due to outliers.
To address this, the bilinear angle-based objective function \citep{zhuang2018baseline,pan2024glomap} is formulated by integrating an additional variable $\lambda_{ij}$ as:
\begin{equation}
\label{eq:bilinear}
||\bm{s}_{ij} - \lambda_{ij}(\bm{s}_i - \bm{s}_j)||_2, \quad \text{s.t.} \quad \lambda_{ij} \ge 0.
\end{equation}
When $\theta \in [0,\pi/2)$, for optimal solutions, Eq.~\eqref{eq:bilinear} is equivalent to calculating the orthogonal distance from $\bm{s}_{ij}$ to $\bm{s}_i - \bm{s}_j$.
In this case, $\lambda_{ij}=\frac{\bm{s}_{ij}\cdot\bm{\hat{s}}_{ij}}{||\bm{s}_i-\bm{s}_j||_2}$ and the residual magnitude equals $\sin{\theta}$.
When $\theta \in [\pi/2, \pi]$, for optimal solutions, $\lambda_{ij}=0$ and the residual magnitude always equals $1$, which makes Eq.~\eqref{eq:bilinear} more robust than Eq.~\eqref{eq:chordal}.
While the redundant scale variable $\lambda_{ij}$ in the bilinear Eq.~\eqref{eq:bilinear} improves convergence and robustness, optimizing the additional variables for each term is both time and memory-consuming.


To eliminate the auxiliary scale variable $\lambda_{ij}$, we formulate a non-bilinear orthogonal-distance objective similar to Eq.~\eqref{eq:bilinear}, but without explicitly optimizing the scale variable. Here, $\bm{s}_{ij}$ is the known unit observation direction, and $\bm{\hat{s}}_{ij}=(\bm{s}_i-\bm{s}_j)/\left\|\bm{s}_i-\bm{s}_j\right\|_2$ is the predicted direction determined by the position variables:
\begin{equation}
\label{eq:non-bilinear}
\mathbb{O}\!\left(\bm{s}_{ij},\hat{\bm{s}}_{ij}\right)
=
\begin{cases}
\left\|
\left(\bm{I}
-\hat{\bm{s}}_{ij}\hat{\bm{s}}_{ij}^{T}\right)
\bm{s}_{ij}
\right\|_2,
& 0\leq\theta<\frac{\pi}{2},
\\[2pt]
1,
& \frac{\pi}{2}\leq\theta\leq\pi .
\end{cases}
\end{equation}
For $\theta<\pi/2$, Eq.~\eqref{eq:non-bilinear} measures the orthogonal distance between the observed and predicted directions. For $\theta\geq\pi/2$, the residual is saturated at one, consistent with the optimal solution of the bilinear formulation under the non-negative scale constraint. With the proposed initialization, this compact non-bilinear formulation provides competitive results without explicitly optimizing per-observation scale variables.

\subsection{Relative Translations Refinement}
\label{sec:rel_trans_refine}
Relative translations offer more direct and robust constraints between cameras than raw feature rays. These constraints can be used to efficiently initialize camera positions and improve convergence during subsequent optimization. After rotation averaging, the optimized global rotations facilitate a more accurate re-estimation of relative translations by consolidating raw camera-to-point constraints from feature matches based on two-view geometry.
However, two significant challenges must be addressed. First, using co-planarity constraints in two-view geometry to estimate relative translations suffers from degeneration, particularly in cases of short baselines or pure rotation. Second, due to the lack of constraints from multiple views, local estimation using only two views is highly sensitive to incorrect feature matches.

To address these issues, we proceed in two steps. During local relative translation re-estimation, we place greater emphasis on feature matches with larger parallax angles, as these provide more reliable co-planarity constraints, as detailed in Section \ref{sec:reestimate}. 
Furthermore, the distribution of the epipolar planes of feature matches also affects the uncertainty in relative translation estimation. Therefore, we first utilize the eigenvalue ratio of the optimization matrix from normal vectors to model this distribution, as detailed in Section \ref{sec:score}. Finally, we evaluate heuristic confidence scores for each relative translation and employ the computed confidence scores to filter out outliers by ensuring global directional consistency of all relative translations in Section \ref{sec:global_filter}.

\begin{figure}[tp]
    \setlength{\abovecaptionskip}{-0.0cm}
	\setlength{\belowcaptionskip}{-0.1cm}
    \centering
    \includegraphics[width=0.6\linewidth, trim = 75mm 50mm 80mm 58mm, clip]{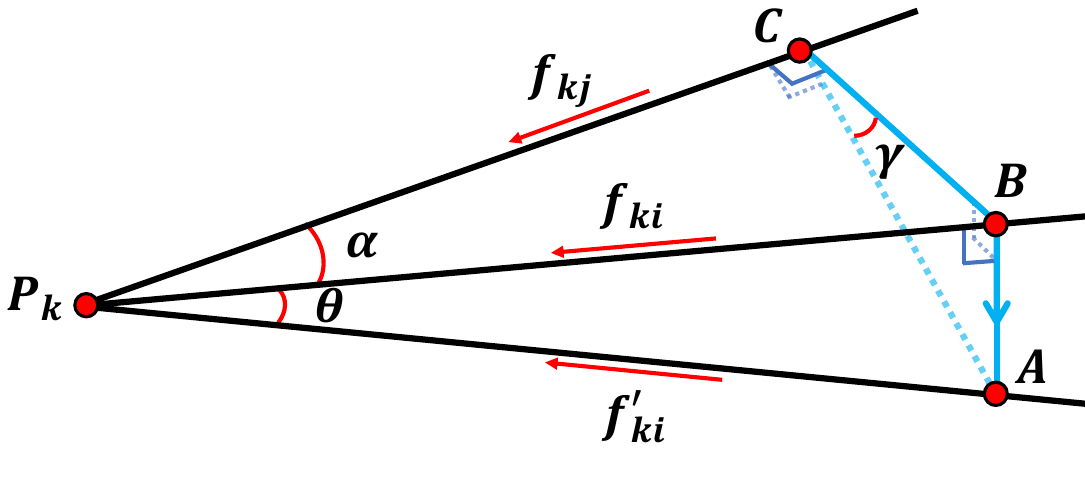}
    \caption{A toy example showing how angular errors in feature rays affect the normal vector of epipolar plane}
    \label{fig:Paralax}
    \vspace{-0.3cm}
\end{figure}
\subsubsection{Local Relative Translation Re-estimation}
\label{sec:reestimate}
In two-view epipolar geometry, the co-planarity constraint is defined as: $\bm{X}_{kj} \cdot (\bm{t}_{ij} \times  \bm{R}_{ij} \bm{X}_{ki}) = 0$.
Given global camera rotations, this constraint is rewritten as:
\begin{equation}
\label{equ:coplanar_global}
\begin{aligned}
\Leftrightarrow 
(\bm{R}_{ij}\bm{X}_{ki} \times \bm{X}_{kj})^T \bm{t}_{ij}&=0\\
\Leftrightarrow 
(\bm{R}_j^T(\bm{R}_{ij}\bm{X}_{ki}\times\bm{X}_{kj}))^T \bm{R}_j^T\bm{t}_{ij}&=0\\
\Leftrightarrow 
(\bm{R}_i^T \bm{X}_{ki} \times \bm{R}_j^T \bm{X}_{kj})\cdot \bm{R}_j^T \bm{t}_{ij} & = 0 \\
\Leftrightarrow (\bm{f}_{ki} \times \bm{f}_{kj}) \cdot \bm{v}_{ij} &=  0,
\end{aligned}
\end{equation}
where $\bm{v}_{ij}$ is the same as defined in Eq.~\eqref{equ:camera pose}.
For Eq.~\eqref{equ:coplanar_global}, each relative translation can be re-estimated using the normal vectors of epipolar planes, which are calculated by $\bm{f}_{ki} \times \bm{f}_{kj}$.
Due to inaccuracies in both camera intrinsic parameters and global rotations, the normal vectors estimated from the feature rays inevitably exhibit angular errors.
The method presented by \citet{ozyesil2015robust} re-estimates relative translations using all normalized normal vectors by minimizing the cosine of the angles between the relative translations and normal vectors, which is equivalent to averaging the sine values of the angular error in the normal vectors.  
However, since the accuracy of the normal vectors is also influenced by the parallax angles, it is unreasonable to apply the same weight to each normal vector during estimation.
To investigate how angular errors in feature rays affect the normal vectors across varying parallax angles, we decompose these errors into components along both the normal direction and the epipolar plane direction. 
Since errors along the epipolar plane direction do not affect the direction of the normal vector, for simplicity, we focus solely on errors along the normal direction.

As shown in Fig.~\ref{fig:Paralax}, two feature rays $\bm{f}_{ki}, \bm{f}_{kj}$ triangulate a 3D point $\bm{P}_k$ with a parallax angle $\alpha$. 
A minor angular error $\theta$ occurring in $\bm{f}_{ki}$ along the normal direction results in a deviation from $\bm{f}_{ki}$ to $\bm{f}_{ki}^\prime$.
We mark a point $A$ on $\bm{f}_{ki}^\prime$ and extend a perpendicular line from point $A$ to $\bm{f}_{ki}$, intersecting it at point $B$. Subsequently, we extend another perpendicular line from point $B$ to $\bm{f}_{kj}$, intersecting it at point $C$.
As the angular error $\theta$ is along the normal direction, line $AB$ is perpendicular to the plane $\{P_k-B-C\}$. Hence, line $P_k C$ is perpendicular to plane $\{A-B-C\}$, implying that the angle $\gamma$ between $AC$ and $BC$ is equal to the angular error of the normal vector. 
According to the geometric relations in Fig.~\ref{fig:Paralax}, the angular deviation $\gamma$ of the epipolar-plane normal satisfies
\begin{equation}
\label{equ:epipolar_plane}
\tan{\gamma}
=\frac{\left\|AB\right\|_2}{\left\|BC\right\|_2}
=\frac{\left\|P_kB\right\|_2\tan{\theta}}{\left\|P_kB\right\|_2\sin{\alpha}}
=\frac{\tan{\theta}}{\sin{\alpha}}.
\end{equation}
For a fixed feature-ray perturbation $\theta$, Eq.~\eqref{equ:epipolar_plane} shows that the angular deviation $\gamma$ decreases as the parallax angle $\alpha$ increases. Therefore, feature matches with larger parallax angles provide more stable epipolar-plane normals, whereas small-parallax matches are more sensitive to perturbations in the feature rays.
Hence, in contrast to employing the  normalized normal vector $\frac{\bm{f}_{ki} \times \bm{f}_{kj}}{||\bm{f}_{ki} \times \bm{f}_{kj}||_2}$ as used by \citet{ozyesil2015robust}, we maintain the reasonable weight $||\bm{f}_{ki} \times \bm{f}_{kj}||_2 = \sin{\alpha}$ for each feature match during estimation.
Next, relative translations are estimated using an IRLS scheme \citep{holland1977robust} as follows:
\begin{equation}
\label{equ:re-estimation}
\min_{\bm{v}_{ij}}
\sum_k \rho\left(\left|(\bm{f}_{ki} \times \bm{f}_{kj}) \cdot \bm{v}_{ij}\right|\right) \quad s.t. \quad||\bm{v}_{ij}||_2=1.
\vspace{-0.1cm}
\end{equation}
The robust estimator function is Cauchy loss function $\rho(\varepsilon) = \log(\beta^2 + \varepsilon^2)$, with the weight function $\phi(\varepsilon) = \frac{\beta^2}{\beta^2 + \varepsilon^2}$, where $\varepsilon$ denotes the residual for each observation and $\beta$ is the loss width. In each iteration, the normalized relative translation is the eigenvector corresponding to the smallest eigenvalue of symmetric matrix $A^TWA$, where each row of matrix $A$ corresponds to a normal vector $\bm{f}_{ki} \times \bm{f}_{kj}$, and $W$ is a diagonal weight matrix for the Cauchy loss.
Furthermore, when errors in normal vectors become significantly large due to low parallax angles, estimating relative translations or verifying feature matches based on coplanarity consistency becomes invalid. Therefore, before estimating relative translations, we initially filter feature matches with parallax angles below a predefined threshold, denoted as $T_\alpha$. 
The complete algorithm of local relative translation re-estimation is shown in Algorithm \ref{algo:re-estimation}.
\begin{algorithm} \footnotesize
	\caption{Local re-estimation for image pair $ij\in \mathcal{E}_c$}
	\label{algo:re-estimation}
	\begin{algorithmic}[1] 	
		\Require Camera rotations $\bm{R}_i$ and $\bm{R}_j$ for images $i$ and $j$, the feature matches set $M_{ij}=\{(X_{ki}, X_{kj}), \cdots \}$, the minimum parallax angle threshold $T_\alpha$, the maximum coplanarity angular residual threshold $T_\beta$, and the minimum number of inlier feature matches $N_m$.
        \Ensure Re-estimated relative translation $\bm{v}_{ij}$ and filtered feature matches $M^\prime_{ij}$.
        \State Initialize an empty set for normal vectors $N_{ij}=\emptyset$ and an empty set for inlier feature matches $M^\prime_{ij}=\emptyset$;
        \For{$(X_{ki},X_{kj})$ \textbf{in} $M_{ij}$}
            \State Calculate feature rays $\bm{f}_{ki},\bm{f}_{kj}$ with known global camera rotations estimated by Eq.~\eqref{equ:camera to point};
            \State $\bm{n}_k \leftarrow \bm{f}_{ki}\times\bm{f}_{kj}$ 
            \State $\alpha_{k} \leftarrow \arccos(\bm{f}_{ki}\cdot\bm{f}_{kj})$;
            \If{$\alpha_{k} > T_\alpha$}
                \State \textbf{Insert} $(\bm{n}_{k},X_{ki},X_{kj})$ into $N_{ij}$;
            \EndIf
        \EndFor
		\State Estimate $\bm{\tilde{v}}_{ij}$ with all normal vectors in $N_{ij}$ by Eq.~\eqref{equ:re-estimation};
        \For{$(\bm{n}_{k},X_{ki},X_{kj})$ \textbf{in} $N_{ij}$}
            \If{$\frac{\left|\bm{\tilde{v}}_{ij}^{T}\bm{n}_k\right|}{\left\|\bm{n}_k\right\|_2}<\sin(T_\beta)$}
                \State \textbf{Insert} $(X_{ki},X_{kj})$ into $M^\prime_{ij}$
            \EndIf
        \EndFor
        \State Select the direction of $\bm{\tilde{v}}_{ij}$ and filter outlier feature matches in $M^\prime_{ij}$ based on the cheirality constraints;
        \If{$|M^\prime_{ij}|\geq N_m$} 
            \State $\bm{v}_{ij} \leftarrow \bm{\tilde{v}}_{ij}$;
        \Else
            \State Discard the image pair $ij$ from the translation graph;
        \EndIf
	\end{algorithmic} 
\end{algorithm}

\begin{figure}[tp]
    \setlength{\abovecaptionskip}{0.0cm}
	\setlength{\belowcaptionskip}{-0.1cm}
    \centering
    \includegraphics[width=\linewidth, trim = 0mm 5mm 0mm 5mm, clip]{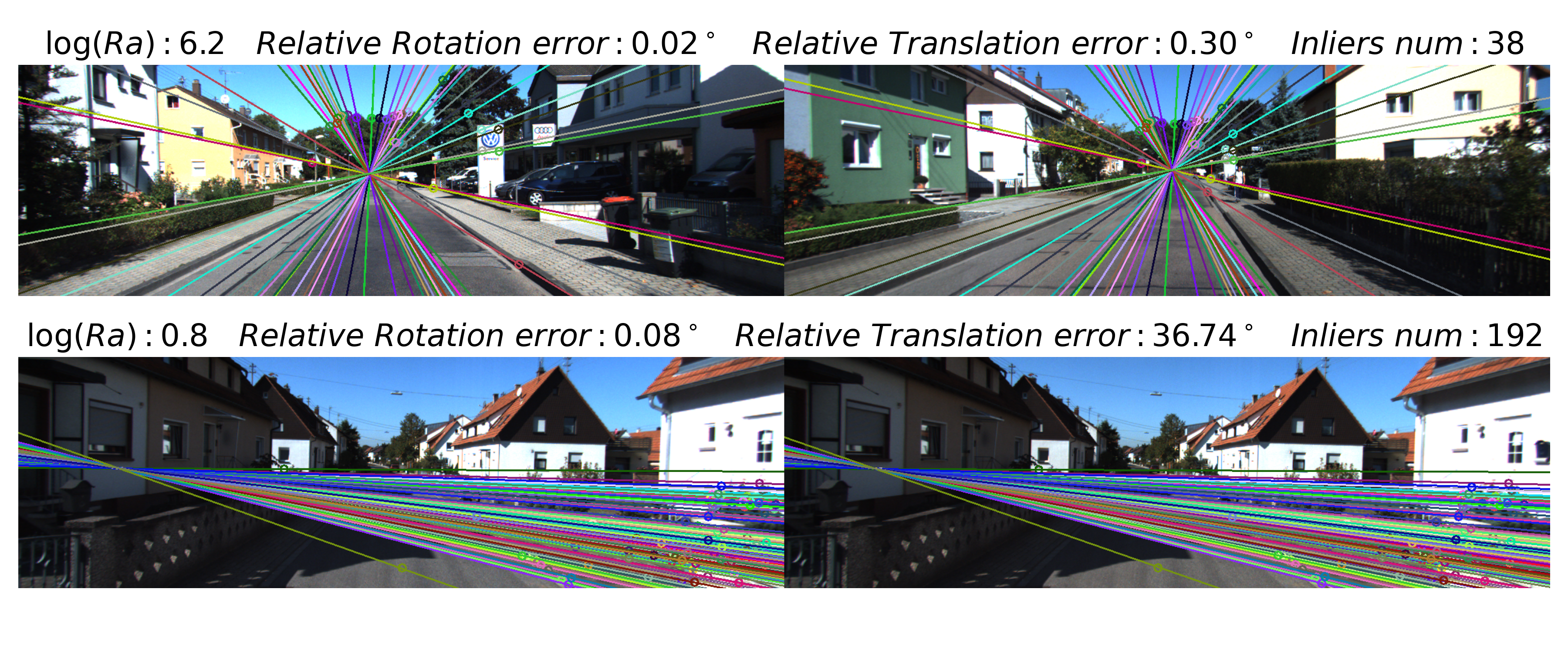}
    \caption{
    This figure shows sample epipolar lines from two overlapping image pairs. Despite similar relative rotation accuracy, the bottom pair, with a narrower epipolar-line distribution, has a larger relative translation error than the top pair, even with more inlier matches
}
    \label{fig:matches}
\end{figure}
\subsubsection{Relative Translation Confidence}
\label{sec:score}
During relative translation estimation, not only the parallax angles but also the distribution of feature matches in the image planes, which can be interpreted as the distribution of epipolar planes in 3D space, influence the accuracy.
Two classical types of epipolar plane distributions are shown in Fig.~\ref{fig:matches}. 
For convenience, we approximate the distribution of epipolar planes in 3D space by the distribution of epipolar lines in the image plane.
After filtering feature matches with parallax angles below $T_\alpha$, the top image pair has a small number of evenly distributed feature matches, while the bottom image pair has a large number of concentrated feature matches. Despite having more inliers, the bottom pair shows lower relative translation accuracy due to the narrower epipolar plane distribution.

After local re-estimation, we construct the normal matrix $\bm{A}$ using the unnormalized epipolar-plane normals $\bm{f}_{ki}\times\bm{f}_{kj}$ from the filtered feature matches in $M^\prime_{ij}$ and use the spectrum of $\bm{A}^{T}\bm{A}$ to evaluate the reliability of each relative translation.
The magnitude of each normal vector, $\left\|\bm{f}_{ki}\times\bm{f}_{kj}\right\|_2=\sin{\alpha_k}$, naturally incorporates the parallax-dependent weight of the corresponding feature match. We denote the eigenvalues of $\bm{A}^{T}\bm{A}$ in ascending order as $(e_1,e_2,1)$, where the largest eigenvalue is normalized to one. The smallest eigenvalue $e_1$ measures the residual inconsistency of the normal vectors along the estimated relative translation direction, while $e_2$ reflects the strength of the second constrained direction in the plane orthogonal to the relative translation. A larger $e_2$ therefore indicates a broader and better-conditioned directional distribution of the epipolar-plane normals.

We define the eigenvalue ratio $Ra=e_2/e_1$ as a spectral confidence measure. A larger $Ra$ indicates both weaker residual inconsistency and stronger separation of the estimated translation direction from the remaining eigenspace, and therefore generally corresponds to a more stable and identifiable relative translation estimate.


\begin{figure}[tp]
    \centering
    \includegraphics[width=\linewidth, trim = 0mm 5mm 0mm 5mm, clip]{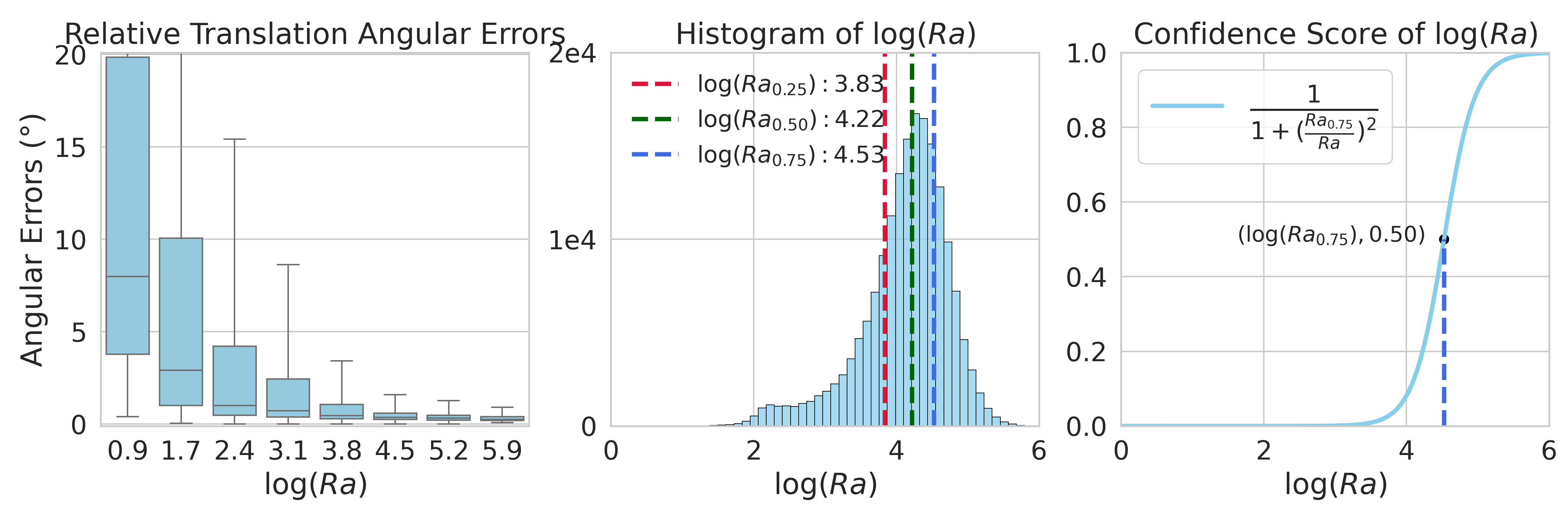}
    \caption{
    For the KITTI-05 view graph~\citep{geiger2013vision} with 177k edges, we show the relative translation error versus $\log(RA)$ (left), the distribution and quartiles of $\log(RA)$ (middle), and its mapping to confidence scores (right)
    }
    \label{fig:EigenRatio}
\end{figure}

Figure~\ref{fig:EigenRatio} illustrates the relationship between relative translation angular errors and $\log(Ra)$ on KITTI-05~\citep{geiger2013vision}, which contains 177k image pairs. Larger eigenvalue ratios generally correspond to lower angular errors, with most relative translations in the top 25\% group, $\log(Ra)>\log(Ra_{0.75})$, exhibiting errors below $5^\circ$.
\subsubsection{Global Filtering with Confidence}
\label{sec:global_filter}
To filter relative translations based on global consistency, we map the eigenvalue ratio to a confidence score $C_{ij} \in (0,1)$ for each image pair $ij \in \mathcal{E}_c$. 
Since relative translations with the top 25\% eigenvalue-ratio generally exhibit lower angular errors, we use the 75th percentile $Ra_{0.75}$ as the reference value and map it to a confidence score of 0.5:
\begin{equation}
\label{equ:conficence}
C_{ij} = \frac{1}{1+10^{2(\log(Ra_{0.75})-\log(Ra))}} = \frac{1}{1+(\frac{Ra_{0.75}}{Ra})^2}.
\end{equation}
The mapping curve of Eq.~\eqref{equ:conficence} is shown in Fig.~\ref{fig:EigenRatio} (right).
Next, we use global direction consistency to compute the coarse camera positions $\bm{\tilde{t}}_i$ for images $i\in \mathcal{V}_c$ by optimizing the angle-based objective function with the confidence score:
\begin{equation}
\label{equ:global_filter}
\min_{\substack{\bm{\tilde{t}}_i,i\in \mathcal{V}_c}}
\sum_{ij\in \mathcal{E}_c} C_{ij} \cdot \rho( \mathbb{O}(\bm{v}_{ij},\frac{\bm{\tilde{t}}_i-\bm{\tilde{t}}_j}{||\bm{\tilde{t}}_i-\bm{\tilde{t}}_j||_2})),
\end{equation}
where the robust estimator function $\rho(\cdot)$ is the Cauchy loss function and the LM \citep{levenberg1944method} solver is used as optimizer. 


Finally, based on the confidence scores and coarse camera positions, we remove globally inconsistent relative translations while preserving the connectivity of the view graph. We first extract a maximum spanning tree from the remaining view graph using the confidence score of each edge and retain all tree edges as a connectivity backbone. For each non-tree edge $ij\in\mathcal{E}_c$, we compute the temporary relative translation $\bm{\tilde{v}}_{ij}=(\bm{\tilde{t}}_i-\bm{\tilde{t}}_j)/\left\|\bm{\tilde{t}}_i-\bm{\tilde{t}}_j\right\|_2$ and discard the edge if the angular difference between $\bm{v}_{ij}$ and $\bm{\tilde{v}}_{ij}$ exceeds $10^\circ$, or if the median parallax angle of its feature matches is below $0.5^\circ$. Although a small number of noisy tree edges may remain, the confidence-based tree construction favors reliable connections. Their influence is mitigated by the robust $L_1$ initialization and further reduced by confidence weighting in the subsequent non-bilinear angle-based refinement.

\subsection{Hybrid Explicit Initialization}
\label{sec:Decouple}
Instead of directly applying bilinear angle-based optimization with random initialization as in \citet{pan2024glomap}, we first use convex distance-based objective functions to initialize the camera positions and 3D points. For efficiency, we decouple the initialization into two stages: first, we estimate the initial camera positions using pure relative translations, and then triangulate the 3D points and refine the camera positions through alternating triangulation. Finally, we employ a non-bilinear angle-based objective function with hybrid camera-to-camera and camera-to-point constraints to efficiently refine both the camera positions and 3D points.


\subsubsection{Camera Position Initialization}
With refined relative translations, we employ the cross-product form objective function under the $L_1$ norm, using the ADMM optimizer~\citep{boyd2011distributed,lorenz2019non} to robustly estimate the initial camera positions:
\begin{align}
\label{eq:PTA}
\min_{\substack{\bm{t}_i,i\in \mathcal{V}_c}}
\sum_{ij\in \mathcal{E}_c}||&\bm{v}_{ij}\times(\bm{t}_{i}-\bm{t}_{j})||_1, 
 \\
\text{s.t.} \sum_{i\in \mathcal{V}_c} \bm{t}_{i} = 0,\quad &\bm{v}_{ij}\cdot (\bm{t}_i-\bm{t}_j) \geq 1, \quad \forall ij\in \mathcal{E}_c,\notag
\end{align}
where the equality constraint $\sum_{i\in \mathcal{V}_c} \bm{t}_{i} = 0$ is applied to remove inherent positional ambiguity, and the inequality constraints $\bm{v}_{ij}\cdot (\bm{t}_i-\bm{t}_j) \geq 1, \forall ij\in \mathcal{E}_c$ are used to resolve scale ambiguity and restrict the direction for each term.
This camera-only stage is intended to provide a non-collapsed and directionally consistent initialization rather than an accurate final solution. Although directional camera graphs may become weakly constrained under degenerate motion configurations, the subsequent alternating triangulation and hybrid angle-based refinement introduce additional camera-to-point constraints from feature tracks, which substantially improve the robustness of the initialization in practical view-track graphs.

\subsubsection{Alternating Triangulation}
After performing translation averaging, we first triangulate the 3D points using fixed camera positions. 
Since the camera positions are fixed, there is no positional or scale ambiguity. The unconstrained convex cross-product form objective function is formulated as follows:
\begin{equation}
\label{eq:Triangulation}
\min_{\substack{\bm{P}_k, k\in \mathcal{V}_p}}
\sum_{ki\in \mathcal{E}_p}||\bm{f}_{ki}\times(\bm{P}_{k}-\bm{t}_{i})||_1.
\end{equation}
The $L_1$ norm is used here to maintain robustness against outlier feature points in feature tracks.

However, since camera-to-point constraints are ignored during translation averaging, cameras with insufficient camera-to-camera constraints may exhibit low accuracy.
To ensure a more consistent topological structure between cameras and 3D points for subsequent angle-based refinement, camera positions are re-estimated by inverse triangulating the cameras with fixed 3D points: 
\begin{equation}
\label{eq:Inv_Triangulation}
\min_{\substack{\bm{t}_i,i\in \mathcal{V}_c}}
\sum_{ki\in \mathcal{E}_p}||\bm{f}_{ki}\times(\bm{P}_{k}-\bm{t}_{i})||_1.
\end{equation}
Finally, the 3D points are re-triangulated using the re-estimated camera positions with the objective function in Eq.~\eqref{eq:Triangulation}. 
With fixed camera positions, different 3D points in Eq.~\eqref{eq:Triangulation} can be optimized independently and in parallel. Likewise, with fixed 3D points, different cameras in Eq.~\eqref{eq:Inv_Triangulation} can be updated independently. The ADMM optimizer~\citep{boyd2011distributed} is used to solve these least absolute deviation subproblems.

\subsubsection{Non-bilinear Angle-based Refinement}
After the convex initialization, we first filter the feature points for each image whose corresponding 3D points lie behind the image planes.
We then utilize the non-bilinear angle-based objective function to refine the positions of cameras and 3D points.
To enhance robustness, rather than relying solely on the camera-to-point constraints from feature tracks, as in \citep{liu2019robust, cai2021pose, pan2024glomap}, we also incorporate relative translations weighted by their confidence scores to directly constrain the relationship between cameras.
Let $\hat{\bm{t}}_{ij}=(\bm{t}_i-\bm{t}_j)/\left\|\bm{t}_i-\bm{t}_j\right\|_2$ and $\hat{\bm{f}}_{ki}=(\bm{P}_k-\bm{t}_i)/\left\|\bm{P}_k-\bm{t}_i\right\|_2$. Based on Eq.~\eqref{eq:non-bilinear}, our objective function is formulated as
\begin{equation}
\label{eq:optim_formu_l2}
\min_{\substack{\bm{t}_i,\bm{P}_k\\ i\in\mathcal{V}_c,\;k\in\mathcal{V}_p}}
\sum_{ki\in\mathcal{E}_p}\rho\left(\mathbb{O}(\bm{f}_{ki},\hat{\bm{f}}_{ki})\right)
+
\sum_{ij\in\mathcal{E}_c}C_{ij}\rho\left(\mathbb{O}(\bm{v}_{ij},\hat{\bm{t}}_{ij})\right),
\end{equation}
where $\rho(\cdot)$ is the Cauchy loss function and the LM method~\citep{levenberg1944method} is used for optimization.

\subsection{Joint Optimization}
\label{sec:joint}
Given the initialization of camera poses and 3D points, we perform joint optimization to refine both camera parameters and 3D points. However, reprojection-based bundle adjustment is unbounded and highly non-convex, making the process sensitive to outlier feature tracks and inaccurate camera parameter initialization. Existing methods \citep{pan2024glomap,theia-manual} first filter out feature points with initial camera poses based on angular errors to enhance robustness before bundle adjustment. However, when the accuracy of the initial camera rotations from rotation averaging is insufficient, the fixed camera rotations during initialization constrain the accuracy of the estimated camera positions, leading to the erroneous filtering of many inlier feature tracks.
Consequently, directly filtering feature tracks and performing bundle adjustment is prone to local optima.
Furthermore, because the number of feature points varies significantly across images, optimization tends to prioritize images with more feature points, resulting in uneven error distribution and local optima. 
To reduce dependence on the accuracy of initial camera rotations and obtain more accurate camera poses and feature tracks for final bundle adjustment, we first select a spatially balanced subset of reliable feature tracks for each image and then improve camera poses using selected feature tracks through angle-based refinement and subsequent bundle adjustment. Our joint optimization consists of three steps: \begin{enumerate}[label=\alph*)] \item \textbf{Selective angle-based refinement}: We first formulate an angle-based objective with these selected feature tracks to robustly refine camera poses. \item \textbf{Selective bundle adjustment}: Using the selected feature tracks, camera poses, camera intrinsics, and 3D points are further refined simultaneously via bundle adjustment. \item \textbf{Complete bundle adjustment}: With improved camera poses, we re-filter the feature tracks and use bundle adjustment to refine camera parameters and 3D points using all reliable feature tracks. \end{enumerate} Next, we describe each step in detail.

\subsubsection{Coverage-Balanced Feature Track Selection}
\label{sec:feature tracks}
We select a subset of reliable feature tracks to improve efficiency and balance camera and image-plane coverage during camera pose refinement. Since longer tracks generally provide stronger camera-to-point constraints, we first sort all tracks in descending order of length and iteratively select a track if it contains an image whose current coverage is below $S_1$. This process continues until each image is covered by at least $S_1$ selected tracks.

Since long tracks are often concentrated in highly textured regions, we further divide each image into square grids with side length $\sqrt{\frac{H\times W}{S_2}}$, where $H$ and $W$ are the image height and width, respectively, and $S_2$ is the expected number of grids. For each uncovered grid, we select the feature point belonging to the longest track. In this paper, we set $S_1=100$ and $S_2=50$, and denote the selected feature tracks by $\{\mathcal{V}_p^s,\mathcal{E}_p^s\}$.

\subsubsection{Selective Angle-based Refinement} 
Given the initialization of camera poses and 3D points, where camera rotations are obtained through rotation averaging, a small angle threshold $T_\gamma$ is typically used to filter out outlier feature points in feature tracks based on the angle between the observed feature ray and the direction from the camera to the 3D point. However, when the accuracy of global camera rotations is insufficient, directly applying a small threshold to filter feature points may discard many inlier feature points, potentially causing subsequent bundle adjustment to converge to incorrect local optima. To address this issue, we first apply a relatively larger threshold for feature point filtering.
We then jointly refine camera rotations, positions, and 3D points using a bounded angle-based objective function. Unlike the unbounded reprojection error, the angular residual has a fixed upper bound and therefore limits the influence of feature points with large geometric errors, providing a robust initialization for subsequent bundle adjustment.
Following the chordal angular residual used in 1DSfM~\citep{wilson2014robust}, the objective function is formulated as follows:
\begin{equation}
\label{eq:angle_ba}
\min_{\substack{\bm{t}_i, \bm{R}_i ,\bm{P}_k\\ k\in \mathcal{V}_p^s, i\in \mathcal{V}_c;}}
\sum_{ki\in \mathcal{E}_p^s}\rho(||\bm{R}_{i}^T\bm{\hat{X}}_{ki} - \frac{\bm{P}_k-\bm{t}_i}{||\bm{P}_k-\bm{t}_i||_2}||_2),
\end{equation}
where the robustifier $\rho(\cdot)$ is the Huber loss, which further reduces the influence of the remaining outlier feature points, and the LM method~\citep{levenberg1944method} is used for optimization.
Similar to \citet{pan2024glomap,moulon2017openmvg}, camera rotations are first fixed and then jointly optimized with camera positions and 3D points, which improves the accuracy of camera poses. We infer that with coarse camera rotations, relatively accurate camera positions can be estimated through nearby 3D points, and then, more accurate camera positions enhance the optimization of camera rotations using distant 3D points.

\subsubsection{Selective Bundle Adjustment} 
Since the angle error metric does not account for the spatial distribution of feature points in the image plane, we employ bundle adjustment to further refine the camera intrinsics and poses using the selected feature tracks.
The objective function is formulated as follows:
\begin{equation}
\label{eq:sba}
\min_{\substack{\bm{t}_i,\bm{R}_i ,\bm{\pi}_i ,\bm{P}_k\\ k\in \mathcal{V}_p^s, i\in \mathcal{V}_c;}}
\sum_{ki\in \mathcal{E}_p^s}\rho(||\pi_i(\bm{R}_i(\bm{P}_k-\bm{t}_i))-\bm{x}_{ki}||_2),
\end{equation}
where the robust estimator $\rho(\cdot)$ is the Huber loss function
and LM \citep{levenberg1944method} method is used as the optimizer.
The camera rotations are also first fixed and then jointly optimized with camera intrinsics and 3D points to improve convergence.

\subsubsection{Complete Bundle Adjustment} 
With improved camera poses, we utilize a RANSAC-based scheme \citep{schoenberger2016sfm} to filter outlier feature points in feature tracks and triangulate all 3D points.
Finally, camera parameters and 3D points are refined through a complete bundle adjustment with all reliable camera-to-point constraints.
The reprojection-based objective function is formulated as follows:
\begin{equation}
\label{eq:cba}
\min_{\substack{\bm{t}_i,\bm{R}_i ,\bm{\pi}_i ,\bm{P}_k\\ k\in \mathcal{V}_p, i\in \mathcal{V}_c;}}
\sum_{ki\in \mathcal{E}_p}\rho(||\pi_i\left(\bm{R}_i(\bm{P}_k-\bm{t}_i)\right)-\bm{x}_{ki}||_2),
\end{equation}
where the robust estimator and the optimization method are the same as those used in Eq.~\eqref{eq:sba}. 

\begin{table*}
    \caption{Camera-pose AUC on the 1DSfM \citep{wilson2014robust} dataset. $N_t$ and $N_c$ denote the number of input and registered images, respectively.
    For global methods, the best results are shown in bold and the second-best results are underlined.
    CReTA~\citep{manam2022correspondence} fails in scene ROF.
    } 	
	\setlength{\tabcolsep}{1.0pt}
	\centering
    \scalebox{0.71}{
        \begin{tabular}{|c|c||ccc|c||ccc|c||ccc|c||ccc|c||ccc|c||ccc|c|}
            \hline		
            \multicolumn{2}{|c||}{method} &			
            \multicolumn{4}{c||}{COLMAP} &
            \multicolumn{4}{c||}{CReTA} &
            \multicolumn{4}{c||}{LiGT} &
            \multicolumn{4}{c||}{HETA} &
            \multicolumn{4}{c||}{GLOMAP} &
            \multicolumn{4}{c|}{HETA++}\\			
            \hline
            \multirow{2}{*}{Data} &  \multirow{2}{*}{$N_t$} 
            &\multicolumn{3}{c|}{AUC@} &\multirow{2}{*}{$N_c\uparrow$}  
            &\multicolumn{3}{c|}{AUC@}  &\multirow{2}{*}{$N_c\uparrow$}  
            &\multicolumn{3}{c|}{AUC@}  &\multirow{2}{*}{$N_c\uparrow$}  
            &\multicolumn{3}{c|}{AUC@}  &\multirow{2}{*}{$N_c\uparrow$}  
            &\multicolumn{3}{c|}{AUC@}  &\multirow{2}{*}{$N_c\uparrow$}  
            &\multicolumn{3}{c|}{AUC@}  &\multirow{2}{*}{$N_c\uparrow$}  \\
            & &$3^\circ\uparrow$  & $5^\circ$ &  $10^\circ$
            & &$3^\circ\uparrow$  & $5^\circ\uparrow$ &  $10^\circ\uparrow$
            & &$3^\circ\uparrow$  & $5^\circ\uparrow$ &  $10^\circ\uparrow$
            & &$3^\circ\uparrow$  & $5^\circ\uparrow$ &  $10^\circ\uparrow$
            & &$3^\circ\uparrow$  & $5^\circ\uparrow$ &  $10^\circ\uparrow$
            & &$3^\circ\uparrow$  & $5^\circ\uparrow$ &  $10^\circ\uparrow$
            &\\
            \hline
            {ALM} & 577 
            & 44.3 & 58.0 & 73.1 & 501
            & 39.3 & 52.2 & 66.9 & \textbf{567} 
            & 21.8 & 32.7 & 48.1 & 547 
            & 39.7 & {52.6} & {67.3} & \textbf{567} 
            & \textbf{39.9} & \underline{52.7} & \underline{67.4} & \textbf{567} 
            & \underline{39.8} & \textbf{52.8} & \textbf{67.6} & \textbf{567}  \\ 
            {ELS} & 227 
            & 48.6 & 62.7 & 77.1 & 217
            & 45.2 & 58.9 & \underline{73.2} & \textbf{226} 
            & 44.0 & 57.2 & 71.7 & 215 
            & \underline{45.6} & \underline{59.3} & 73.0 & \textbf{226}  
            & 45.2 & 58.8 & 72.5 & \textbf{226}
            & \textbf{45.8} & \textbf{59.8} & \textbf{73.7} & \textbf{226}\\
            {GDM} & 677 
            & 17.7 & 25.7 & 35.2 & 590 
            & 15.5 & 22.9 & 32.2 & \textbf{673} 
            & 5.0 & 7.2 & 10.8 & 645  
            & \underline{15.8} & \underline{23.4} & \underline{32.9} & 671  
            & 15.6 & 22.9 & 32.3 & \textbf{673}
            & \textbf{15.9} & \textbf{23.5} & \textbf{33.0} & \textbf{673}\\
            {MDR} & 341 
            & 29.7 & 44.3 & 62.1 & 178
            & 26.9 & 39.1 & 54.5 & \textbf{340} 
            & 21.6 & 31.7 & 45.4 & 316  
            & \textbf{27.5} & \textbf{40.3} & \textbf{56.4} & \textbf{340}  
            & 26.8 & 39.1 & 54.8 & \textbf{340}
            & \underline{27.3} & \underline{39.8} & \underline{55.7} & \textbf{340} \\
            {MND} & 450  
            & 52.4 & 64.4 & 76.9 & 403 
            & 46.9 & 58.6 & 71.6 & \textbf{449}
            & 41.1 & 51.1 & 62.3 & 438  
            & \underline{47.4} & \underline{59.3} & \underline{72.1} & 448  
            & 47.2 & 59.1 & 71.9 & \textbf{449}
            & \textbf{47.6} & \textbf{59.5} & \textbf{72.5} & \textbf{449}\\
            {ND} & 553 
            & 50.2 & 62.8 & 75.6 & 479 
            & 47.2 & 59.5 & 72.4 & \textbf{552} 
            & 30.1 & 42.0 & 56.0 & 535  
            & 47.1 & 59.5 & 72.5 & \textbf{552}  
            & \underline{47.4} & \underline{59.9} & \underline{72.9} & \textbf{552}
            & \textbf{47.8} & \textbf{60.2} & \textbf{73.0} & {\textbf{552}}\\
            {NYC} & 332 
            & 48.2 & 60.1 & 73.9 & 296 
            & 39.8 & 52.0 & 66.9 & \textbf{329} 
            & 0.1 & 0.2 & 1.0 & {316}  
            & \underline{43.4} & 54.3 & 67.4 & \textbf{329}  
            & 43.3 & \underline{54.4} & \underline{67.8} & \textbf{329}
            & \textbf{43.8} & \textbf{55.0} & \textbf{68.2} & {\textbf{329}}\\
            {PDP} & 338 
            & 48.4 & 62.7 & 76.9 & 295
            & 44.2 & 57.4 & 71.3 & \textbf{337} 
            & 34.5 & 46.6 & 59.9 & {312}  
            & \underline{44.6} & \underline{58.2} & \underline{72.6} & 336  
            & \underline{44.6} & 58.1 & 72.4 & \textbf{337}
            & \textbf{45.1} & \textbf{58.4} & \textbf{72.8} & {\textbf{337}}\\
            {PIC} & 2152
            & 52.5 & 64.2 & 76.7 & 1838 
            & \underline{44.5} & 56.0 & 69.1 & \textbf{2146} 
            & 3.8 & 9.4 & 22.0 & {216}  
            & {44.2} & \underline{56.3} & \underline{69.8} & 2145  
            & {44.2} & 56.2 & 69.6 & \textbf{2146}
            & \textbf{46.3} & \textbf{58.1} & \textbf{71.3} & {\textbf{2146}}\\
            {ROF} & 1084
            & 73.8 & 81.4 & 88.5 & 918 
            & - & - & - & - 
            & 14.3 & 23.0 & 38.9 & 1043  
            & 49.5 & 61.8 & 75.5 & \textbf{1082}  
            & \underline{67.8} & \underline{75.9} & \underline{84.1} & \textbf{1082}
            & \textbf{68.4} & \textbf{76.2} & \textbf{84.4} & {\textbf{1082}}\\
            {TFG} & 5058
            & 44.9 & 57.9 & 72.3 & 3989 
            &  36.6& 49.3 & 64.4 & \textbf{5050}
            & 0.8 & 2.8 & 9.4 & 4715   
            & 30.8 & 42.1 & 59.9 & 5048
            & \underline{38.2} & \underline{50.4} & \underline{65.2} & \textbf{5050}
            & \textbf{39.1} & \textbf{51.4} & \textbf{66.2} & \textbf{5050}\\
            {TOL} & 472
            & 56.0 & 68.9 & 81.1 & 396 
            & \underline{46.0} & \underline{58.4} & \underline{71.2} & \textbf{472} 
            & 0.4 & 1.2 & 4.3 & 444  
            & 45.2 & {57.7} & \underline{71.2} & \textbf{472}  
            & 45.5 & 57.7 & 70.9 & \textbf{472}
            & \textbf{47.5} & \textbf{60.3} & \textbf{73.2} & {\textbf{472}}\\
            {USQ} & 789
            & 19.3 & 31.6 & 48.6 & 637
            & 13.7 & 24.2 & 38.7 & \textbf{784} 
            & 9.4 & 15.6 & 25.1 & 681  
            & 14.3 & \underline{25.3} & \underline{40.1} & \textbf{784}  
            & \textbf{14.4} & 25.2 & 40.0 & \textbf{784}
            & \textbf{14.6} & \textbf{25.6} & \textbf{40.5} & {\textbf{784}}\\
            {VNC} & 836
            & 40.7 & 55.7 & 71.4 & 713 
            & 35.3 & 49.5 & 64.6 & \textbf{834} 
            & \textbf{37.7} & \textbf{51.5} & \textbf{66.6} & 754  
            & \underline{35.4} & 49.7 & 65.1 & \textbf{834}  
            & 35.1 & {49.9} & {65.9} & {\textbf{834}}
            & \underline{35.4} & \underline{50.3} & \underline{66.5} & {\textbf{834}}\\
            {YKM} & 437
            & 56.5 & 69.5 & 80.9 & 337
            & 43.9 & 55.6 & 66.7 & \textbf{434} 
            & 38.8 & 48.7 & 58.5 & 401
            & \textbf{45.5} & \textbf{57.4} & \textbf{69.1} & \textbf{434}  
            & 43.8 & 55.9 & 67.6 & \textbf{434}
            & \underline{45.1} & \underline{57.1} & \underline{68.5} & {\textbf{434}}\\
            \hline
            {Average} & 954.9
            & 45.5 & 58.0 & 71.4 & 785.8
            & 37.5 & 49.5 & 63.1 & 942.4 
            & 20.2 & 28.1 & 38.7 & 771.9
            & 38.4 & 50.5 & 64.3 & 951.2  
            & \underline{39.9} & \underline{51.7} & \underline{65.0} & \textbf{951.7}
            & \textbf{40.6} & \textbf{52.5} & \textbf{65.8} &\textbf{951.7} \\
            \hline
        \end{tabular}
    }
	
	\label{tab:1DSFM}
\end{table*}

\section{Experiments}
\label{sec:experiment}
The experiment is performed on an Ubuntu 20.04.5 LTS platform, with 64 GB of memory and a 12th Gen Intel(R) Core(TM) i7-12700 CPU @ 2.10 GHz, with 20 cores. 
To demonstrate the performance of our pipeline, called HETA++, we conduct extensive experiments on various sequential and unordered datasets.
We also present ablation experiments to study the performance of three key components in our system.

\noindent\textbf{Baselines.}
To demonstrate the superiority of our method, we compare HETA++ with the robust incremental method COLMAP~\citep{schoenberger2016sfm} and several state-of-the-art global methods. These include the camera-only translation averaging method CReTA~\citep{manam2022correspondence}, the implicit method LiGT~\citep{cai2021pose}, the explicit method GLOMAP~\citep{pan2024glomap}, and our previous work HETA~\citep{tao2024revisiting}. All geometry-based methods use the same view-track graph as input. The global methods share the same image matches, relative poses, and global rotations. Image matches and relative poses are computed using COLMAP~\citep{schoenberger2016sfm} and refined with PoseLib~\citep{PoseLib}, while global rotations are estimated using the method of \citet{chatterjee2017robust}.
Since CReTA-BATA~\citep{manam2022correspondence} extends BATA with correspondence-based relative translation refinement and generally outperforms CReTA-RLUD, we use it as the representative camera-only baseline. As CReTA, LiGT, and HETA focus on translation averaging, their 3D points are triangulated using RANSAC before jointly refining camera parameters and scene structure through bundle adjustment. For fairness, we apply the GLOMAP bundle adjustment module~\citep{pan2024glomap} to all three methods, using three rounds by default, followed by the structure refinement of \citet{pan2024glomap}.
We additionally include the state of the art feed-forward method $\pi^3$+BA~\citep{wang2025pi}, which directly predicts camera poses and scene geometry before bundle adjustment. 


\noindent\textbf{Datasets.}
For sequential evaluation, we use KITTI~\citep{geiger2013vision}, ETH3D MVS (rig)~\citep{ETH3DDataset}, and LaMAR~\citep{sarlin2022lamar}. KITTI is captured by two cameras mounted on a driving vehicle, with predominantly near-collinear camera trajectories. Both cameras are used but treated independently for all compared methods. ETH3D MVS (rig) contains indoor and outdoor scenes with approximately 1,000 frames captured by a multi-camera rig; the rig pose is not fixed for any method. LaMAR is a large-scale indoor–outdoor benchmark containing tens of thousands of images captured by various AR devices and smartphones.

For unordered evaluation, we use 1DSfM~\citep{wilson2014robust} and ETH3D MVS (DSLR)~\citep{ETH3DDataset}. The 1DSfM dataset contains medium- to large-scale Internet image collections captured by diverse cameras. Following the standard protocol, we use the largest connected component of the released view graph and the provided Bundler reconstruction as reference for direct comparison with previous translation averaging methods. ETH3D MVS (DSLR) contains several small-scale indoor and outdoor scenes with high-resolution images.
Image matches for LaMAR are computed using its recommended pipeline~\citep{sarlin2022lamar}, while those for the other datasets are obtained using COLMAP~\citep{schoenberger2016sfm}.

\noindent\textbf{Metrics.} 
For most datasets, we report the Area Under the Recall Curve (AUC) scores, which are computed based on the maximum relative rotation and translation errors for all image pairs, similar to the metrics used by \citet{pan2024glomap,he2024dfsfm,lindenberger2021pixsfm}.
For large-scale sequential datasets like KITTI~\citep{geiger2013vision} and LaMAR~\citep{sarlin2022lamar}, where most camera trajectories are nearly collinear, the relative error is ineffective in capturing scale drift.
For the KITTI dataset, we report the median and mean distance errors after globally aligning the reconstruction to the ground truth using a robust RANSAC-based~\citep{chum2003locally} similarity transform~\citep{Umeyama}, following the evaluation protocols of \citet{zhuang2018baseline,wilson2014robust,ozyesil2015robust,manam2022correspondence}.
For the LaMAR dataset, cameras with sparse feature tracks are often optimized far from the scene, leading to significant mean position errors across many methods. Therefore, we compare both the median position errors and the AUC scores.

\begin{scaledtable}[t]
\caption{
Reconstruction runtimes in seconds on the 1DSfM
dataset~\protect\citep{wilson2014robust}, excluding the common
view-track graph construction stage. The best and second-best results
are shown in bold and underlined.
}
\label{tab:1DSFM_time}

\setlength{\tabcolsep}{1.5pt}
\renewcommand{\arraystretch}{0.95}

\resizebox{\columnwidth}{!}{%
\begin{tabular}{@{}lrrrrrr@{}}
\toprule
Method & COLMAP & CReTA & LiGT & HETA & GLOMAP & HETA++ \\
\midrule
ALM & 6028  & 1271   & 2400  & \underline{1229} & 1606 & \textbf{953} \\
ELS & 566   & 121    & 113   & \textbf{97} & 184 & \underline{102} \\
GDM & 4952  & \textbf{478} & 560 & 660 & 816 & \underline{550} \\
MDR & 1563  & \underline{252} & 282 & 309 & 307 & \textbf{189} \\
MND & 3693  & \underline{356} & 868 & 417 & 455 & \textbf{299} \\
ND  & 9890  & 1269 & \underline{1174} & 1338 & 1199 & \textbf{894} \\
NYC & 1285  & \underline{82} & \textbf{42} & 283 & 325 & 229 \\
PDP & 1353  & 152 & \textbf{148} & \underline{192} & 256 & 231 \\
PIC & 25319 & 42868 & \textbf{6732} & 32095 & 28951 & \underline{16020} \\
ROF & 11054 & -- & \textbf{580} & \underline{1010} & 3791 & 2990 \\
TFG & 74544 & 164906 & \textbf{51516} & 138640 & 203537 & \underline{64656} \\
TOL & 3695  & 739 & \textbf{59} & 693 & 671 & \underline{320} \\
USQ & 3170  & 865 & 885 & \underline{669} & 894 & \textbf{466} \\
VNC & 17735 & 1672 & 1968 & 1992 & \underline{1587} & \textbf{1446} \\
YKM & 3748  & 463 & \underline{421} & 595 & 504 & \textbf{332} \\
\bottomrule
\end{tabular}%
}
\end{scaledtable}

\begin{table*}
    \caption{Camera-pose AUC on the ETH3D MVS (DSLR) \citep{ETH3DDataset} dataset.  $N_t$ and $N_c$ denote the number of input and registered images, respectively. The best results are shown in bold. For the average accuracy, the second-best results are underlined.}
	\setlength{\tabcolsep}{1.0pt}
	\centering
    \scalebox{0.63}{
        \begin{tabular}{|c|c||ccc|c||ccc|c||ccc|c||ccc|c||ccc|c||ccc|c||cc|c|}
            \hline		
            \multicolumn{2}{|c||}{method} &			
            \multicolumn{4}{c||}{COLMAP} &
            \multicolumn{4}{c||}{CReTA} &
            \multicolumn{4}{c||}{LiGT} &
            \multicolumn{4}{c||}{HETA} &
            \multicolumn{4}{c||}{GLOMAP} &
            \multicolumn{4}{c||}{HETA++} &
            \multicolumn{3}{c|}{$\pi^3$+BA}\\			
            \hline
            \multirow{2}{*}{Data} &  \multirow{2}{*}{$N_t$} 
            &\multicolumn{3}{c|}{AUC@} &\multirow{2}{*}{$N_c\uparrow$}  
            &\multicolumn{3}{c|}{AUC@}  &\multirow{2}{*}{$N_c\uparrow$}  
            &\multicolumn{3}{c|}{AUC@}  &\multirow{2}{*}{$N_c\uparrow$}  
            &\multicolumn{3}{c|}{AUC@}  &\multirow{2}{*}{$N_c\uparrow$}  
            &\multicolumn{3}{c|}{AUC@}  &\multirow{2}{*}{$N_c\uparrow$}  
            &\multicolumn{3}{c|}{AUC@}  &\multirow{2}{*}{$N_c\uparrow$}
            &\multicolumn{2}{c|}{AUC@} &\multirow{2}{*}{$N_c\uparrow$}  \\
            & &$3^\circ\uparrow$  & $5^\circ$ &  $10^\circ$
            & &$3^\circ\uparrow$  & $5^\circ\uparrow$ &  $10^\circ\uparrow$
            & &$3^\circ\uparrow$  & $5^\circ\uparrow$ &  $10^\circ\uparrow$
            & &$3^\circ\uparrow$  & $5^\circ\uparrow$ &  $10^\circ\uparrow$
            & &$3^\circ\uparrow$  & $5^\circ\uparrow$ &  $10^\circ\uparrow$
            & &$3^\circ\uparrow$  & $5^\circ\uparrow$ &  $10^\circ\uparrow$
            & &$3^\circ\uparrow$ & $5^\circ\uparrow$ &\\
            \hline
            {botanical} & 30
            & \textbf{88.4} & \textbf{96.1} & \textbf{97.7} & 17 
            &42.7  &50.1  &56.5  &\textbf{30} 
            &32.1  &38.3  &40.2  &\textbf{30} 
            &84.5  &94.7  &96.8  &\textbf{30}
            &84.5  &94.7  &96.8  &\textbf{30}  
            &84.5  &94.7  &96.8  &\textbf{30}   & 6.6 & 14.8 & \textbf{30} \\ 
            {boulders} & 26
            & 89.9 & 96.6 & \textbf{98.0} & \textbf{26} 
            &89.9  & 96.6 &\textbf{98.0}  & \textbf{26} 
            & \textbf{90.0} &\textbf{96.7}  &\textbf{98.0} & \textbf{26} 
            &89.9  & 96.6 &\textbf{98.0}  & \textbf{26}
            & 89.9 & 96.6 & \textbf{98.0} &  \textbf{26}
            &89.9  &96.6  &\textbf{98.0}  & \textbf{26}  & 70.1 & 78.9 & \textbf{26} \\
            {bridge} & 110
            & 91.5 & 97.2 & 98.3 & \textbf{110} 
            &91.2  &96.9  &98.1  & \textbf{110} 
            &91.7  &97.2  &98.3  & \textbf{110} 
            &91.7  & 97.2 & 98.3 & \textbf{110} 
            &91.7  &97.2  &98.3  & \textbf{110} 
            &91.7  &97.2  &98.3  & \textbf{110}  & 11.3 & 25.4 & \textbf{110} \\
            {courtyard} & 38
            & 88.3 & 96.1 & 97.7 & \textbf{38} 
            & \textbf{87.7} & \textbf{95.9} &\textbf{97.5} & \textbf{38}
            & 87.1 & 95.7 & 97.4 & \textbf{38} 
            & 87.0 & 95.7 & 97.4 & \textbf{38}
            & 87.0 & 95.7 & 97.4 & \textbf{38} 
            & 87.0 & 95.7 &97.4  &  \textbf{38} & 73.6 & 83.7 & \textbf{38} \\
            {delivery} & 44
            & 92.3 & 97.4 & 98.5 &  \textbf{44}
            &\textbf{92.5}  &\textbf{97.5}  &\textbf{98.5} & \textbf{44}
            &\textbf{92.5}  &\textbf{97.5}  &\textbf{98.5} & \textbf{44} 
            &\textbf{92.5}  &\textbf{97.5}  &\textbf{98.5} & \textbf{44}
            &\textbf{92.5}  &\textbf{97.5}  &\textbf{98.5} & \textbf{44}
            &\textbf{92.5}  &\textbf{97.5}  &\textbf{98.5} & \textbf{44} & 34.4 & 47.5 & \textbf{44} \\
            {door} & 7
            & \textbf{92.4} & \textbf{97.5} & \textbf{98.5} & \textbf{7} 
            & 91.4 & 97.1 & 98.3 & \textbf{7} 
            & 91.4 & 97.1 & 98.3 & \textbf{7} 
            & 91.4 & 97.1 & 98.3 & \textbf{7}
            & 91.4 & 97.1 & 98.3 & \textbf{7} 
            & 91.4 & 97.1 & 98.3 & \textbf{7} & 78.7 & 85.6 & \textbf{7} \\
            {electro} & 45
            & \textbf{88.5} & \textbf{96.0} & \textbf{97.5} &  41
            & 80.1 & 86.8 & 88.8 & \textbf{45} 
            & {86.9} & {95.0} & {96.9} & 42 
            & 83.9 & 92.8 & 95.4 & \textbf{45}
            & 83.1 & 90.9 & 92.6 & \textbf{45} 
            & 83.9 & 92.4 & 95.1 & \textbf{45} & 44.2 & 60.3 & \textbf{45} \\
            {exhibition} & 68
            & 84.9 & 94.3 & 96.4 & \textbf{68} 
            &  \textbf{85.1}& \textbf{94.2} &  \textbf{96.5}& \textbf{68} 
            &  \textbf{85.1}& \textbf{94.2} &  \textbf{96.5}& \textbf{68}  
            &  \textbf{85.1}& \textbf{94.2} &  \textbf{96.5}& \textbf{68} 
            &  \textbf{85.1}& \textbf{94.2} &  \textbf{96.5}& \textbf{68} 
            &  \textbf{85.1}& \textbf{94.2} &  \textbf{96.5}& \textbf{68}  & 1.9 & 2.6 & \textbf{68} \\
            {facade} & 76
            &90.1  & 96.7 & 98.0 & \textbf{76} 
            & \textbf{90.4} & \textbf{96.8}&  \textbf{98.1}& \textbf{76}  
            & 66.2 & 70.5 & 71.5 & \textbf{76} 
            & \textbf{90.4} & \textbf{96.8}&  \textbf{98.1}& \textbf{76}
            & \textbf{90.4} & \textbf{96.8}&  \textbf{98.1}& \textbf{76}
            & \textbf{90.4} & \textbf{96.8}&  \textbf{98.1}& \textbf{76} & 53.7 & 67.4 & \textbf{76} \\
            {kicker} & 31
            & 91.8 & 97.3 & 98.4 & 30 
            & 85.8 & 91.1 & 92.2 &  \textbf{31}
            & \textbf{91.9} & \textbf{97.3} &  \textbf{98.4}& 30 
            & 85.8 & 90.9 & 93.8 & \textbf{31}
            & 85.8 & 90.9 & 93.8 &  \textbf{31}
            & 85.8 & 90.9 & 93.8  &\textbf{31} & 85.2 & 90.9 & \textbf{31} \\
            {lecture} & 23
            & \textbf{91.5} & 93.6 & 96.1 &  \textbf{23}
            & 82.1 & 93.8 & \textbf{96.3} & \textbf{23}            
            & 82.3 & \textbf{93.9} & \textbf{96.3} & \textbf{23}   
            & 82.2 & 93.8 & \textbf{96.3} & \textbf{23} 
            & 82.2 & 93.8 & \textbf{96.3} & \textbf{23}
            & 82.2 & 93.8 & \textbf{96.3} & \textbf{23} & 76.3 & 84.2 & \textbf{23} \\
            {living} & 65
            & \textbf{87.7} & \textbf{95.7} & \textbf{97.4} & \textbf{65} 
            & 87.5 & 95.6 & 97.3 & \textbf{65} 
            & 87.4 & 95.6 & 97.3 & \textbf{65} 
            & 87.5 & 95.6 & 97.3 & \textbf{65}
            & 87.5 & 95.6 & 97.3 & \textbf{65} 
            & 87.5 & 95.6 & 97.3 & \textbf{65} & 50.1 & 64.9 & \textbf{65} \\
            {lounge} & 10
            & 90.6 & 96.9 & 98.1 &  \textbf{6}
            & 91.0 & \textbf{97.0} & \textbf{98.2} & \textbf{6} 
            & 90.7 & 96.9 & 98.1 & \textbf{6} 
            & \textbf{91.1} & \textbf{97.0} & \textbf{98.2} & \textbf{6}
            & \textbf{91.1} & \textbf{97.0} & \textbf{98.2} & \textbf{6}
            & \textbf{91.1} & \textbf{97.0} & \textbf{98.2} & \textbf{6} & 66.3 & 76.7 & \textbf{10} \\
            {meadow} & 15
            & 59.1 & 83.3 & 89.1 &  14
            & 13.7 & 29.7 & 40.8 &  \textbf{15}
            & 11.1 & 16.3 & 17.5 & 13 
            & 59.3 & 75.8 & 79.9 & \textbf{15}
            & 66.8 & 87.2 & 92.4 & \textbf{15} 
            & \textbf{75.0} & \textbf{91.5} & \textbf{94.9} & \textbf{15} & 74.8 & 84.2 & \textbf{15} \\
            {observatory} & 27
            & 63.1 & 86.0 & 91.5 &  \textbf{27}
            & {64.0} & \textbf{86.5} & \textbf{91.8} &  \textbf{27} 
            & {64.0} & \textbf{86.5} & \textbf{91.8} &  \textbf{27}
            & {64.0} & \textbf{86.5} & \textbf{91.8} &  \textbf{27}
            & {64.0} & \textbf{86.5} & \textbf{91.8} &  \textbf{27}
            & {64.0} & \textbf{86.5} & \textbf{91.8} &  \textbf{27} 
            & \textbf{64.2} & 76.2 & \textbf{27} \\
            {office} & 26
            & 59.8 & 81.9 & 88.4 &  \textbf{26}
            & 61.2 & \textbf{82.7} & \textbf{89.0} & \textbf{26} 
            & 59.5 & 81.6 & 88.0 & \textbf{26} 
            & \textbf{61.3} & \textbf{82.7} & \textbf{89.0} &\textbf{26}
            & 61.2 & \textbf{82.7} & \textbf{89.0} & \textbf{26} 
            & \textbf{61.3} & \textbf{82.7} & \textbf{89.0} &\textbf{26} & 41.6 & 51.5 & \textbf{26} \\
            {old\_computer} & 54
            & \textbf{81.6} & \textbf{93.6} & \textbf{96.2} & \textbf{54}  
            & 49.2 & 56.4 & 57.9 & \textbf{54} 
            & 23.1 & 28.1 & 29.3 & \textbf{54}
            & 49.2 & 56.4 & 57.9 & \textbf{54}
            & 49.2 & 56.4 & 57.9 & \textbf{54} 
            & 49.3 & 56.4 & 57.9 & \textbf{54} & 54.8 & 69.2 & \textbf{54} \\
            {pipes} & 14
            & 89.8 & 96.6 & 98.0 &  \textbf{14}
            & \textbf{90.4} & \textbf{96.8} & \textbf{98.1} & \textbf{14}  
            & 90.2 & 96.7 & 98.0 & \textbf{14} 
            & \textbf{90.4} & \textbf{96.8} & \textbf{98.1} & \textbf{14}
            & \textbf{90.4} & \textbf{96.8} & \textbf{98.1} & \textbf{14} 
            & \textbf{90.4} & \textbf{96.8} & \textbf{98.1} & \textbf{14} & 80.8 & 88.1 & \textbf{14} \\
            {playground} & 38
            & 89.4 & 96.4 & 97.8 & \textbf{38} 
            & 79.2 & 92.4 & 95.5 & \textbf{38} 
            & 42.7 & 51.1 & 56.1 & \textbf{38} 
            & \textbf{90.0} & \textbf{96.6} & \textbf{98.0} & \textbf{38}
            & \textbf{90.0} & 96.5 & \textbf{98.0} & \textbf{38} 
            & 89.8 & 96.5 & 97.9 & \textbf{38} & 53.0 & 66.9 & \textbf{38} \\
            {relief} & 31
            & 93.8 & 97.9 & \textbf{98.8} & \textbf{31} 
            & 93.8 & 97.9 & \textbf{98.8} & \textbf{31} 
            & \textbf{93.9} & \textbf{98.0} & \textbf{98.8} & \textbf{31} 
            & 93.8 & 97.9 & \textbf{98.8} & \textbf{31}
            & 93.8 & 97.9 & \textbf{98.8} & \textbf{31} 
            & 93.8 & 97.9 & \textbf{98.8} & \textbf{31} & 12.0 & 26.0 & \textbf{31} \\
            {relief\_2} & 31
            & \textbf{93.7} & \textbf{97.9} & \textbf{98.7} &  \textbf{31}
            & 93.6 & \textbf{97.9} & \textbf{98.7} &  \textbf{31}
            & 93.6 & \textbf{97.9} & \textbf{98.7} &  \textbf{31}
            & 93.6 & \textbf{97.9} & \textbf{98.7} &  \textbf{31}
            & 93.6 & \textbf{97.9} & \textbf{98.7} &  \textbf{31}
            & 93.6 & \textbf{97.9} & \textbf{98.7} & \textbf{31} & 13.6 & 24.3 & \textbf{31} \\
            {statue} & 11
            & \textbf{97.2} & \textbf{99.1} & \textbf{99.4} &  \textbf{11}
            & 97.1 & 99.0 & \textbf{99.4} & \textbf{11} 
            & 97.1 & 99.0 & \textbf{99.4} & \textbf{11} 
            & 97.1 & 99.0 & \textbf{99.4} & \textbf{11}
            & \textbf{97.2} & \textbf{99.1} & \textbf{99.4} & \textbf{11} 
            & 97.1 & \textbf{99.1} & \textbf{99.4} & \textbf{11} & 87.0 & 92.2 & \textbf{11} \\
            {terrace} & 23
            & 90.0 & 96.7 & 98.0 & \textbf{23} 
            & \textbf{91.3} & \textbf{97.1} & \textbf{98.3} & \textbf{23} 
            & 90.5 & 96.8 & 98.1 & \textbf{23} 
            & 90.6 & 96.9 & 98.1 & \textbf{23}
            & 90.6 & 96.9 & 98.1 & \textbf{23} 
            & 90.6 & 96.9 & 98.1 & \textbf{23} & 80.8 & 88.2 & \textbf{23} \\
            {terrace\_2} & 13
            & 89.5 & 96.5 & 97.9 & \textbf{13} 
            & \textbf{89.8} & \textbf{96.6} & \textbf{98.0} & \textbf{13} 
            & \textbf{89.8} & \textbf{96.6} & \textbf{98.0} & \textbf{13} 
            & \textbf{89.8} & \textbf{96.6} & \textbf{98.0} & \textbf{13}
            & \textbf{89.8} & \textbf{96.6} & \textbf{98.0} & \textbf{13} 
            & \textbf{89.8} & \textbf{96.6} & \textbf{98.0} & \textbf{13} & 84.1 & 90.4 & \textbf{13} \\
            {terrains} & 42
            & 79.5 & 92.7 & 95.6 & \textbf{42} 
            & 78.6 & 89.4 & 91.7 & \textbf{42} 
            & \textbf{83.2} & \textbf{94.1} & \textbf{96.4} & \textbf{42} 
            & \textbf{83.2} & \textbf{94.1} & \textbf{96.4} & \textbf{42}
            & \textbf{83.2} & \textbf{94.1} & \textbf{96.4} & \textbf{42} 
            & \textbf{83.2} & \textbf{94.1} & \textbf{96.4} & \textbf{42} & 78.0 & 86.5 & \textbf{42} \\
            \hline
            {Average} & 35.9 
            & \textbf{86.2} & \textbf{94.8} & \textbf{96.8} & 35.0 
            & 80.0 & 88.5 & 90.9 & {35.8}
            & 76.6 & 84.3 & 86.2 & 35.5
            & 84.2 & 92.7 & 94.7 & {35.8}
            & 84.5 & 93.1 & 95.1 & {35.8}
            & \underline{84.8} & \underline{93.3} & \underline{95.3} & {35.8} & 55.1 & 65.1 & \textbf{35.9} \\
            \hline
        \end{tabular}
   }
   \vspace{-0.2cm}
	\label{tab:dslr}
\end{table*}

\begin{table*}
\caption{Camera position accuracy on the KITTI~\citep{geiger2013vision} dataset. $N_t$ and $N_c$ denote the number of input and registered images, respectively; $\tilde{e}$ and $\bar{e}$ denote the median and mean camera position error in meters, respectively; $T$ represents the runtime in seconds. The best results are shown in bold and the second-best results are underlined.} 	
	\setlength{\tabcolsep}{1.0pt}
	\centering
    \scalebox{0.69}{
        \begin{tabular}{|c|c||c|c|c|c||c|c|c|c||c|c|c|c||c|c|c|c||c|c|c|c||c|c|c|c|}
            \hline		
            \multicolumn{2}{|c||}{method} &			
            \multicolumn{4}{c||}{COLMAP} &
            \multicolumn{4}{c||}{CReTA} &
            \multicolumn{4}{c||}{LiGT} &
            \multicolumn{4}{c||}{HETA} &
            \multicolumn{4}{c||}{GLOMAP} &
            \multicolumn{4}{c|}{HETA++}\\			
            \hline
            Data & $N_t$ 
            & $\tilde{e}\downarrow$ & $\bar{e} \downarrow$ & $N_c\uparrow$ & $T\downarrow$
            & $\tilde{e}\downarrow$ & $\bar{e}\downarrow $ & $N_c\uparrow$ & $T\downarrow$
            & $\tilde{e}\downarrow$ & $\bar{e}\downarrow $ & $N_c\uparrow$ & $T\downarrow$
            & $\tilde{e}\downarrow$ & $\bar{e}\downarrow $ & $N_c\uparrow$ & $T\downarrow$
            & $\tilde{e}\downarrow$ & $\bar{e}\downarrow $ & $N_c\uparrow$ & $T\downarrow$
            & $\tilde{e}\downarrow$ & $\bar{e}\downarrow $ & $N_c\uparrow$ & $T\downarrow$\\
            \hline
            {00} & 9082 
            & 0.9  & 3.7  & 9082  & 1.5e5
            & 1.2  & 2.0 & 9082  & 2809
            & 33.6  & 69.1 & 9082  & 10220
            & \underline{0.7}  & 0.9  & 9082 & 3353 
            &  \underline{0.7} & \textbf{0.8}  & 9082 & 4035
            & \textbf{0.6}  & \textbf{0.8}  & 9082 & \textbf{2394} \\ 
            {01} & 2202 
            & 1.5  & \underline{3.0}  & 2202  & 1.2e4
            & 1.0  & 5.7  & 2202 & 1006
            & 1.6  & 5e2  & 2202  & 1984
            & \textbf{0.5}  & 6.2  & 2202 & 1016
            &  1.3 & 73.4  &  2202 & 1503
            & \textbf{0.5}  & \textbf{2.6}  & 2202& \textbf{763} \\ 
            {02} & 9322 
            & \underline{4.0}  & 12.3  & 9322  & 1.8e5
            & 5.1 & 9.0  & 9322  & 2978
            & 2e2  & 5e2  & 9322  & 3998
            & 4.6  & \underline{8.2}  & 9322&  3419
            & 5.4  & 1e3  & 9322 &6472
            & \textbf{2.1}  & \textbf{2.9}  & 9322 &\textbf{2666} \\ 
            {03} & 1602 
            & \textbf{0.2}  & 0.8  & 1602  & 1.8e4
            & \textbf{0.2}  & \textbf{0.2}  & 1602& 470
            & 0.3  & 0.6  & 1602  & 818
            &  \textbf{0.2} & \textbf{0.2}  & 1602& 500
            & 0.3  & 0.5  & 1602  & 850
            & \textbf{0.2}  & \textbf{0.2}  & 1602&  \textbf{397}\\ 
            {04} & 542
            &  \textbf{0.1} & \textbf{0.2}  & 542  & 1526
            &  \textbf{0.1} & \textbf{0.2}  & 542 &  77
            &  \textbf{0.1} & \textbf{0.2}  & 542 & 149
            &  \textbf{0.1} & \textbf{0.2}  & 542 & 98
            &  \textbf{0.1} & \textbf{0.2}  & 542 & 178
            &  \textbf{0.1} & \textbf{0.2}  & 542 & \textbf{75}\\ 
            {05} & 5522 
            & 0.8  &  2.1 & 5522  & 4.9e4
            & 0.1  & 2.3  & 5522  & 2456
            & 0.4 & 99.2 & 5522 & 6501
            & \textbf{0.1}  & \textbf{0.4}  & 5522 & 2548
            &  \textbf{0.1} & \textbf{0.4}  & 5522 & 3251
            & \textbf{0.1}  & \textbf{0.4}  & 5522  &\textbf{2269}\\ 
            {06} & 2202 
            & 0.2  & 1.5  &  2202 & 2.1e4
            & \textbf{0.1}  & 0.5  &   2202&574
            & \textbf{0.1}  & 2e3  &   2202&1188
            &  \textbf{0.1} & \textbf{0.2}  &   2202&624
            & \textbf{0.1}  &  0.5 &   2202& 890
            &  \textbf{0.1} & \textbf{0.2}  & 2202 & \textbf{508}\\ 
            {07} & 2202 
            & 0.5  & 2.1  & 2202  & 1.7e4
            & \textbf{0.1}  & \textbf{0.2}  &  2202& 561
            & 0.3 & 15.6 & 2202  & 1352
            & \textbf{0.1}  & \textbf{0.2}  & 2202  & 674
            &  \textbf{0.1} & \textbf{0.2}  &  2202 &910
            & \textbf{0.1}  & \textbf{0.2}  &  2202 &\textbf{522}\\ 
            {08} & 8142
            & 6.0  & 11.5  & 8142  & 1.0e5
            & 3.7 & 8.1  &  8142 & 2531
            & 1e2  & 6e2  &  8142  & 2975
            & \textbf{2.0} & \underline{3.0}  & 8142  & 3759
            & 3.0  & 4.4  &  8142 & 5345
            & \underline{2.3}  & \textbf{2.7} & 8142 &\textbf{2102} \\ 
            {09} & 3182 
            & 1.1  & 3.4  & 3182 & 2.0e4
            & \underline{0.8} & \underline{1.3}  &  3182& 581
            & 56.3  & 4e2  &   3182& 1479
            & \underline{0.8}  & 1.6  &   3182& 703
            & \underline{0.8}  & \underline{1.3}  &   3182&1284
            & \textbf{0.5}  & \textbf{1.1}  &   3182&\textbf{522}\\
            {10} & 2402 
            &  2.4 & 4.2  &  2402 & 1.3e4
            & \textbf{0.5}  & 2.6  &  2402 & 660
            & 6.8  & 1e2  &  2402 & 898
            &  \textbf{0.5} &  2.7 &  2402& 675
            & 0.6  & \textbf{2.0}  &  2402 &1124
            & \textbf{0.5}  & \underline{2.6}  & 2402& \textbf{496} \\ 
            \hline 
            {Average} & 4218.4  
            & 1.6 & 4.1  & \textbf{4218.4} &  5e4
            & 1.2  & 2.9  & \textbf{4218.4} & 1337 
            & 36.3  & 3e2  & \textbf{4218.4} & 2869
            &  \underline{0.9} &  \underline{2.2} & \textbf{4218.4} & 1579 
            & 1.1  &  98.5 &\textbf{ 4218.4} & 2349
            & \textbf{0.6}  & \textbf{1.3}  & \textbf{4218.4} & \textbf{1156} \\ 
            \hline 
        \end{tabular}
    }
    \vspace{-0.2cm}
	\label{tab:KITTI}
\end{table*}
\subsection{Evaluation On Unordered Data}
\begin{figure}[htbp]
    \vspace{-0.3cm}
    \centering
    \subfloat[Roman Forum]{\includegraphics[width=0.23\textwidth]{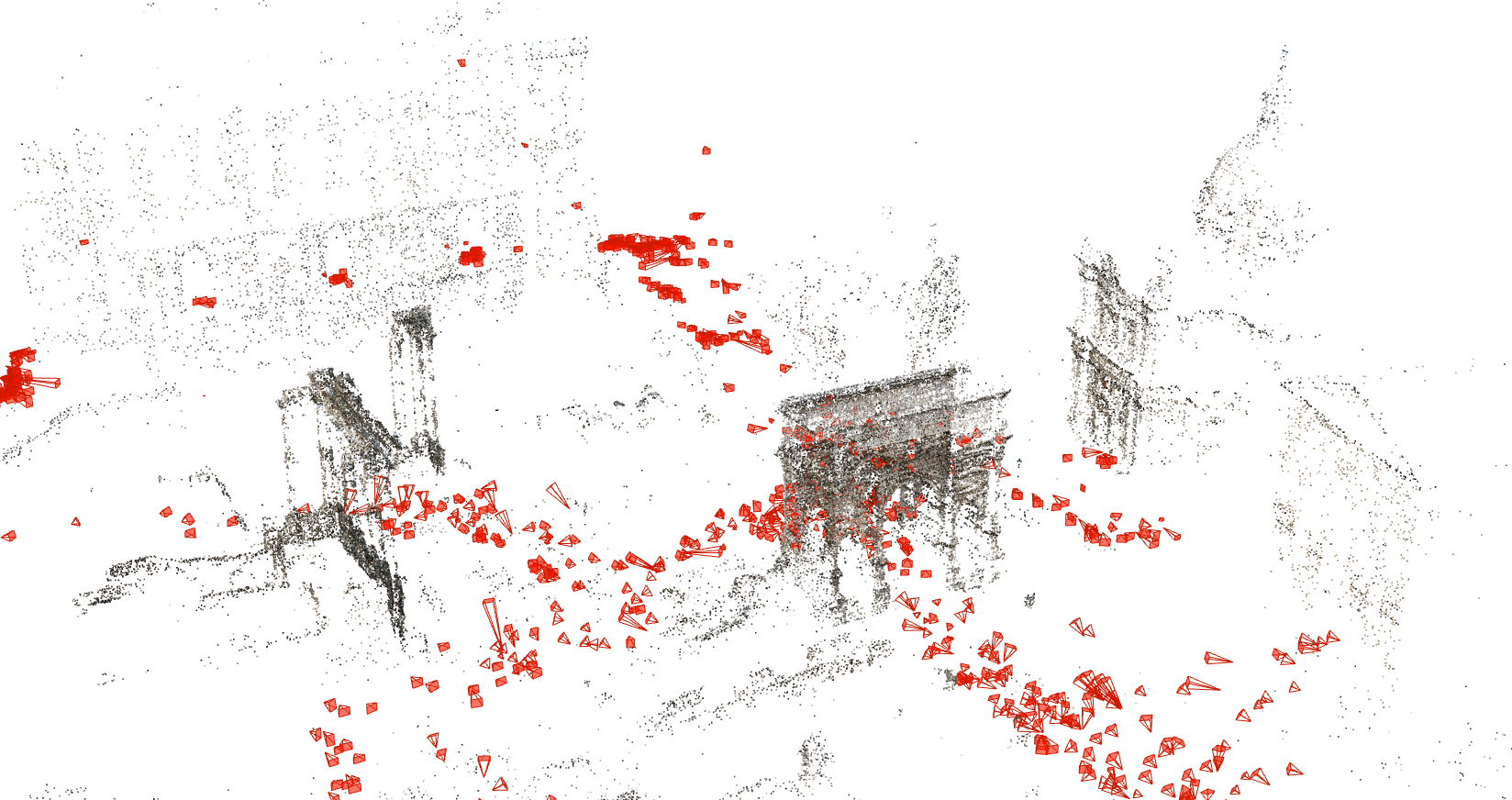}\label{fig:ROF}}
    \hfill
    \subfloat[Piccadilly]{\includegraphics[width=0.23\textwidth]{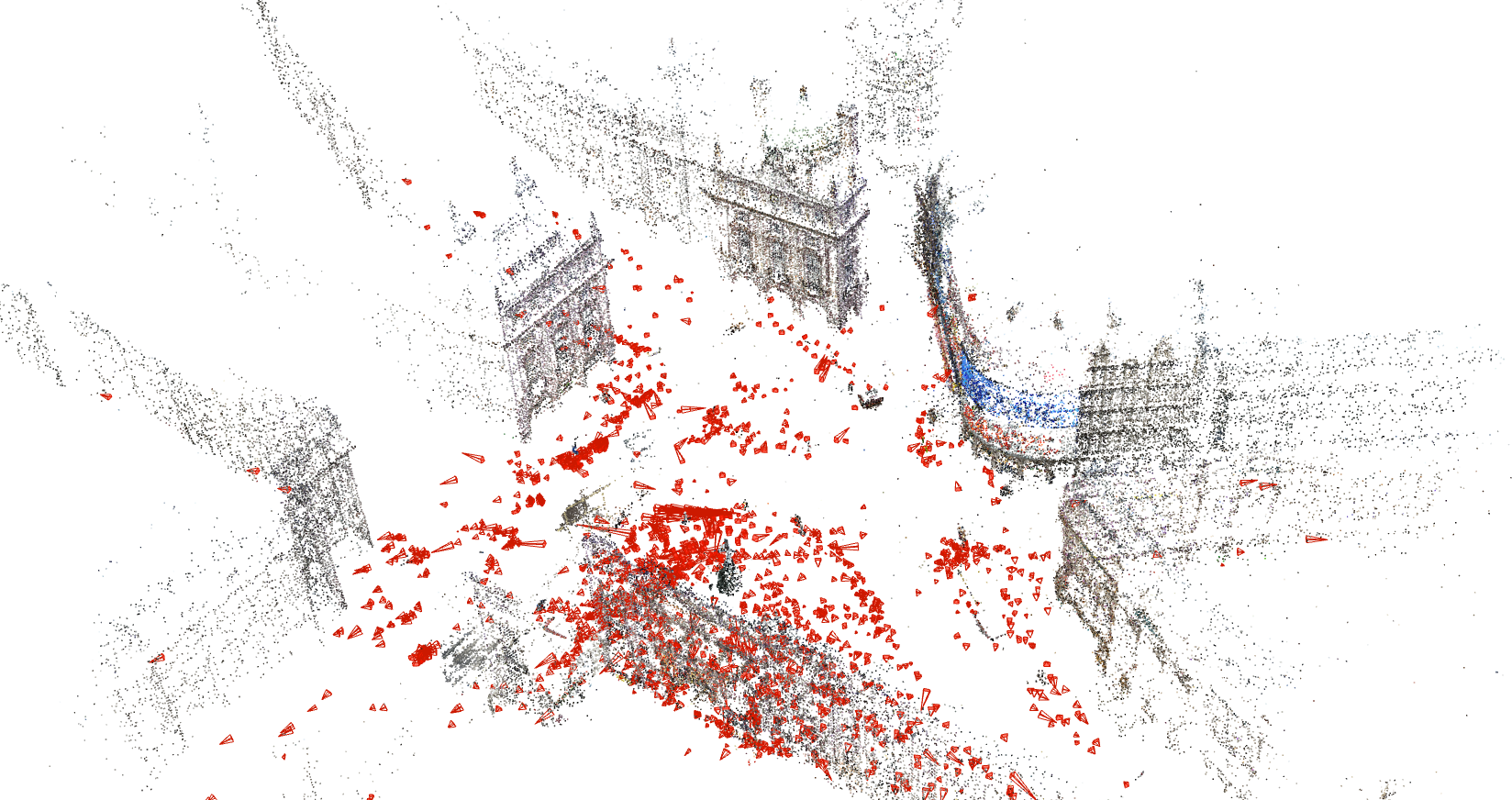}\label{fig:PIC}}
    \caption{Sample results produced by our HETA++}
    \label{fig:1dsfm_qua}
    \vspace{-0.2cm}
\end{figure}
\subsubsection{1DSfM Dataset}
Due to the limited accuracy of the provided camera intrinsics and the significant presence of outlier feature matches in the 1DSfM dataset, the estimated relative poses exhibit large errors. As a result, the accuracy of the global camera rotations estimated by \citet{chatterjee2017robust} is relatively low. The dataset presents particular challenges, including numerous outlier feature matches in the feature tracks and low-precision camera rotations. The reconstruction results after bundle adjustment are shown in Table~\ref{tab:1DSFM}, with the corresponding runtime of all methods presented in Table~\ref{tab:1DSFM_time}. Since all methods utilize the same view track graph, the associated time costs are excluded from the runtime comparison.

Compared to global methods, COLMAP fails to register many images. Therefore, we compare the AUC scores among the global methods. Among these, our method, HETA++, achieves the best performance in terms of accuracy. In terms of the number of registered images, CReTA, GLOMAP, and our method perform equally well. We then compare the efficiency of these methods, with our method outperforming the others. The LiGT method is sensitive to outliers in the feature tracks due to its reliance on implicit 3D points, which causes it to lose many images and results in a relatively lower AUC score.

Comparing the implicit method LiGT with explicit methods such as HETA, GLOMAP, and HETA++ shows that explicit 3D point optimization improves global translation estimation. CReTA performs worse because it lacks angle-based point refinement after translation averaging. By introducing this refinement before bundle adjustment and selecting reliable feature tracks, HETA++ achieves higher accuracy and efficiency than HETA and GLOMAP.

In conclusion, for the 1DSfM dataset, our method performs the best and we show two reconstruction results in Fig.~\ref{fig:1dsfm_qua}.


\subsubsection{ETH3D MVS (DSLR) Dataset}
For this small-scale unordered dataset, the time costs of all compared methods are similar; therefore, we primarily focus on comparing their accuracy, as shown in Table~\ref{tab:dslr}. All global methods perform comparably to the incremental COLMAP in most scenes, except for the ``old\_computer'' dataset. The ``old\_computer'' scene contains many symmetric textures, leading to numerous erroneous edges in the view-track graph. As a result, the global rotation averaging module produces inaccurate camera rotations, causing all global methods to fail. Our proposed HETA++ outperforms other methods in the weak-texture scene ``meadow", demonstrating the robustness of our system in handling scenes with few feature tracks. Although $\pi^3$+BA registers all input images, its average accuracy remains notably lower than that of geometry-based methods, indicating that BA cannot fully compensate for inaccurate feed-forward initialization.

\begin{figure*}	
	\centering
	\includegraphics[width=\linewidth, trim = 0mm 0mm 0mm 0mm, clip]{KITTI.png}
	\caption{Comparison of camera motion trajectories on the sequence 02 and sequence 08 of KITTI~\citep{geiger2013vision} odometry benchmark. The sample state-of-the-art SfM methods include COLMAP~\citep{schoenberger2016mvs}, CReTA~\citep{manam2022correspondence}, LiGT~\citep{cai2021pose},
    HETA~\citep{tao2024revisiting},
    GLOMAP~\citep{pan2024glomap} and our method HETA++}
	\label{fig:KITTI_result}
\end{figure*}
\subsection{Evaluation On Sequential Data}
\subsubsection{KITTI Dataset}
\begin{table*}
    \caption{Camera-pose AUC on the ETH3D MVS (rig) ~\citep{ETH3DDataset} dataset.  $N_t$ and $N_c$ denote the number of input and registered images. The best results are shown in bold and the second best results are underlined.}	
	\setlength{\tabcolsep}{2.0pt}
	\centering
    \scalebox{0.65}{
        \begin{tabular}{|c|c||ccc|c||ccc|c||ccc|c||ccc|c||ccc|c||ccc|c|}
            \hline		
            \multicolumn{2}{|c||}{method} &			
            \multicolumn{4}{c||}{COLMAP} &
            \multicolumn{4}{c||}{CReTA} &
            \multicolumn{4}{c||}{LiGT} &
            \multicolumn{4}{c||}{HETA} &
            \multicolumn{4}{c||}{GLOMAP} &
            \multicolumn{4}{c|}{HETA++} \\			
            \hline
            \multirow{2}{*}{Data} &  \multirow{2}{*}{$N_t$} 
            &\multicolumn{3}{c|}{AUC@} &\multirow{2}{*}{$N_c\uparrow$}  
            &\multicolumn{3}{c|}{AUC@}  &\multirow{2}{*}{$N_c\uparrow$}  
            &\multicolumn{3}{c|}{AUC@}  &\multirow{2}{*}{$N_c\uparrow$}  
            &\multicolumn{3}{c|}{AUC@}  &\multirow{2}{*}{$N_c\uparrow$}  
            &\multicolumn{3}{c|}{AUC@}  &\multirow{2}{*}{$N_c\uparrow$}  
            &\multicolumn{3}{c|}{AUC@}  &\multirow{2}{*}{$N_c\uparrow$}  \\
            & &$1^\circ\uparrow$  & $3^\circ\uparrow$ &  $5^\circ\uparrow$
            & &$1^\circ\uparrow$  & $3^\circ\uparrow$ &  $5^\circ\uparrow$
            & &$1^\circ\uparrow$  & $3^\circ\uparrow$ &  $5^\circ\uparrow$
            & &$1^\circ\uparrow$  & $3^\circ\uparrow$ &  $5^\circ\uparrow$
            & &$1^\circ\uparrow$  & $3^\circ\uparrow$ &  $5^\circ\uparrow$
            & &$1^\circ\uparrow$  & $3^\circ\uparrow$ &  $5^\circ\uparrow$ &\\
            \hline
            \text{delivery\_area} & 948 
            & 71.3 & 89.2 & 93.3 & \textbf{948}
            & 76.6 & 91.4 & 94.7 & \textbf{948} 
            & \underline{77.2} & \underline{91.6} & \underline{94.8} & \textbf{948} 
            & \textbf{77.4} & \textbf{91.7} & \textbf{94.8} & \textbf{948} 
            & 76.6 & 91.4 & 94.7 & \textbf{948} 
            & 76.9 & 91.5 & 94.7 & \textbf{948}  \\ 
            {electro} & 1200
            & 36.4 & 64.9 & 74.9 & \textbf{1200} 
            & 48.3 & 74.6 & 83.4 & \textbf{1200}
            & 24.5 & 64.4 & 77.1 & \textbf{1200}
            & \textbf{49.5} & \textbf{75.3} & \textbf{83.9} & \textbf{1200}
            & 48.0& 74.2& 83.2& \textbf{1200}
            & \underline{49.0}& \underline{74.8}& \underline{83.6}&  \textbf{1200}\\
            {forest} & 1028
            & 0 & 0 & 0 & \textbf{1027} 
            & 70.4 & 89.1 & 93.3 & 1022
            & 56.7 & 69.8 & 74.1 & 999
            & 70.2 & \textbf{89.0} & \textbf{93.3} & 1022
            & \textbf{70.3} & \textbf{89.0} & \textbf{93.3} & 1022
            & \textbf{70.3}& \textbf{89.0}& \textbf{93.3}& 1022 \\
            {playground} & 960
            & 0.3 & 6.6 & 12.1 & \textbf{954}
            & 37.4 & 71.0 & 80.9 & \textbf{954}  
            & 11.0 & 23.0 & 32.4 & 953
            & 36.9 & 70.5 & 80.6 & \textbf{954}   
            & \underline{38.0} & \underline{71.6} & \underline{81.1} & \textbf{954}
            & \textbf{38.7}& \textbf{72.2}& \textbf{81.5}& \textbf{954}\\
            {terrains} & 660
            & 46.3 & 76.7 & 84.7 & \textbf{660}
            & 47.4 & 77.1 & 84.9 & \textbf{660}  
            & \textbf{49.2} & \textbf{78.1} & \textbf{85.6} & \textbf{660}    
            & 48.0 & 77.5 & 85.1 & \textbf{660}
            & 47.5 & 77.2 & 84.9 & \textbf{660}
            & \underline{48.5}& \underline{77.6} & \underline{85.0}& \textbf{660}\\
            \hline
            {Average} & 959.2
            & 30.9 & 47.5 & 53.0 & \textbf{957.8}
            & 56.0 & 80.6 & 87.4 & 956.8  
            & 43.7 & 65.4 & 72.8 & 952    
            & \underline{56.4} & \underline{80.8} & \underline{87.5} & 956.8
            & 56.1 & 80.7 & 87.4 & 956.8
            & \textbf{56.7} & \textbf{81.0} & \textbf{87.6} & 956.8\\
            \hline
        \end{tabular}
    }
	\label{tab:rig}
    \vspace{-0.2cm}
\end{table*}
   
\begin{table*}
    \caption{Camera-pose accuracy on LaMAR \citep{sarlin2022lamar} dataset. $N_t$ and $N_c$ denote the number of input and registered images, respectively.
    $\tilde{e}$ denotes the median camera position error in meters. $T$ represents the runtime in seconds. The best results are shown in bold and the second-best are underlined.}
	\setlength{\tabcolsep}{1pt}
	\centering
    \scalebox{0.67}{
        \begin{tabular}{|c|c||c|cc|c|c||c|cc|c|c||c|cc|c|c||c|cc|c|c||c|cc|c|c|}
            \hline		
            \multicolumn{2}{|c||}{method} &			
            \multicolumn{5}{c||}{COLMAP} &
            \multicolumn{5}{c||}{CReTA} &
            \multicolumn{5}{c||}{HETA} &
            \multicolumn{5}{c||}{GLOMAP} &
            \multicolumn{5}{c|}{HETA++} \\			
            \hline
            \multirow{2}{*}{Data} &  \multirow{2}{*}{$N_t$} 
            &\multirow{2}{*}{$\tilde{e}\downarrow$} &\multicolumn{2}{c|}{AUC@} &\multirow{2}{*}{$N_c\uparrow$} &\multirow{2}{*}{$T\downarrow$}    
            &\multirow{2}{*}{$\tilde{e}\downarrow$} &\multicolumn{2}{c|}{AUC@} &\multirow{2}{*}{$N_c\uparrow$} &\multirow{2}{*}{$T\downarrow$}    
            &\multirow{2}{*}{$\tilde{e}\downarrow$} &\multicolumn{2}{c|}{AUC@} &\multirow{2}{*}{$N_c\uparrow$} &\multirow{2}{*}{$T\downarrow$}    
            &\multirow{2}{*}{$\tilde{e}\downarrow$} &\multicolumn{2}{c|}{AUC@} &\multirow{2}{*}{$N_c\uparrow$} &\multirow{2}{*}{$T\downarrow$}    
            &\multirow{2}{*}{$\tilde{e}\downarrow$} &\multicolumn{2}{c|}{AUC@} &\multirow{2}{*}{$N_c\uparrow$} &\multirow{2}{*}{$T\downarrow$}    \\
            & & &$5^\circ\uparrow$  & $10^\circ\uparrow$ & 
            & & &$5^\circ\uparrow$  & $10^\circ\uparrow$ & 
            & & &$5^\circ\uparrow$  & $10^\circ\uparrow$ & 
            & & &$5^\circ\uparrow$  & $10^\circ\uparrow$ & 
            & & &$5^\circ\uparrow$  & $10^\circ\uparrow$ & &  \\
            \hline
            \text{CAB} & 33587 
            & 7.8& 4.7& 14.0&26635 & 302199 
            & 15.5& 2.6& 5.4&\textbf{30914}& 5694
            & \textbf{7.2}& \textbf{4.8}& 12.0& 30897&8163
            & 13.2& 3.9& 9.6& \textbf{30914}& \textbf{7832}
            & \underline{7.5}& \underline{4.7}& \textbf{14.6}& \textbf{30914}&8001\\ 
            \hline
            \text{HGE} & 25881 
            & 3.7& 19.5& 28.1& 24254& 331921
            &1.8 &27.8 & 42.2& \textbf{24770}& 8697
            & \textbf{1.0}& \underline{34.3}& \underline{53.6}& 24765& 9333
            & 1.8& 32.0& 50.8& \textbf{24770}& 8733
            & \textbf{1.0}& \textbf{41.9}& \textbf{56.3}& \textbf{24770}& \textbf{7827} \\
            \hline
            \text{LIN} & 37677 
            &1.6 & 45.2 & 64.3 & 35927& 658440
            & 0.4 & 73.8& 85.5& 36418& \textbf{12639}
            & \underline{0.3}& \underline{78.9}& \underline{88.5}& 36434& 15368
            & \underline{0.3}& 78.2& 87.9& \textbf{36437} & 14922
            & \textbf{0.2}& \textbf{82.3}& \textbf{90.3}& \textbf{36437}& {14823} \\
            \hline
        \end{tabular}
    }
	\label{tab:lamar}
\end{table*}
   
The quantitative reconstruction results are presented in Table~\ref{tab:KITTI}, where HETA++ generally achieves the highest accuracy and efficiency among all the methods compared.
Fig.~\ref{fig:KITTI_result} shows the computed camera trajectories for the two largest datasets, 02 and 08. Our results are closer to the ground truth camera motion trajectories.

Since most camera trajectories in the KITTI dataset are collinear, the pure translation averaging method CReTA encounters degenerate cases due to its reliance solely on camera-to-camera constraints, resulting in poor camera pose initialization for the bundle adjustment.
The implicit global translation estimation method LiGT estimates the camera positions based on matrix decomposition, which enhances efficiency but compromises accuracy and robustness.
With a reasonable initialization, HETA outperforms GLOMAP in terms of robustness in scenes 01 and 02, where many outlier feature matches exist in feature tracks. By selecting reliable feature tracks, our method achieves improved accuracy and efficiency, particularly in the largest scene 02.

\begin{figure}[htbp]
    \centering
    \subfloat[ bridge build by HETA++]{\includegraphics[width=0.23\textwidth]{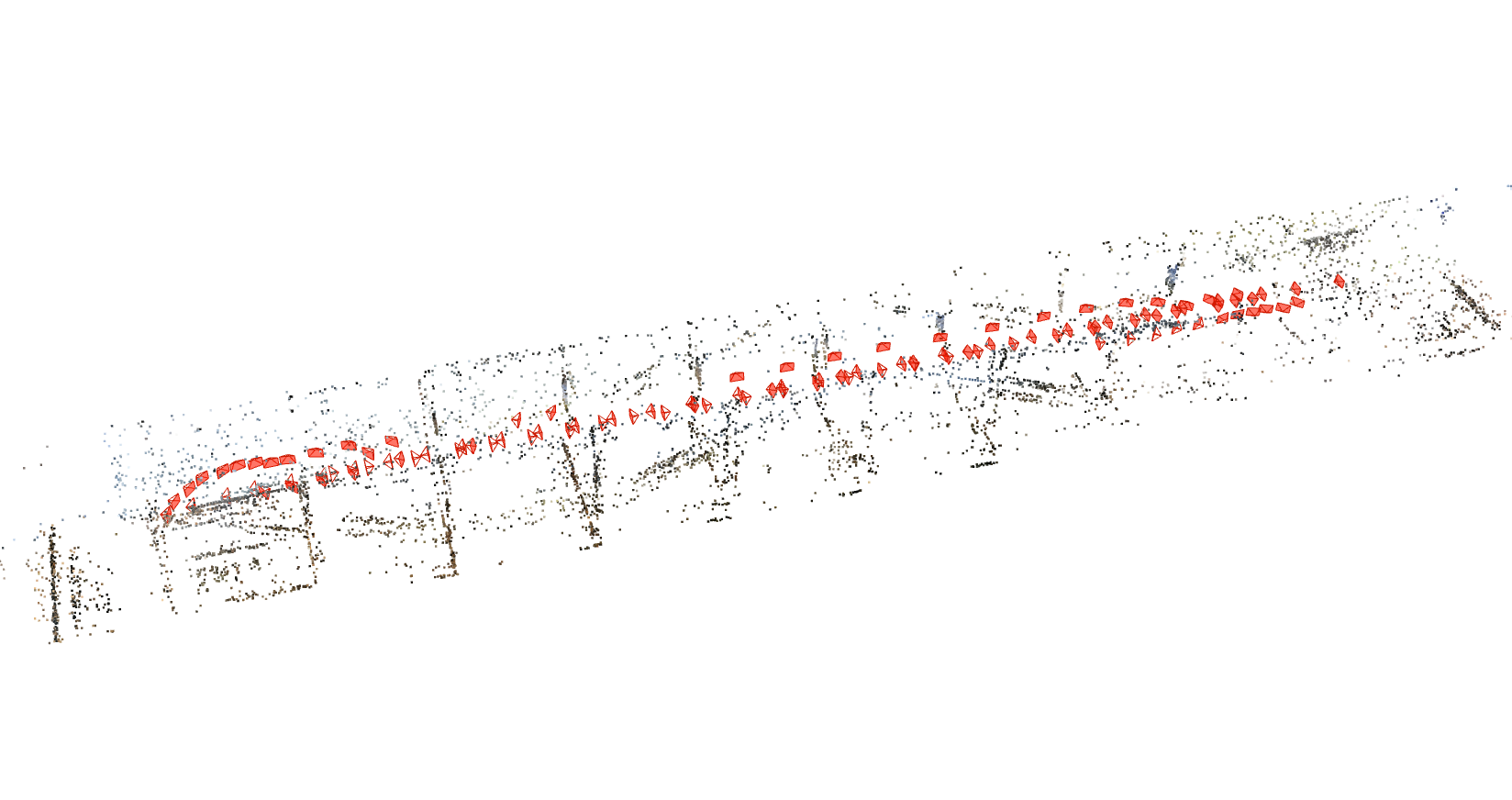}\label{fig:bridge}}
    \hfill
    \subfloat[electro build by HETA++]{\includegraphics[width=0.23\textwidth]{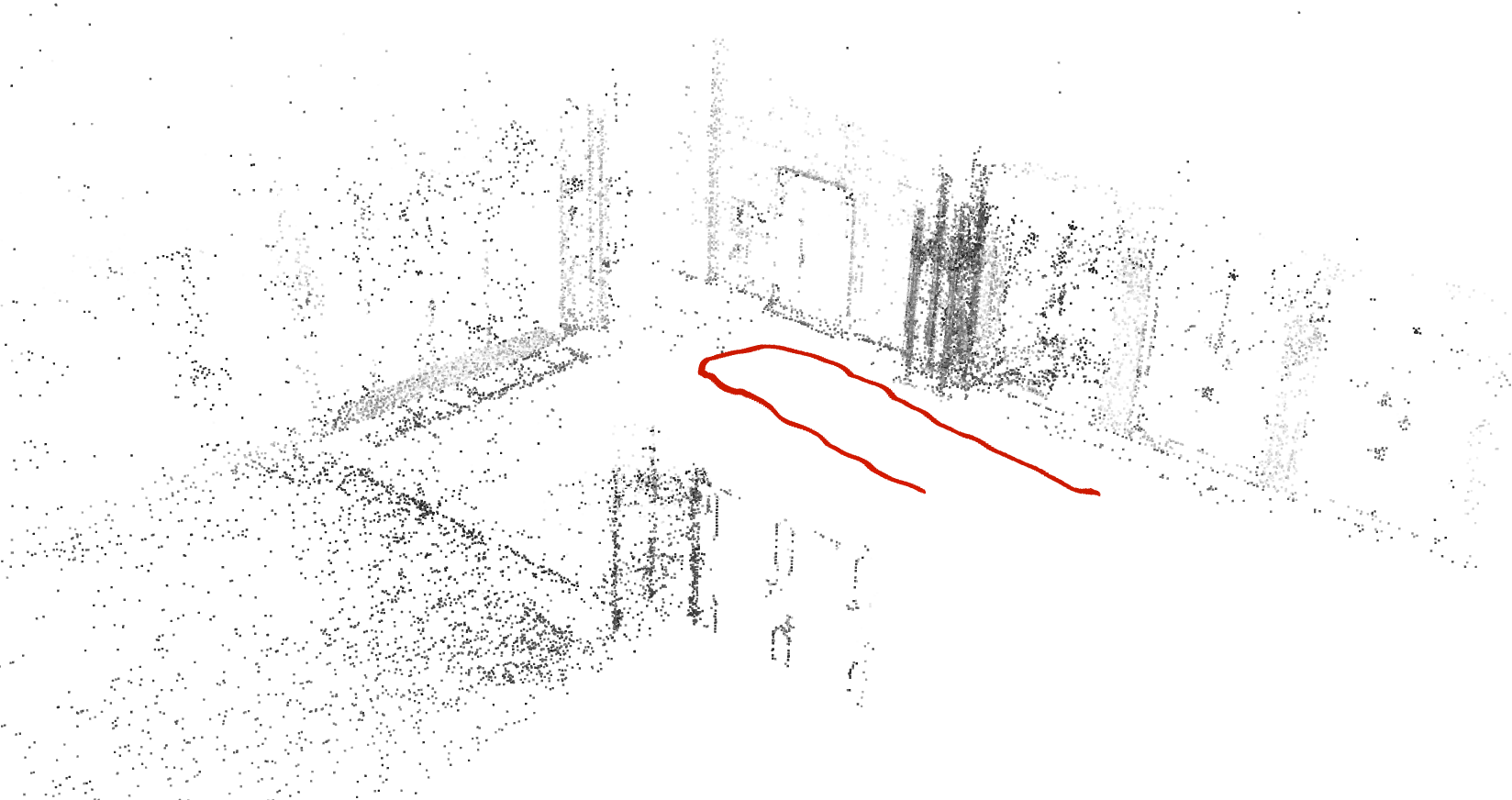}\label{fig:electro}}\\
    \subfloat[forest build by  HETA++]{\includegraphics[width=0.23\textwidth]{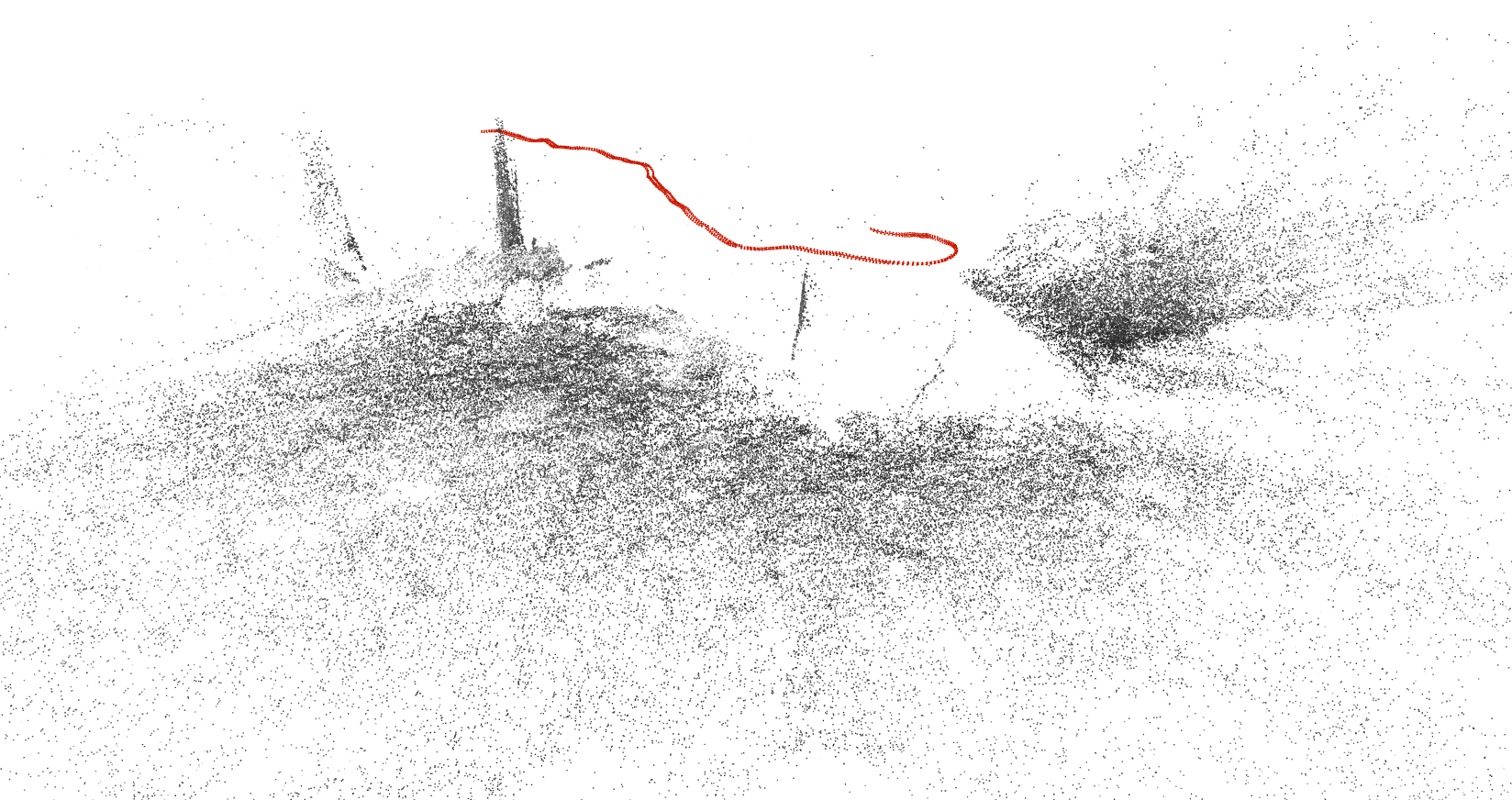}\label{fig:forest_hetapp}}
    \hfill
    \subfloat[forest build by COLMAP]{\includegraphics[width=0.23\textwidth]{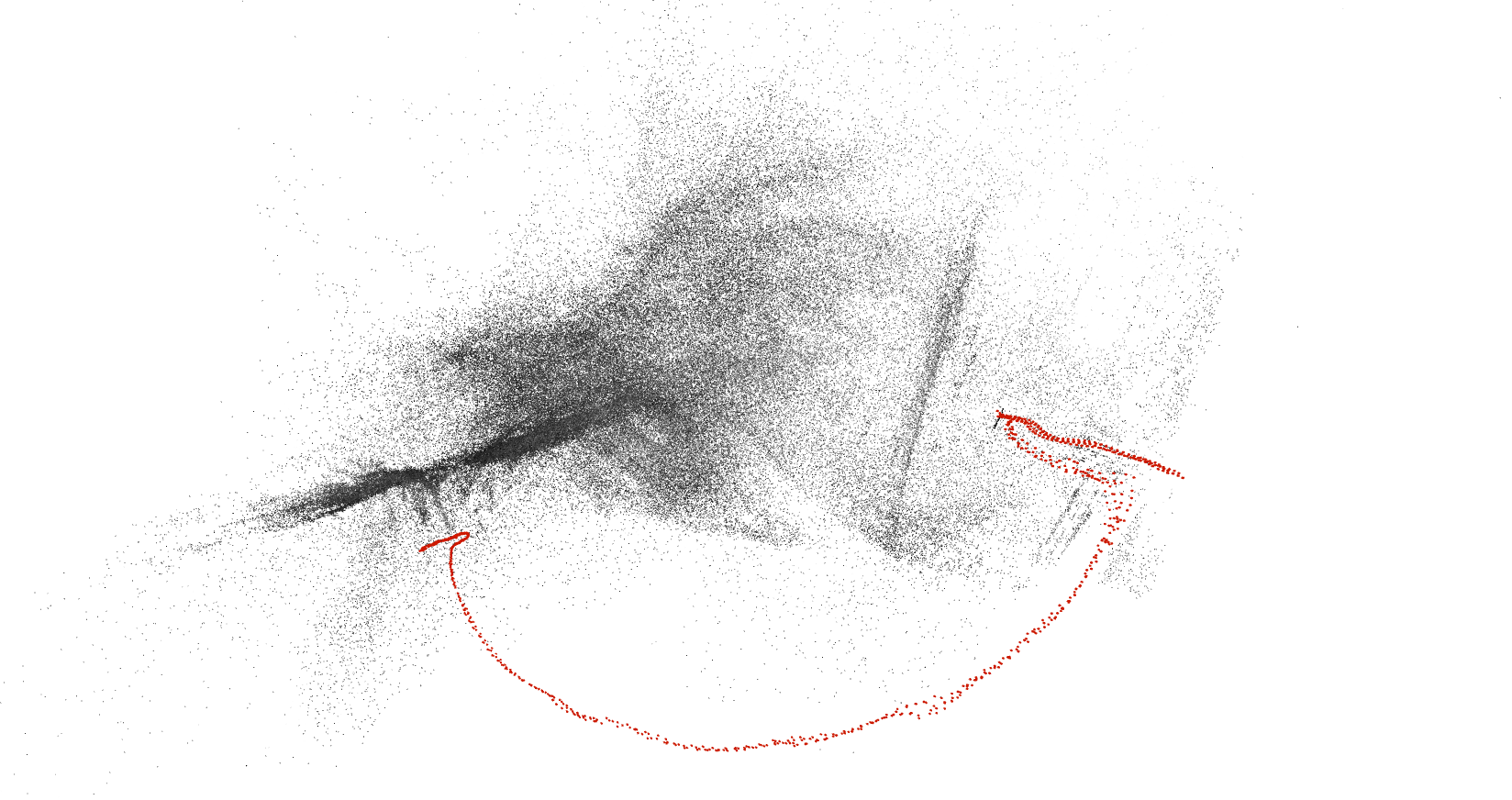}\label{fig:forest_colamp}}
    \caption{This figure shows some reconstruction results of ETH3D MVS~\citep{ETH3DDataset}}
    \label{fig:ETH3D_qua}
\end{figure}
Consequently, these reconstruction results demonstrate the superiority of HETA++ in terms of both accuracy and efficiency. Compared to GLOMAP, HETA++ provides a more reasonable initialization with greater robustness.

\subsubsection{ETH3D MVS (rig) Dataset}
Similar to GLOMAP~\citep{pan2024glomap}, we compare the AUC scores for all methods and present the results in Table~\ref{tab:rig}.
For this dataset, the accuracy of all global methods is comparable, with HETA++ achieving the best accuracy on average.
Due to the absence of a camera motion loop, COLMAP performs slightly worse than global methods.
Note that COLMAP reconstructs a wrong result in the scene ``forest", despite  registering the largest number of images among all methods. 
Fig.~\ref{fig:ETH3D_qua} shows the corresponding reconstruction results produced by COLMAP and HETA++.
\begin{figure*}[htbp]
    \vspace{-0.4cm}
    \centering
    \subfloat[CAB]{\includegraphics[width=0.33\textwidth]{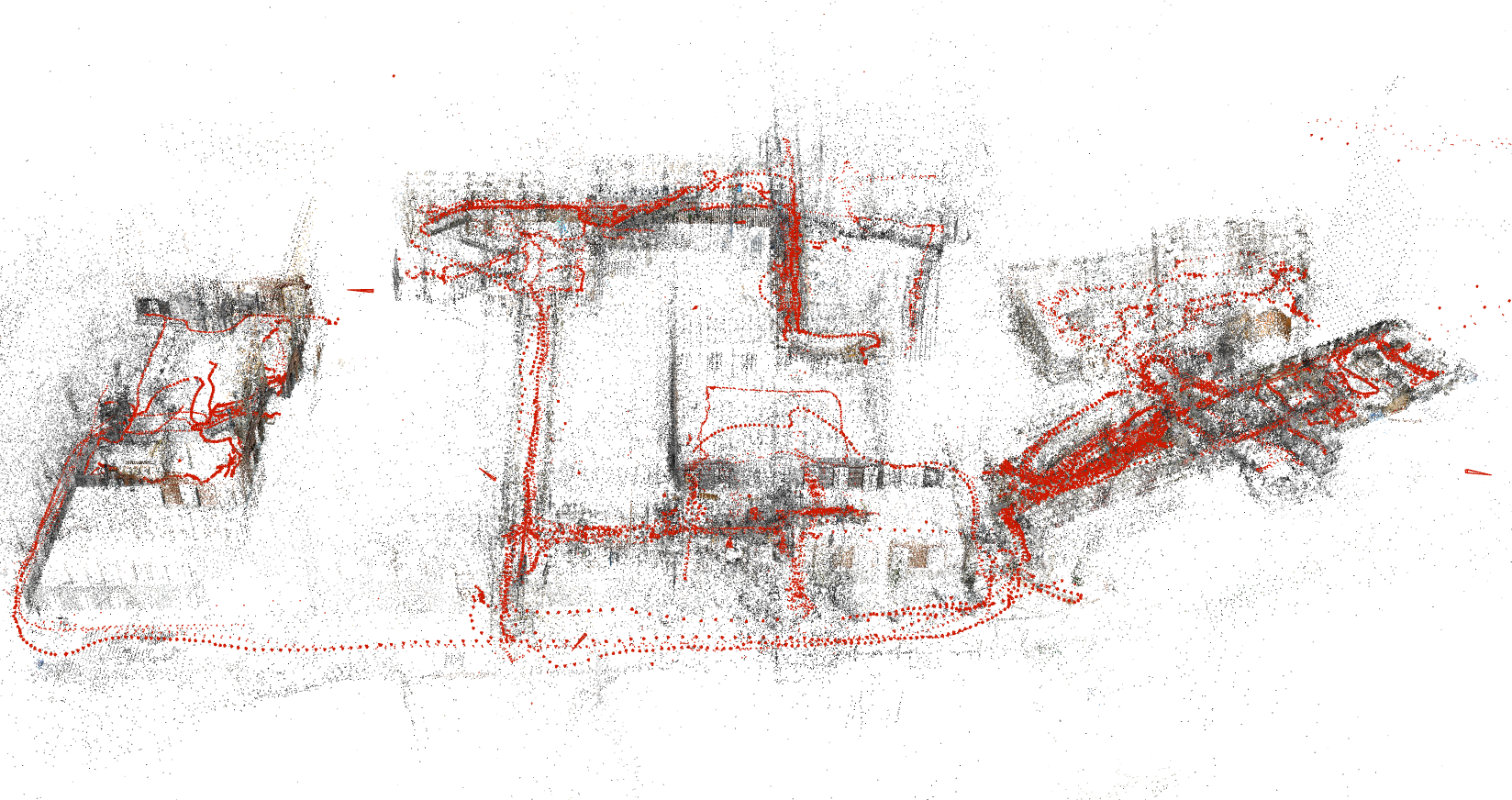}\label{fig:CAB}}
    \hfill
    \subfloat[HGE]{\includegraphics[width=0.33\textwidth]{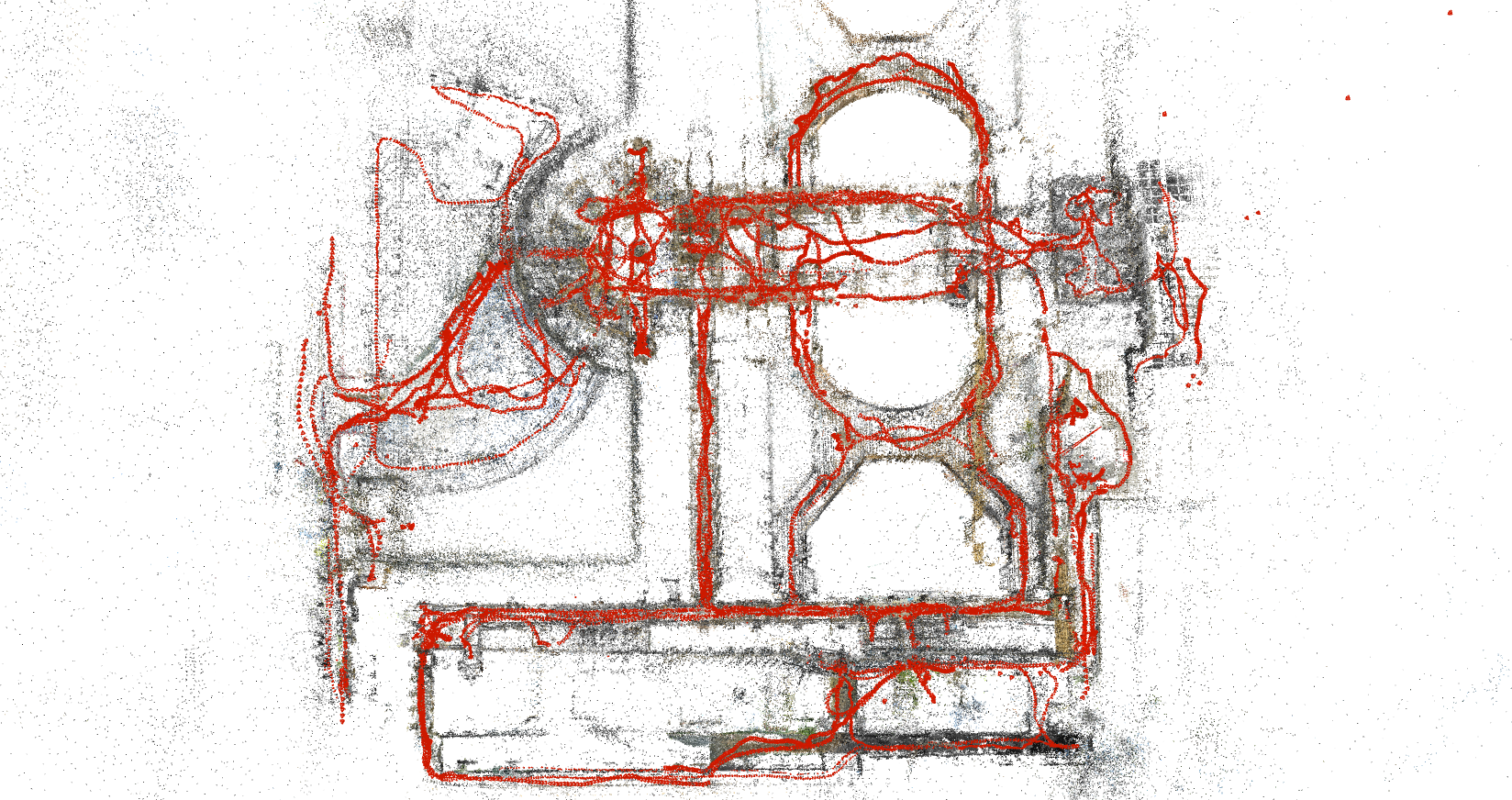}\label{fig:HGE}}
    \hfill
    \subfloat[LIN]{\includegraphics[width=0.33\textwidth]{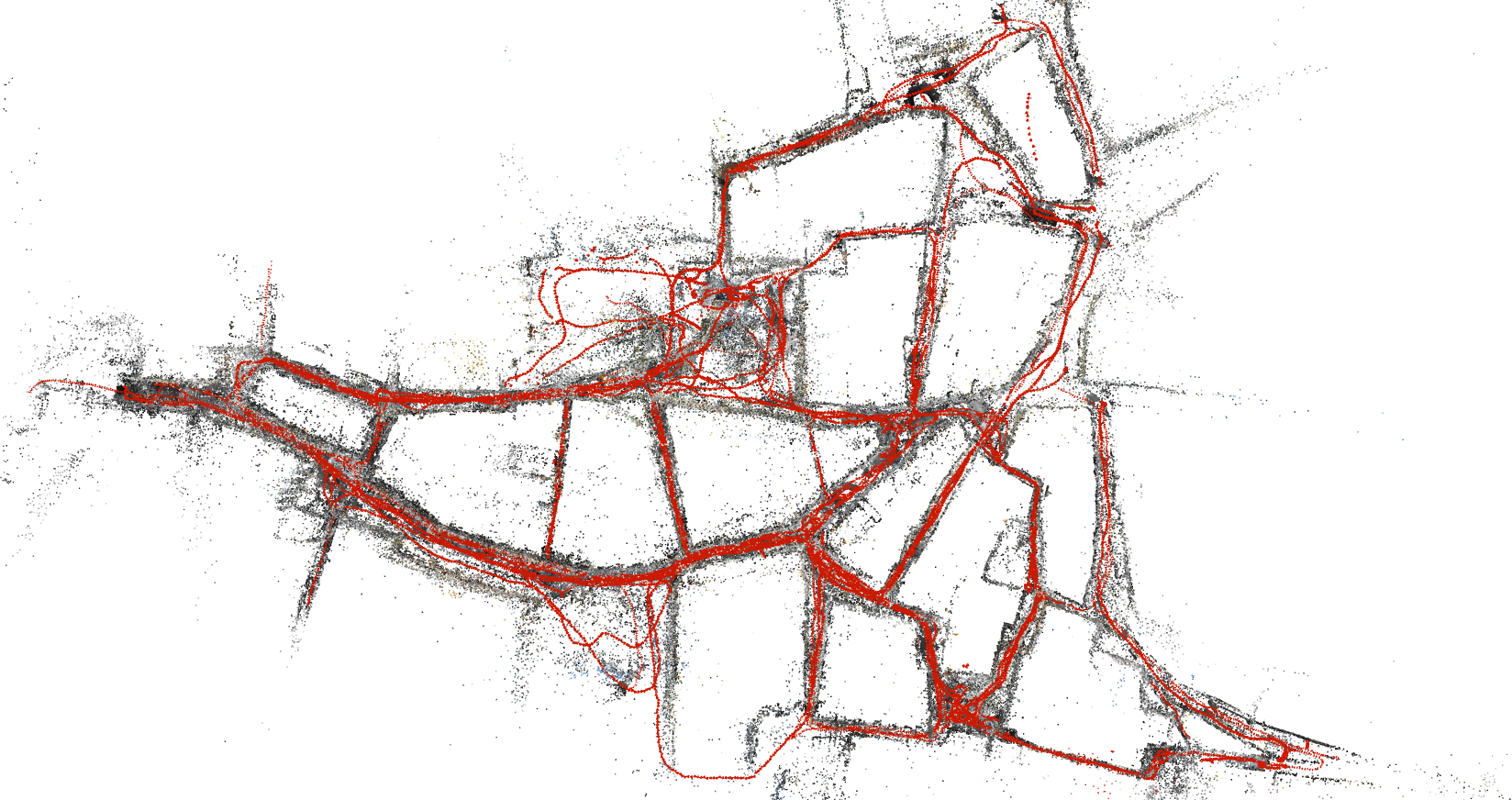}\label{fig:LIN}}
    \caption{This figure shows the qualitative results of LaMAR dataset \citep{sarlin2022lamar} reconstructed by HETA++}
    \label{fig:lamar}
\end{figure*}
\begin{figure*}	
	\setlength{\abovecaptionskip}{-0.4cm}
	\setlength{\belowcaptionskip}{-0.2cm}
	\vspace{-0.2cm}
    \centering
    \includegraphics[width=\linewidth, trim = 0mm 53mm 0mm 53mm, clip]{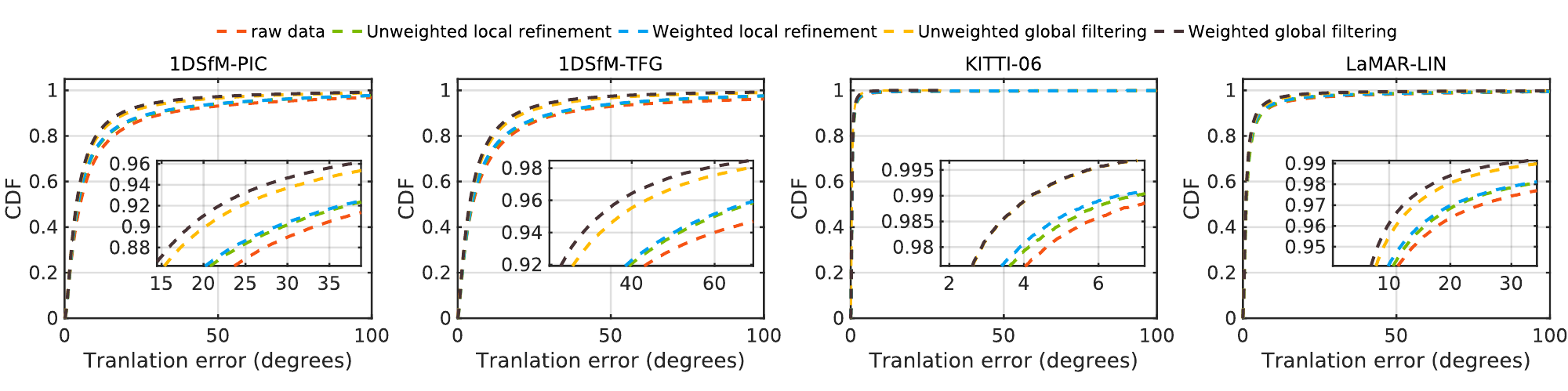}
	\caption{This figure shows the cumulative distribution functions of the relative translation angle errors for 1DSfM-PIC, 1DSfM-ROF, KITTI-06 and LaMAR-LIN, after local and global refinement with or without weights}
	\label{fig:CDF}
    \vspace{-0.2cm}
\end{figure*}
\subsubsection{LaMAR Dataset}
The quantitative reconstruction results are shown in Table~\ref{tab:lamar}. All methods perform poorly in scene CAB due to drastic illumination changes and repetitive facades, which lead to numerous outlier feature matches in the view graph. Our method achieves the best accuracy and efficiency in scene HGE. In scene LIN, HETA++ provides more accurate reconstructions, achieving the highest AUC score and the lowest median camera position error. Both GLOMAP and HETA++ register the same number of images, but in terms of efficiency, our method outperforms GLOMAP. The reconstruction results produced by our method are shown in Fig.~\ref{fig:lamar}, demonstrating the scalability of our hybrid explicit global SfM system for large-scale datasets.

\subsection{Ablation Study}

\subsubsection{Relative Translation Refinement}

The hyperparameters in Algorithm~\ref{algo:re-estimation} are set to $\beta=\sin(1^\circ)\sin(5^\circ)$, $T_\alpha=0.5^\circ$, $T_\beta=1^\circ$, and $N_m=20$ in all experiments.
We conduct experiments to evaluate the impact of both local and global refinement, with and without reasonable weights. 
We select four datasets to show the effectiveness of our relative translation refinement, including two large-scale unordered image datasets in 1DSfM~\citep{wilson2014robust}, one sequential dataset in KITTI \citep{geiger2013vision} and
one large-scale dataset in LaMAR \citep{sarlin2022lamar}.
The cumulative distribution functions of the relative translation errors are shown in Fig.~\ref{fig:CDF}. Both local re-estimation and global consistency filtering improve the relative translation accuracy, while retaining the parallax-dependent normal magnitudes further improves the results. For densely connected view graphs such as KITTI-06, the proposed refinement removes most relative translations with large angular errors, making the stricter directional constraints of the cross-product formulation suitable for camera position initialization.

\begin{table*}
	\caption{A comparison of three initialization methods with the same bundle adjustment (BA) module from GLOMAP~\citep{pan2024glomap}, including GLOMAP~\citep{pan2024glomap}, HETA~\citep{tao2024revisiting}, and HETA++ in the first three columns. A comparison of three refinement schemes with the same initialization method from HETA++, including BA in GLOMAP, joint optimization (JO) without feature track selection, and JO with feature track selection in HETA++ in the last three columns. $\tilde{e}$ and $\bar{e}$ denote the median and mean camera position errors (in meters), respectively. $T_1$ is the runtime of initialization (in seconds), and $T_2$ is the runtime of BA or JO (in seconds).}
	\setlength{\tabcolsep}{2.5pt}
	\centering
    \scalebox{0.75}{
        \begin{tabular}{|c||cc|cc||cc|cc||cc|cc||cc|cc||cc|cc|}
            \hline		
            \multicolumn{1}{|c||}{method} &
            \multicolumn{4}{c||}{{GLOMAP$_{init}$} + BA} &
            \multicolumn{4}{c||}{{HETA$_{init}$} + BA}&
            \multicolumn{4}{c||}{{HETA++$_{init}$} + BA} &
            \multicolumn{4}{c||}{{HETA++$_{init}$} + JO w/o S}&
            \multicolumn{4}{c|}{{HETA++$_{init}$} + JO w/ S}\\			
            \hline
            KITTI 
            & $\tilde{e}\downarrow$ & $\bar{e}\downarrow $ & $T_1\downarrow$ & $T_2\downarrow$
            & $\tilde{e}\downarrow$ & $\bar{e}\downarrow $ & $T_1\downarrow$ & $T_2\downarrow$
            & $\tilde{e}\downarrow$ & $\bar{e}\downarrow $ & $T_1\downarrow$ & $T_2\downarrow$
            & $\tilde{e}\downarrow$ & $\bar{e}\downarrow $ & $T_1\downarrow$ & $T_2\downarrow$
            & $\tilde{e}\downarrow$ & $\bar{e}\downarrow $ & $T_1\downarrow$ & $T_2\downarrow$\\
            \hline
            {00} 
            & 0.7 & \textbf{0.8} & 1586 & 1207
            & 0.7 & 0.9 & 735 & 1360
            & 0.7 & \textbf{0.8} & \textbf{280} & 1115
            & {0.7} & \textbf{0.8} &  \textbf{280} & 1284
            & \textbf{0.6} & \textbf{0.8} & \textbf{280}  & \textbf{901}\\  
            {02} 
            & 5.4 & 1e3 & 3163 & 2281
            & 4.6 & 8.2 & 586 & 1898
            &  3.8 & 8.1  & \textbf{417} & 1616
            & 3.2 & 6.0 & \textbf{417} & 2102
            & \textbf{2.1} & \textbf{2.9} & \textbf{417} & \textbf{1221}\\ 
            {05} 
            & \textbf{0.1} & \textbf{0.4} & 1020 & 1031 
            & \textbf{0.1} & \textbf{0.4} & 387 & 968
            & \textbf{0.1} & \textbf{0.4} & \textbf{197} & \textbf{861}
            & \textbf{0.1} & \textbf{0.4} & \textbf{197}& 1470
            & \textbf{0.1} & \textbf{0.4} & \textbf{197} & \textbf{915}\\ 
            {08} 
            & 3.0 & 4.4 & 2085 & 2239
            & 2.0 & 3.0 & 619 & 1927
            & \textbf{1.9} & 2.8 & \textbf{250} & 1513
            & 2.4 & 3.3  & \textbf{250} & 1415
            & {2.3} & \textbf{2.7}  & \textbf{250} & \textbf{810}\\ 
            \hline
            {Average} 
            & 2.3 & 2e2 & 1963 & 1689
            & 1.9 & 3.1 & 582 & 1538
            & {1.6} & 3.0 & \textbf{286} & 1276
            & 1.6 & 2.6 & \textbf{286} & 1735
            & \textbf{1.3} & \textbf{1.7}  & \textbf{286} & \textbf{962}\\ 
            \hline
            \multirow{2}{*}{1DSfM}
            &\multicolumn{2}{c|}{AUC@} & \multirow{2}{*}{$T_1\downarrow$} & \multirow{2}{*}{$T_2\downarrow$}
            &\multicolumn{2}{c|}{AUC@} & \multirow{2}{*}{$T_1\downarrow$} & \multirow{2}{*}{$T_2\downarrow$}
            &\multicolumn{2}{c|}{AUC@}& \multirow{2}{*}{$T_1\downarrow$} & \multirow{2}{*}{$T_2\downarrow$}
            &\multicolumn{2}{c|}{AUC@} &\multirow{2}{*}{$T_1\downarrow$} & \multirow{2}{*}{$T_2\downarrow$}
            &\multicolumn{2}{c|}{AUC@} &\multirow{2}{*}{$T_1\downarrow$} & \multirow{2}{*}{$T_2\downarrow$}\\
            & $3^\circ\uparrow$& $5^\circ\uparrow$ & &
            &$3^\circ\uparrow$& $5^\circ\uparrow$ & &
            &$3^\circ\uparrow$& $5^\circ\uparrow$ & & 
            &$3^\circ\uparrow$& $5^\circ\uparrow$ & &
            &$3^\circ\uparrow$& $5^\circ\uparrow$ & &\\
            \hline
            {PIC} 
            & 44.2 & 56.2  &455 & 25737
            & 44.2 & 56.3  & \textbf{150} & 29392
            & 46.1 & 57.9 & 192 & 29897
            & \textbf{46.8} & \textbf{58.6} & 192 & 18674
            & 46.3 & 58.1 & 192 & \textbf{13708}\\  
            {ROF} 
            & 67.8 & 75.9  & 357& 3791
            & 49.5 & 61.8  & 99 & \textbf{336}
            & 67.9 & 75.9 & \textbf{54} & 3953
            & \textbf{68.6}  & \textbf{76.6} & \textbf{54} & 3705
            & {68.4} & {76.2} & \textbf{54} & {2693} \\ 
            {TFG} 
            & 38.2 & 50.4  & 1451& 203537
            & 32.8 & 44.8  & 736 & 138640
            & 38.7  & 50.9  & \textbf{705}  & 124088
            & 39.0 & 51.3 & \textbf{705}& 74135
            &  \textbf{39.1} &  \textbf{51.4} & \textbf{705} & \textbf{55350} \\  
            {VNC} 
            & 35.1&49.9 & 231& 1587
            & 35.4&49.7 & 39 & 1547
            & 35.3  & 49.8  & \textbf{33}& 1290
            & \textbf{35.7} & \textbf{50.6}  &\textbf{33} & 1185
            &  35.4 & 50.3  & \textbf{33}& \textbf{1064}\\ 
            \hline
            {Average} 
            & 46.3&58.1 & 624& 58663
            & 40.5&53.2 & 256 & 42479
            & 47.0 & 58.6 & \textbf{246} & 39807
            & \textbf{47.5} & \textbf{59.3} & \textbf{246} & 24425
            &  47.3 & 59.0  & \textbf{246}& \textbf{18204}\\ 
            \hline 
        \end{tabular}
    } 	
    \vspace{-0.2cm}
	\label{tab:refinement}
\end{table*}
\subsubsection{Hybrid Explicit Initialization}
Compared to HETA~\citep{tao2024revisiting}, which jointly estimates initial camera positions and 3D points with selected feature tracks, HETA++ uses all feature tracks and decouples this step for efficiency. It first initializes camera positions using translation averaging and then estimates both camera positions and 3D points via alternating triangulation. These estimates are further refined with a non-bilinear angle-based objective function. In contrast, GLOMAP~\citep{pan2024glomap} directly estimates the camera positions and 3D points using a bilinear objective function with random initialization. 

We compare the accuracy and efficiency of these three initialization modules from GLOMAP~\citep{pan2024glomap}, HETA~\citep{tao2024revisiting}, and HETA++.
Four large-scale unordered image datasets from 1DSfM~\citep{wilson2014robust} and four large-scale sequential image datasets from KITTI~\citep{geiger2013vision} are used to demonstrate the effectiveness of our method.
For a fair comparison, we use the same view-track graph and bundle adjustment module from GLOMAP~\citep{pan2024glomap}, changing only the initialization module in the global SfM system, as shown in the first three columns of Table~\ref{tab:refinement}. 
Compared to GLOMAP~\citep{pan2024glomap} with random initialization, HETA++ achieves superior robustness, accuracy and efficiency.
Compared to HETA~\citep{tao2024revisiting}, our decoupled method improves both accuracy and efficiency in most testing cases, indicating the effectiveness of our upgraded approach.

Overall, by providing a reliable initialization and avoiding the explicit optimization of auxiliary scale variables for millions of feature-ray observations, the proposed decoupled hybrid explicit initialization substantially reduces the number of optimization variables and peak memory consumption. It also improves optimization efficiency while yielding more accurate camera positions.

\subsubsection{Joint Optimization}
We propose a joint optimization framework to progressively improve the accuracy of camera parameters and 3D points.
Prior to the final complete bundle adjustment, we first refine camera poses using selected feature tracks through angle-based refinement, followed by reprojection-based bundle adjustment.
Feature tracks are selected to balance the error distribution and enhance efficiency.
Given the same HETA++ initialization, we compare three refinement schemes: direct bundle adjustment (BA), joint optimization without feature-track selection (JO w/o Selection), and joint optimization with feature-track selection (JO w/ Selection). The results are reported in the last three columns of Table~\ref{tab:refinement}.
Comparing ``HETA++$_{\mathrm{init}}$ + BA'' with ``HETA++$_{\mathrm{init}}$ + JO w/o Selection'' demonstrates the benefit of bounded angle-based refinement before bundle adjustment. Further comparison with ``HETA++$_{\mathrm{init}}$ + JO w/ Selection'' shows that coverage-aware track selection substantially improves efficiency and KITTI accuracy while maintaining comparable accuracy on 1DSfM.

\begin{figure*}[htbp]
    \centering
     \subfloat[old\_computer]{\includegraphics[width=0.18\textwidth, trim = 0mm 0mm 0mm 0mm, clip]{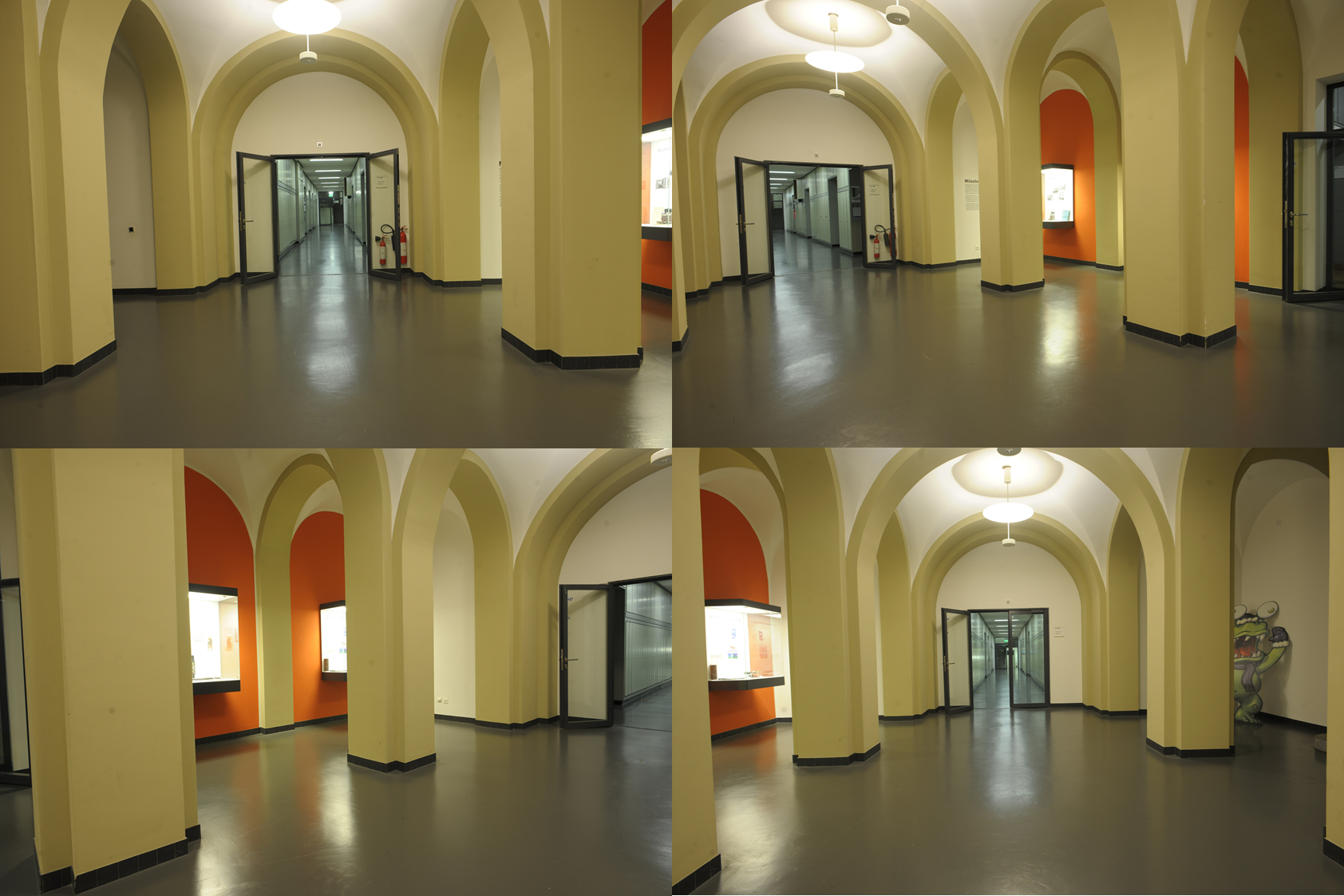}\label{fig:raw_computer}}
    \hfill
    \subfloat[COLMAP]{\includegraphics[width=0.205\textwidth, trim = -20mm 10mm 20mm -10mm, clip]{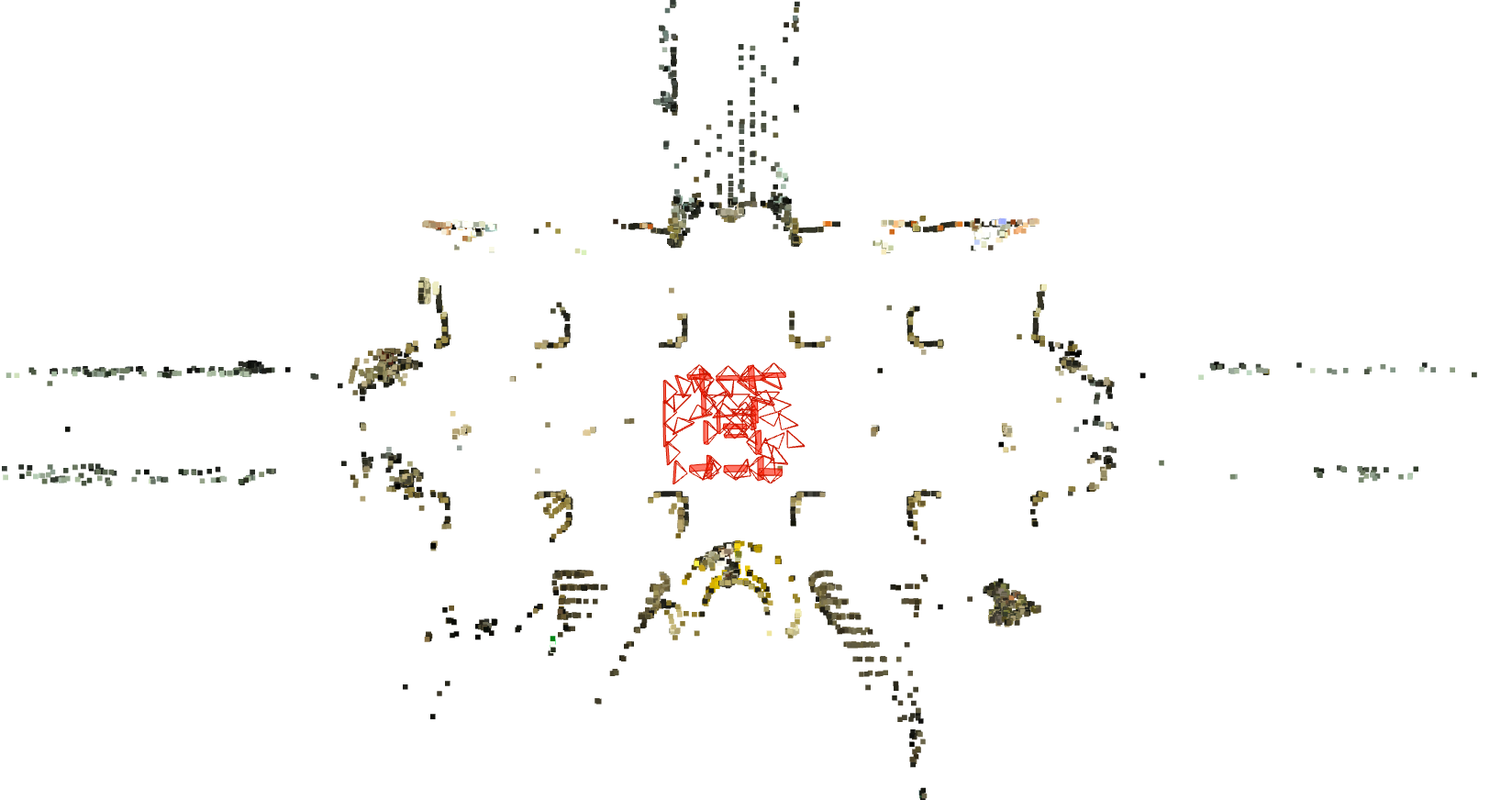}\label{fig:computer_colmap}}
    \hfill
     \subfloat[GLOMAP]{\includegraphics[width=0.205\textwidth, trim = -50mm 0mm 50mm 0mm, clip]{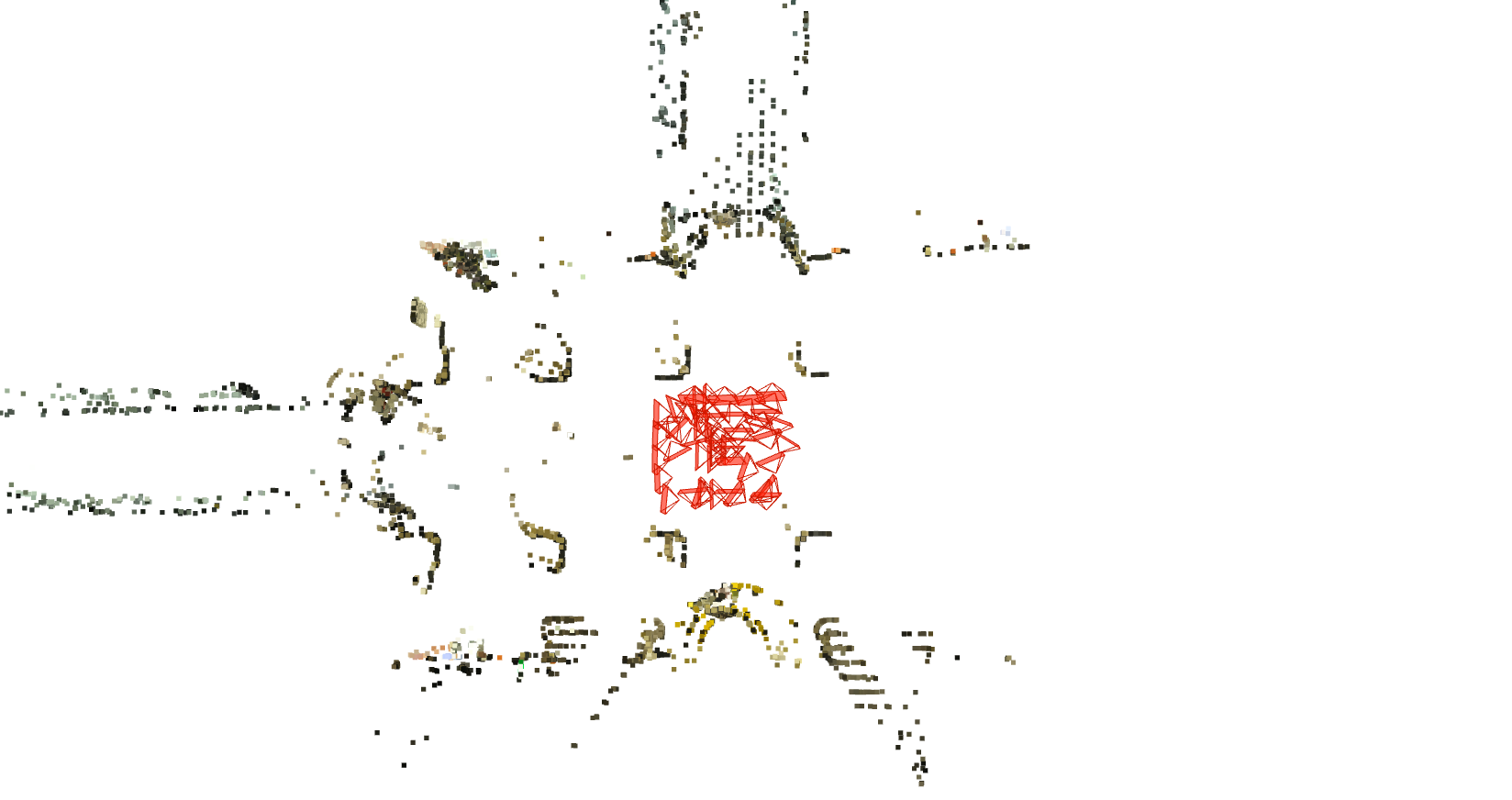}\label{fig:computer_glo}}
    \hfill
    \subfloat[HETA++]{\includegraphics[width=0.205\textwidth, trim = -30mm 0mm 30mm 0mm, clip]{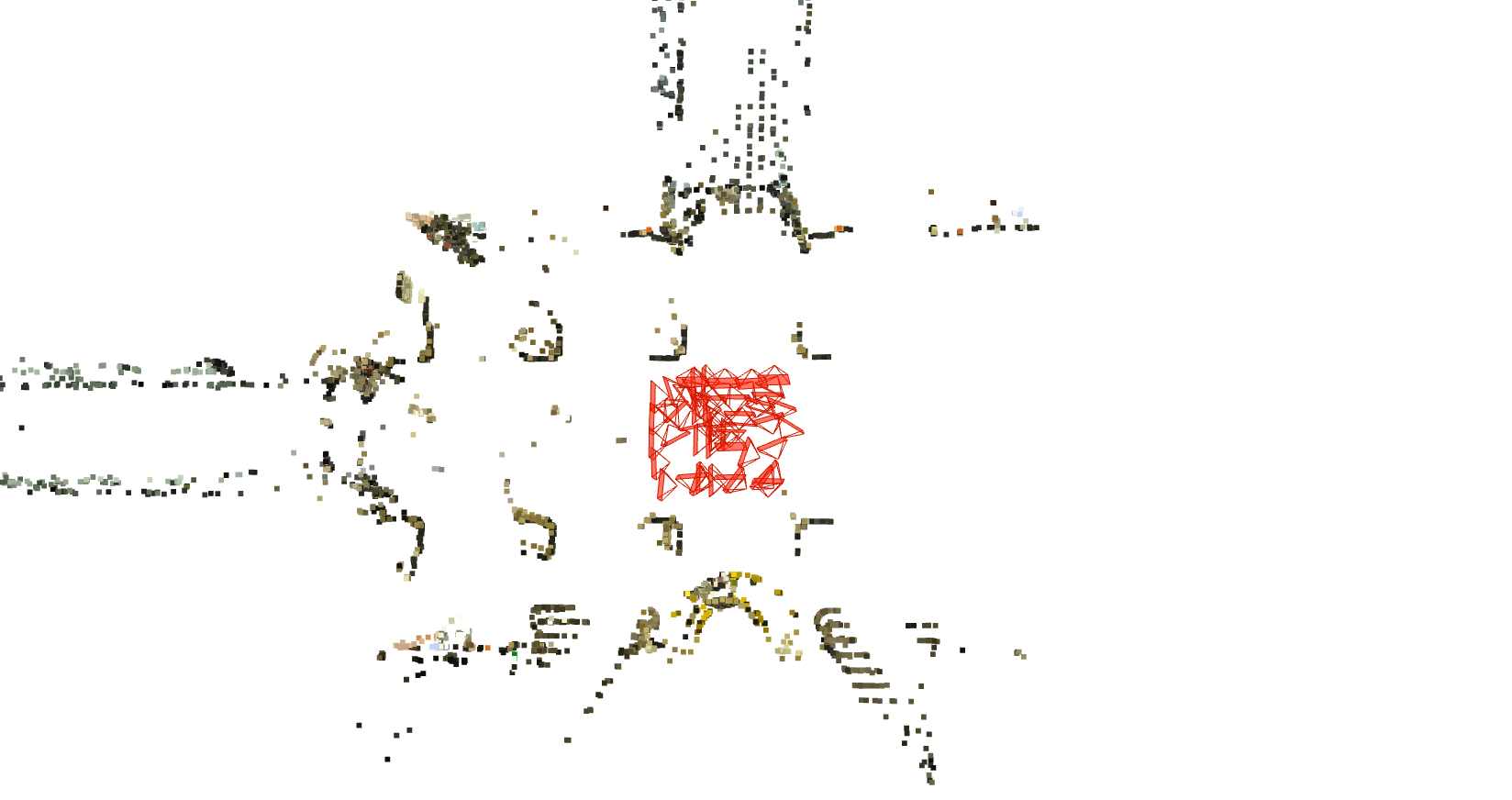}\label{fig:computer_hetapp}}
    \hfill
    \subfloat[HETA++ with camera-triplet-based filtering]{\includegraphics[width=0.205\textwidth, trim = 0mm 0mm 0mm 0mm, clip]{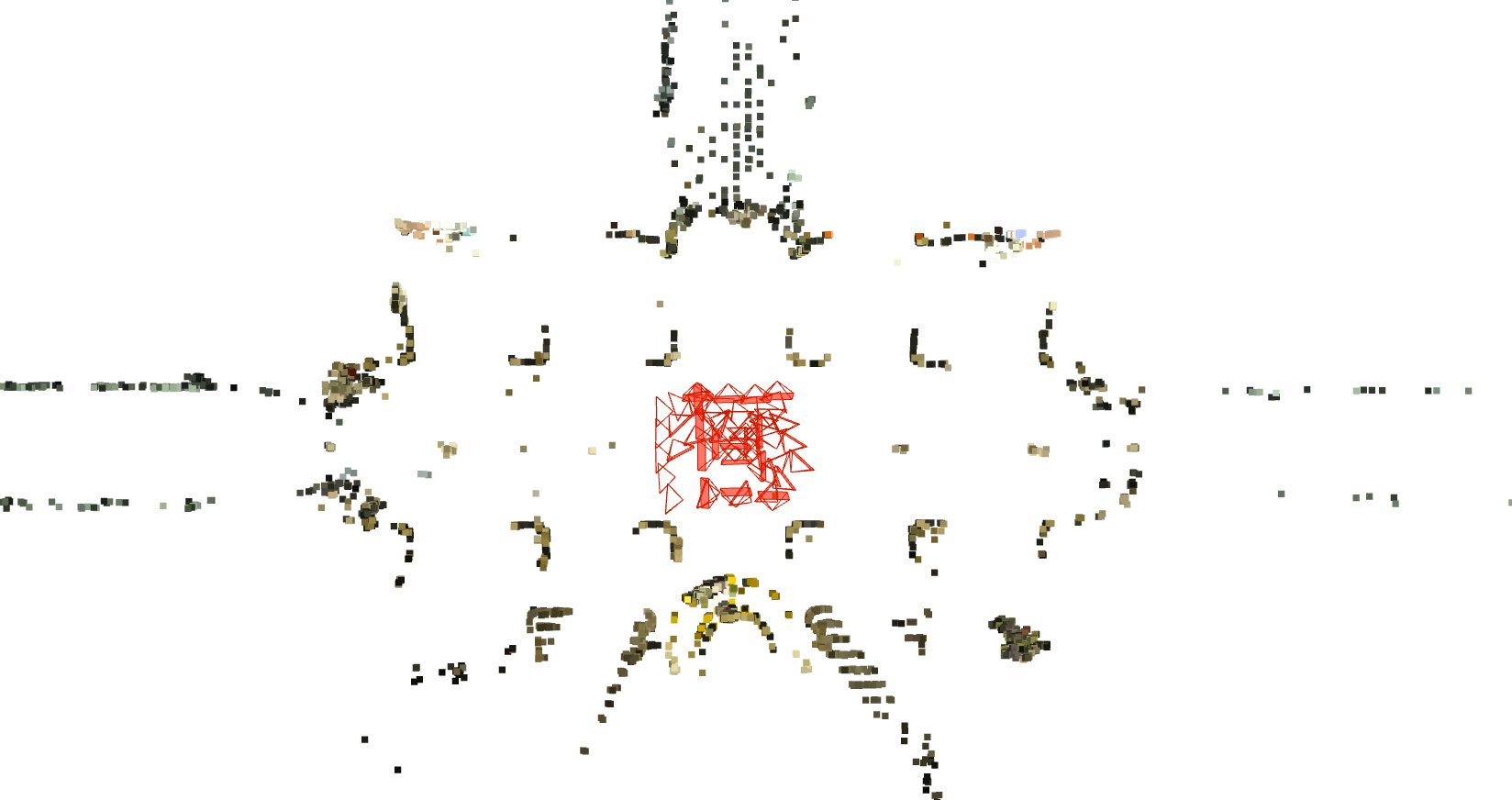}\label{fig:computer_trip}}\\
     \subfloat[Temple of Heaven]{\includegraphics[width=0.18\textwidth, trim = 0mm 0mm 0mm 0mm, clip]{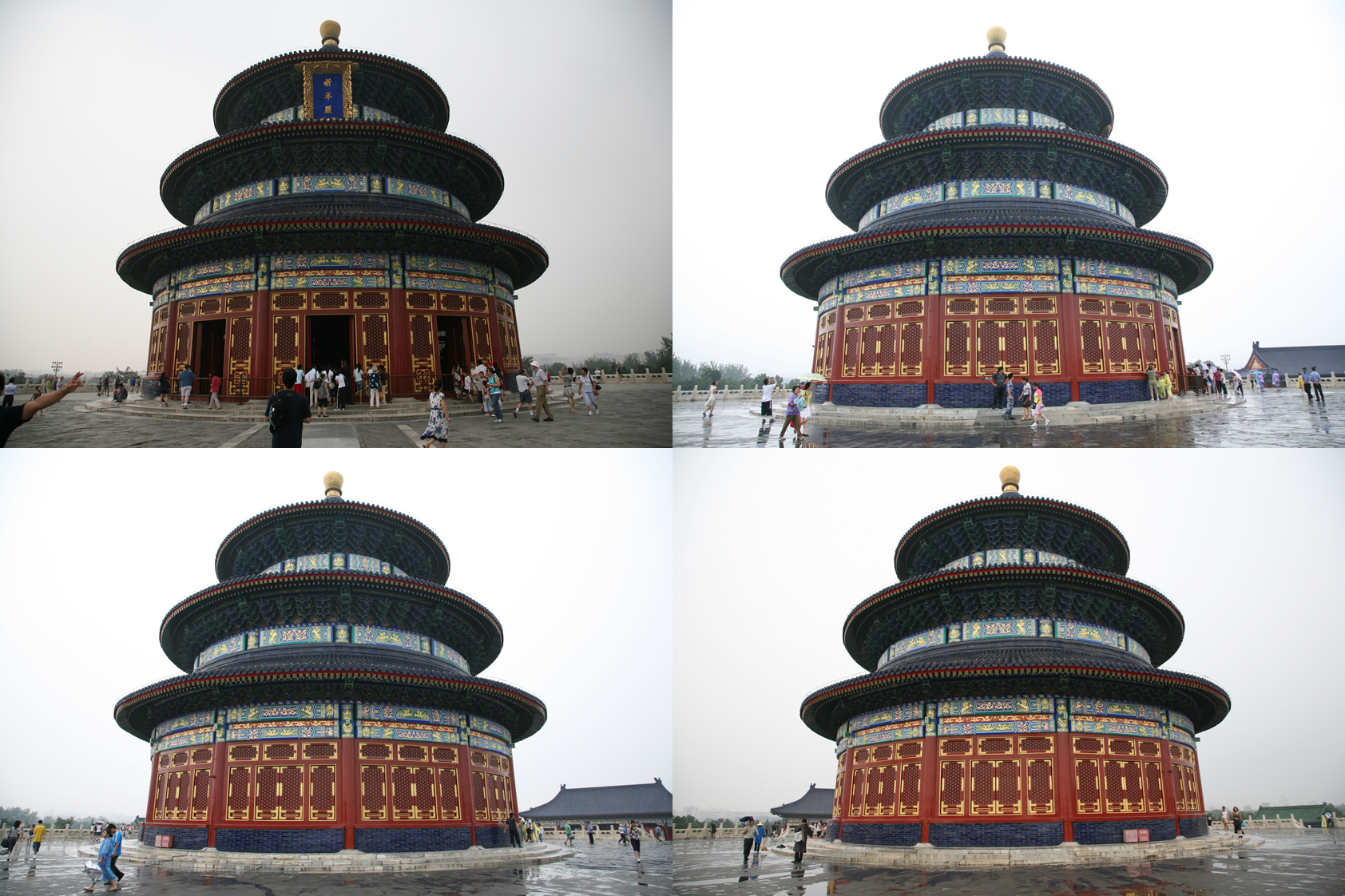}\label{fig:raw_TOH}}
    \hfill
    \subfloat[COLMAP]{\includegraphics[width=0.205\textwidth, trim = 20mm -5mm 20mm 15mm, clip]{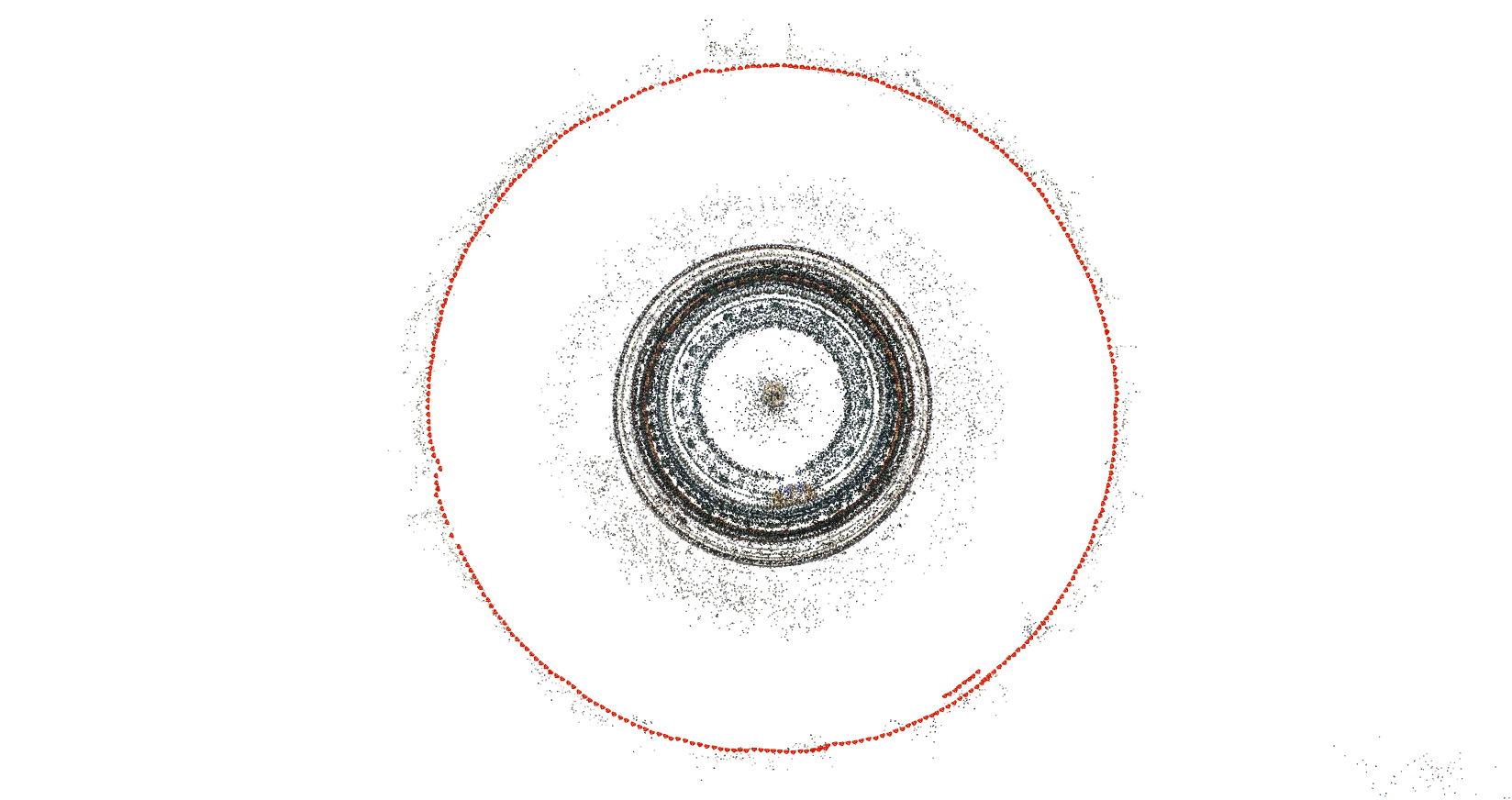}\label{fig:TOH_colmap}}
    \hfill
    \subfloat[GLOMAP]{\includegraphics[width=0.205\textwidth, trim = 98mm 45mm 98mm 52mm, clip]{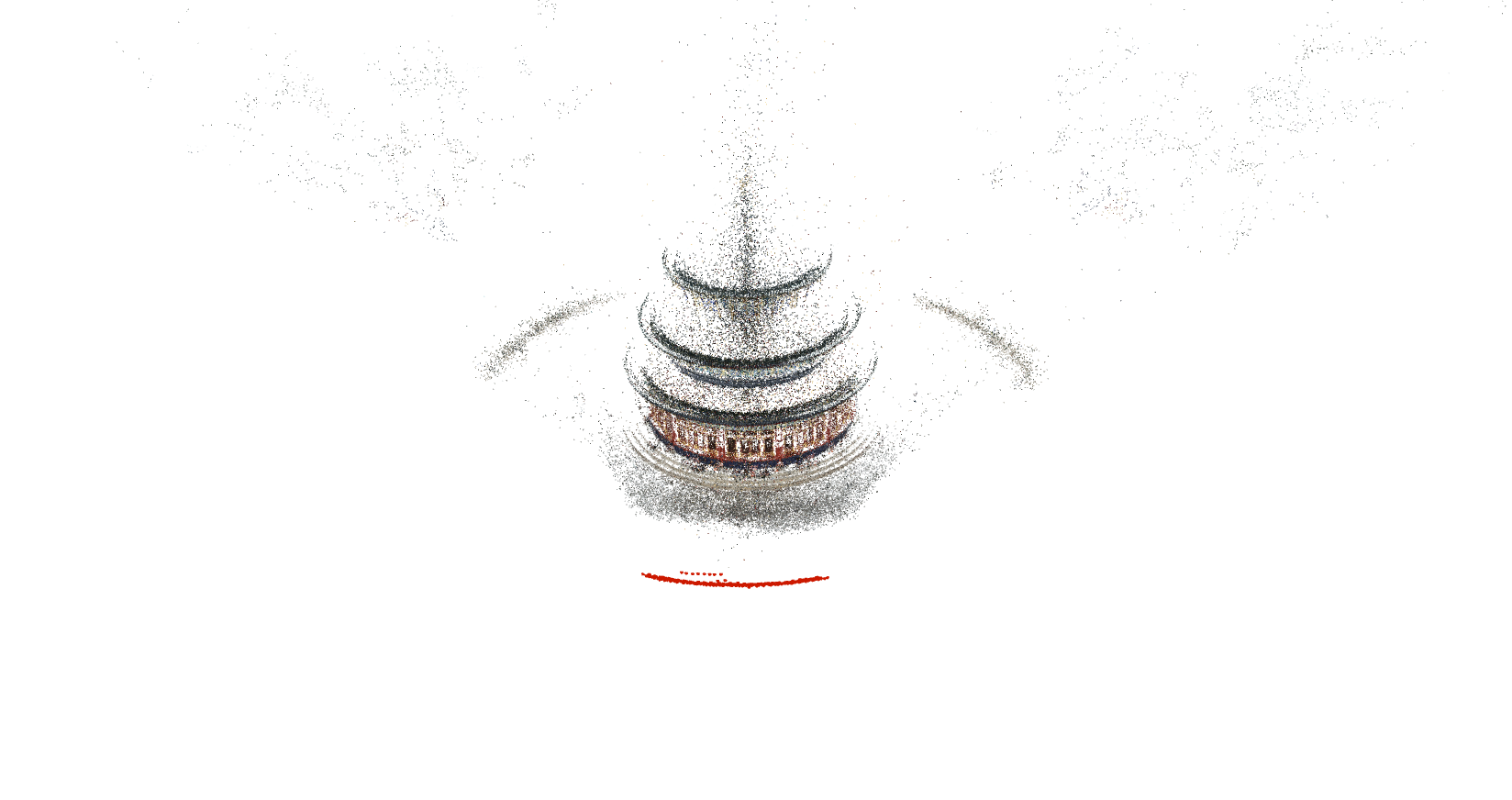}\label{fig:TOH_amb_glo}}
    \hfill
    \subfloat[HETA++]{\includegraphics[width=0.205\textwidth, trim = 98mm 45mm 98mm 57mm, clip]{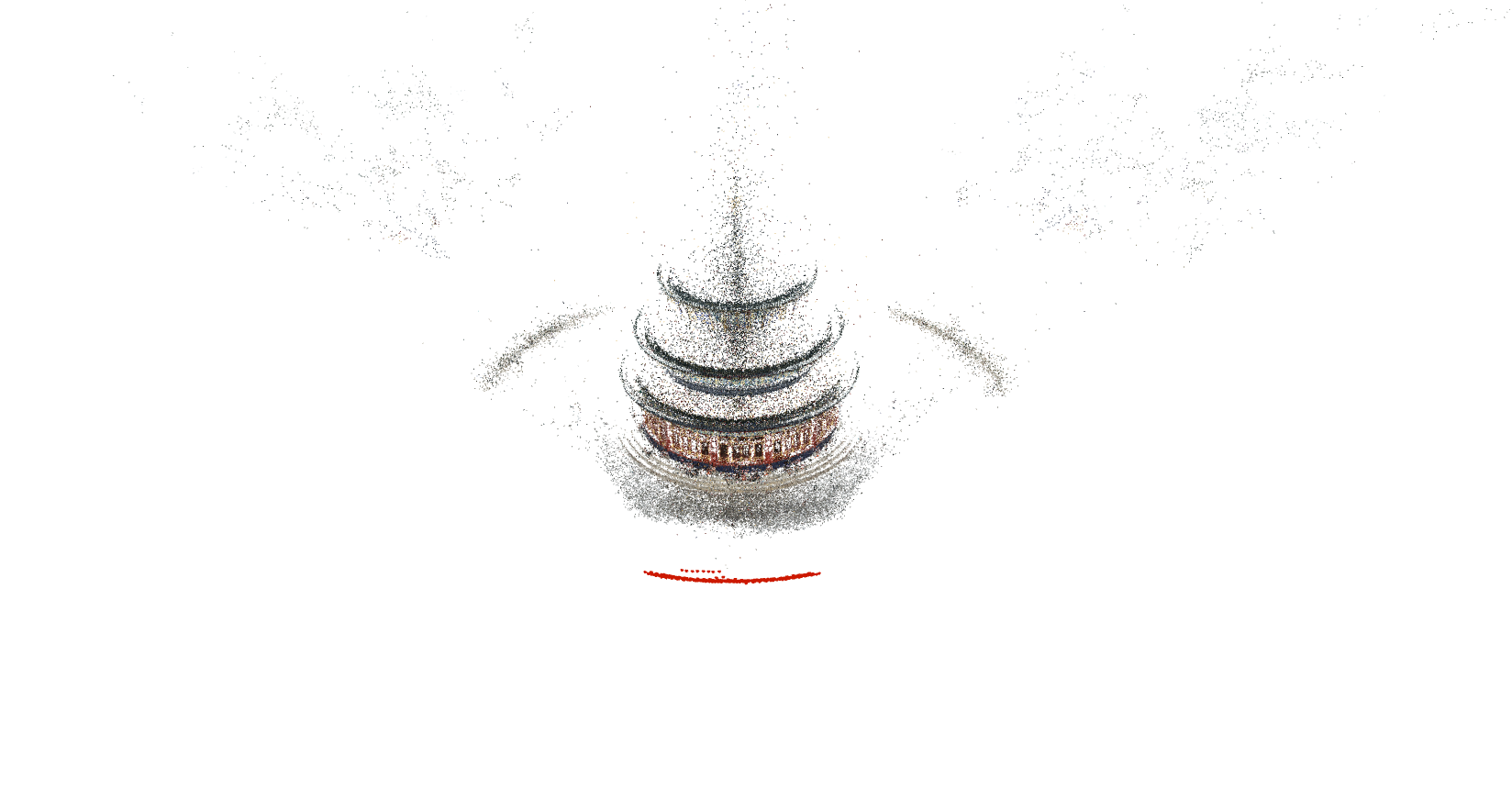}\label{fig:TOH_amb_heta}}
    \hfill
    \subfloat[HETA++ with camera-triplet-based filtering]{\includegraphics[width=0.205\textwidth, trim = 0mm -20mm 0mm 20mm, clip]{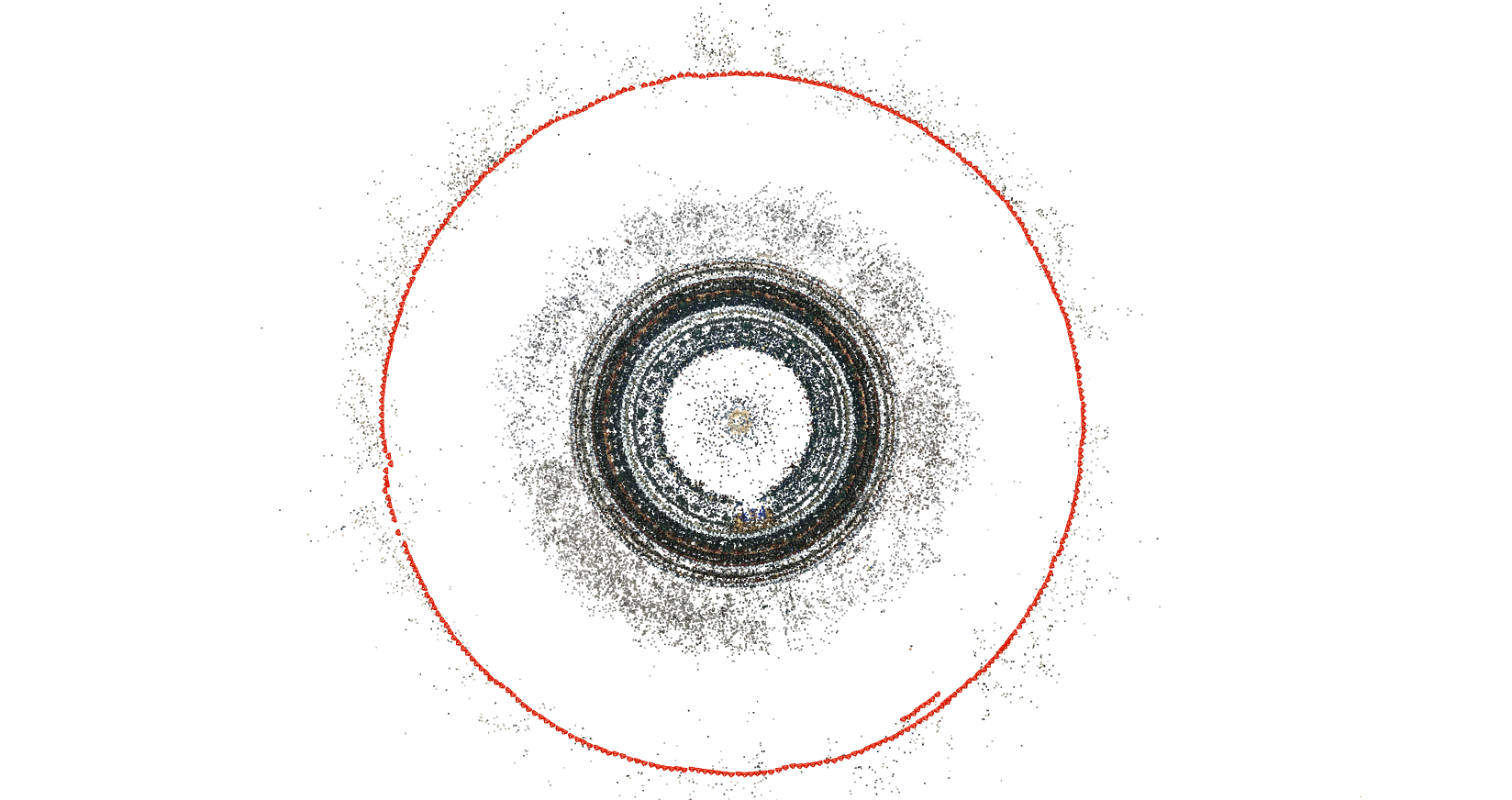}\label{fig:TOH_noamb}}
    \caption{Reconstruction results on two structurally ambiguous scenes:
    (a)--(e) ``old\_computer'' from ETH3D MVS~\citep{ETH3DDataset} and
    (f)--(j) Temple of Heaven~\citep{yan2017distinguishing}.
    Panels (a) and (f) show representative input images;
    (b) and (g) show the reconstructions produced by COLMAP;
    (c) and (h) show those produced by GLOMAP;
    (d) and (i) show those produced by HETA++;
    and (e) and (j) show HETA++ combined with the filtering method of \citet{manam2024camtripsfm}}
    \label{fig:ambigous}
    \vspace{-0.2cm}
\end{figure*}

For the KITTI dataset, camera rotations after rotation averaging exhibit relatively high accuracy. By selecting reliable feature points, our method achieves better convergence in data 02 and improved efficiency in data 08.
In contrast, for the 1DSfM dataset, the camera rotations are less accurate due to a large number of outlier relative rotations in the view graph. Refining camera poses with the bounded angle-based objective function improves accuracy, providing a solid starting point for the subsequent bundle adjustment, thereby enhancing both efficiency and accuracy.
Even without feature track selection, our joint optimization outperforms GLOMAP \citep{pan2024glomap} in most cases, highlighting the necessity of the angle-based camera pose refinement step.

\section{Conclusion and Discussion}
\label{sec:conclusion}
In this paper, we present a global Structure-from-Motion method with hybrid explicit global translation averaging, where ``explicit'' refers to the direct optimization of 3D points, and ``hybrid" signifies the use of both hybrid inputs and solvers, along with hybrid error metrics.
The key idea lies in providing a solid starting point for bundle adjustment to achieve robust and accurate reconstruction. 
We first employ a local-to-global scheme to refine relative translations based on geometric consistency. Then, given the hybrid refined relative translations and feature tracks, we propose a hybrid explicit method to initialize camera positions and 3D points using distance-based initialization followed by angle-based refinement. Finally, before the complete bundle adjustment, we robustly refine both camera rotations and positions with selected feature tracks to reduce reliance on the accuracy of initial camera rotations. This step includes bounded angle-based refinement followed by reprojection-based bundle adjustment. Various real-world experiments demonstrate the efficiency, accuracy, and robustness of our method.

Although our method shows robustness on many datasets, it still struggles with scenes containing ambiguous structures, such as symmetries and repetitive facades, as outlier feature matches cannot be filtered through local geometric consistency. As shown in Fig.~\ref{fig:ambigous}, both GLOMAP and HETA++ fail to reconstruct the scene old\_computer in ETH3D MVS \citep{ETH3DDataset} and the Temple of Heaven dataset \citep{yan2017distinguishing}. By integrating the state-of-the-art disambiguation module \citep{manam2024camtripsfm} to filter unreliable image matches during the view graph construction stage, our method reconstructs these scenes successfully, as shown in Fig.~\ref{fig:computer_trip} and Fig.~\ref{fig:TOH_noamb}.

For ambiguous data, the incremental SfM system can sometimes handle them, as demonstrated by the ``old\_computer'' dataset in Table~\ref{tab:dslr}. This is because, during incremental reconstruction, outliers are filtered repeatedly, and the RANSAC scheme is applied during camera registration. In the future, we will focus on this disambiguation task, specifically tackling incorrect edges in the view-track graph, which benefits both global rotation averaging and global translation averaging.

\backmatter

\section*{Statements and Declarations}

\subsection*{Funding}
This work was supported by the National Natural Science Foundation of China (Grant Numbers: U23A20386, 62572470 and 62402494)

\subsection*{Competing interests}
The authors declare no competing interests.

\subsection*{Data availability}
No new datasets were generated in this study. The KITTI, ETH3D, LaMAR, and 1DSfM datasets analyzed in this work are publicly available from their official repositories, which are cited in the manuscript.

\subsection*{Code availability}
The code will be made publicly available upon publication.

\subsection*{Author contributions}
Peilin Tao conceived the study, developed the methodology and software,
conducted the experiments, analyzed the results, and drafted the manuscript.
Hainan Cui contributed to the experimental design, analysis of the results,
and manuscript revision. Mengqi Rong contributed to data processing and
analysis of the results. Shuhan Shen supervised the project and revised the
manuscript. All authors reviewed and approved the final manuscript.

\bibliography{sn-bibliography}

\end{document}